\documentclass{article}
\usepackage{rapwam_preprint,times}
\usepackage[T1]{fontenc}
\usepackage[protrusion=true,expansion=true]{microtype}
\usepackage{amsmath,amssymb,graphicx,booktabs,tabularx,longtable,array}
\usepackage{algorithm,algpseudocode}
\usepackage{wrapfig,needspace}
\usepackage{hyperref,url}
\hypersetup{pdftitle={RAP-WAM: Learning World-Action Models through Predictive Rehearsal and Action Perturbations},pdfauthor={Yuanpeng He, Ceyao Zhang, Fangjing Li, Kefei Zhu, Lijian Li, Yueheng Li, Tianjia He, Wenpin Jiao, Zhi Jin, Yaodong Yang, Yuanpei Chen}}
\newcommand{\method}{RAP-WAM}
\newcommand{\puga}{response-based credit}
\newcommand{\priorfield}{\mathcal{P}}
\newcommand{\responsefield}{\mathcal{S}}
\newcommand{\errorfield}{\mathcal{E}}
\newcommand{\weightfield}{\mathcal{W}}

\title{RAP-WAM: Learning World-Action Models\\through Predictive Rehearsal\\and Action Perturbations}
\author{%
  \begin{tabular}{@{}c@{}}
    \fontsize{9.5}{11.5}\selectfont
    Yuanpeng He\textsuperscript{1,4}, Ceyao Zhang\textsuperscript{1,4},
    Fangjing Li\textsuperscript{2}, Kefei Zhu\textsuperscript{4},
    Lijian Li\textsuperscript{3}, Yueheng Li\textsuperscript{4}\\[-1pt]
    \fontsize{9.5}{11.5}\selectfont
    Tianjia He\textsuperscript{4}, Wenpin Jiao\textsuperscript{1},
    Zhi Jin\textsuperscript{1}, Yaodong Yang\textsuperscript{1,4},
    Yuanpei Chen\textsuperscript{4}\\[3pt]
    \fontsize{8}{10}\selectfont
    \textsuperscript{1}Peking University\qquad
    \textsuperscript{2}Beijing Jiaotong University\\[-2pt]
    \fontsize{8}{10}\selectfont
    \textsuperscript{3}University of Macau\qquad
    \textsuperscript{4}PKU-PsiBot Joint Lab
  \end{tabular}}

\AtBeginDocument{\parfillskip=0pt plus .70\textwidth\relax}
\AddToHook{env/table/begin}{\parfillskip=0pt plus 1fil\relax}
\AddToHook{env/thebibliography/begin}{\parfillskip=0pt plus 1fil\relax}
\begin{document}
\input{float_page_targets}
\maketitle
\begin{abstract}
Recent world action models draw on video generation priors to jointly predict future observations and robot actions. Adapting these priors to control requires determining how future predictions should guide policy learning. RAP-WAM combines predictive Rehearsal and Action Perturbations (RAP) with future-supervised queries and Coupled Action Projection for policy learning. Predictive rehearsal trains on model-generated inputs while retaining recorded targets, and accumulates recurring errors into a spatial prior. We perturb actions and measure changes in video predictions. Action-response-based spatiotemporal credit assignment combines these responses, prediction error, and the rehearsal prior to allocate visual supervision; spatially aggregating the responses determines supervision intensity across future action steps. No separate scoring network or deployment-time probe is required. The base learns from a video-pretrained backbone without intermediate pretraining on external embodied data. We evaluate the fixed base and its agent-refined, task-conditioned system. On 42 RoboDojo tasks, their average success rates are 12.85\% and 15.62\%, with Scores of 17.86 and 20.59. Across 15 real-robot tasks, the complete system achieves 89.00\% task-macro-average success, exceeding the strongest of five baselines by 37.67 percentage points. Across 315 fixed-output simulation windows, 45.21\% of response-top20 positions fall outside error-top20, revealing information beyond prediction error. Videos and further details are available on the project website: \url{https://rapwam.github.io/}.
\end{abstract}

\section{Introduction}
Generative robot policies combine visual representations with continuous action generation, as exemplified by Diffusion Policy and $\pi_0$ \citep{chi2023diffusion,black2024pi0}. The $\pi$ series has since advanced toward open-world generalization in $\pi_{0.5}$ \citep{physicalintelligence2025pi05} and multimodal steerability and compositional task generalization in $\pi_{0.7}$ \citep{physicalintelligence2026pi07}. World action models additionally predict future observations: UWM couples video and action diffusion \citep{zhu2025uwm}; DreamZero builds on pretrained video diffusion \citep{ye2026dreamzero}; X-WAM incorporates multi-view RGB-D prediction and asynchronous denoising \citep{guo2026xwam}.

OpenWAM studies backbone inheritance, world action interaction, and embodied-pretraining scale \citep{openwam2026project}. We focus on adapting video priors directly to target-task demonstrations: which visual locations and future action steps should receive additional training emphasis? Earlier work offers complementary perspectives. Prioritized replay changes how frequently experience transitions are reused \citep{schaul2016prioritized}, while value-aware model learning incorporates decision structure into prediction losses \citep{farahmand2017valueaware}. These motivate distinguishing prediction accuracy from policy relevance; allocating supervision across visual locations and future action steps in joint video--action models remains an open problem.

We introduce RAP-WAM, coupling predictive rehearsal with action perturbations to guide representations and supervision. Its central signal is the video prediction's response to perturbations of the model's action input. Prediction error measures target mismatch, whereas this response reveals which visual locations and future times depend more strongly on the action representation. Combining them distinguishes prediction difficulty from action dependence, providing a shared basis for prioritizing visual and action prediction.

The framework couples three designs. Recorded future observations supervise predictive queries, giving action generation access to future-oriented features. Predictive rehearsal conditions subsequent demonstration segments on model-predicted observations while retaining recorded targets \citep{huang2025selfforcing,bengio2015scheduled}, and accumulates recurring errors into a spatial prior. The credit mechanism integrates this prior with current error and action response: spatial maps guide visual supervision, while spatially pooled response weights future action steps. Together, these designs turn existing demonstrations into representation targets, rehearsed inputs, and adaptive supervision without extra region annotations or a separate scoring network.

\begin{figure}[t]
\centering\includegraphics[width=136mm]{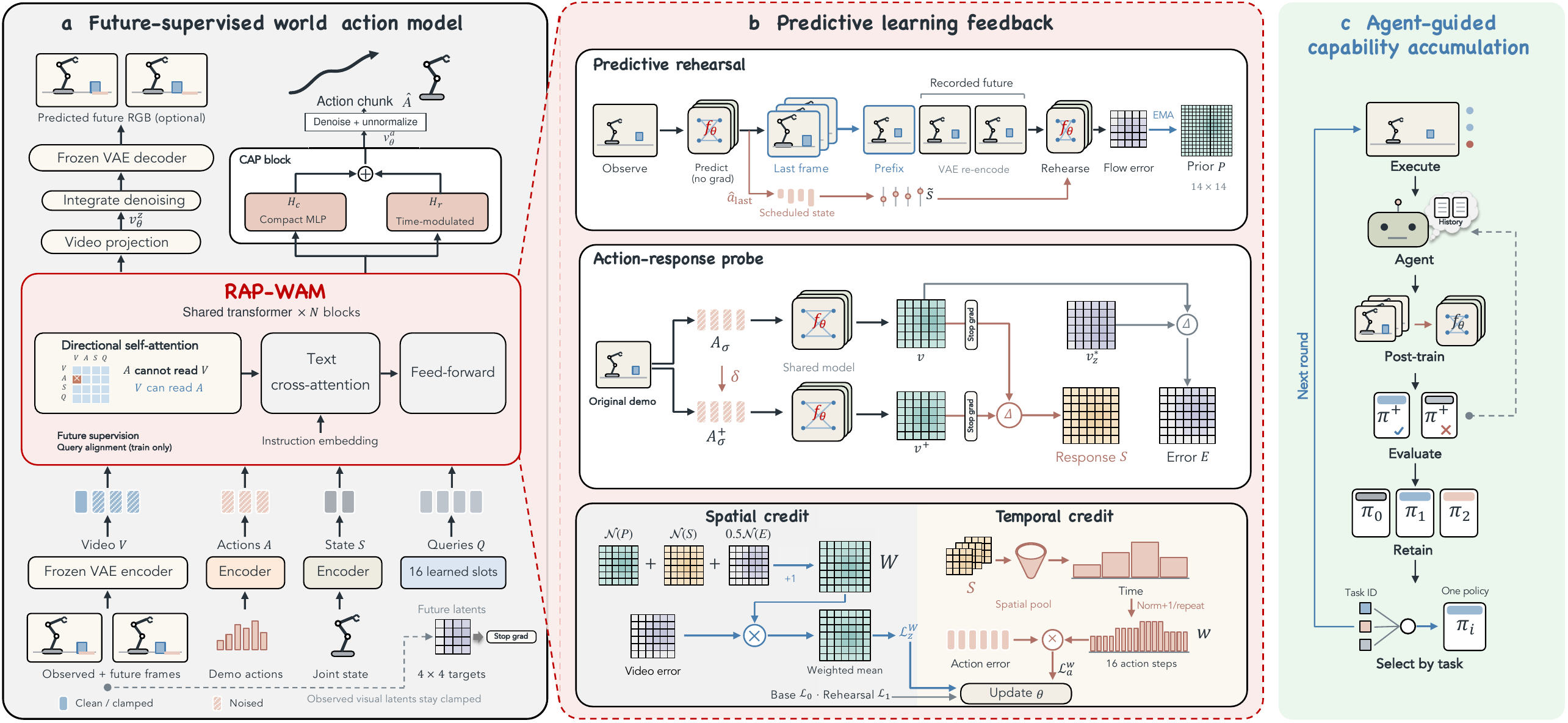}
\caption{\textbf{Predictive learning and agent-guided policy refinement.}
\textbf{a}, future latents supervise query alignment during training. CAP predicts action flow-matching velocity $v_\theta^a$; denoising and unnormalization yield action chunk $\hat A$.
\textbf{b}, rehearsal updates spatial prior $P$. The matched probe produces video-denoising velocity predictions $v$ and $v^+$ under original and perturbed action inputs, sharing model, condition, video noise, and diffusion level. $\Delta$ denotes channel-mean squared discrepancy: $S$ compares $v^+$ with $v$, and $E$ compares $v$ with flow-matching target $v_z^*$. Spatial weights combine separately normalized sources, $W=1+\mathcal N(P)+\mathcal N(S)+0.5\mathcal N(E)$. Temporal weights use $1+\mathcal N_t(\operatorname{mean}_{hw}S)$, repeated over action steps. Weight construction is detached.
\textbf{c}, the agent proposes post-training interventions. Evaluation retains improvements by task; other tasks keep their policies. Accepted and rejected outcomes inform subsequent rounds. Each task selects one complete policy. Scenes and fields are schematic.}
\label{fig:overview}
\end{figure}

Our contributions are as follows:
\begin{list}{\textbullet}{\setlength{\leftmargin}{1.5em}\setlength{\labelwidth}{.7em}\setlength{\labelsep}{.5em}\setlength{\topsep}{2pt}\setlength{\partopsep}{0pt}\setlength{\itemsep}{2pt}\setlength{\parsep}{0pt}}
\item We develop RAP-WAM to adapt video priors through future-supervised action representations and predictive rehearsal: recorded futures shape action-generating features, while rehearsal exposes the policy to observations affected by its own prediction errors.
\item We design an action-response-driven spatiotemporal credit mechanism: action response, current prediction error, and a historical error prior guide visual supervision, while spatially pooling the same response allocates supervision across future action steps.
\item We integrate execution-guided post-training with task-specific policy retention, using evaluation feedback to refine demonstration sampling and preserve complementary specializations.
\end{list}

\section{Related Work}
\paragraph{Generative control and world action models.}
Beyond the action and world action formulations introduced above, UniPi derives actions from video plans through inverse dynamics \citep{du2023unipi}. Motus and MotuBrain use multimodal expert architectures \citep{bi2025motus,motubrain2026}. GigaWorld-Policy emphasizes action-centered joint learning \citep{ye2026gigaworldpolicy}; AHA-WAM uses asynchronous video planning \citep{cai2026ahawam}, while Fast-WAM retains video co-training without test-time future imagination \citep{yuan2026fastwam}. Our focus is how existing target demonstrations supervise a joint policy. Further context on action interfaces and adaptive supervision appears in Appendix~\ref{app:extended-related}.

\paragraph{Post-training for embodied policies.}
RECAP improves $\pi_{0.6}$ into $\pi^{*}_{0.6}$ through reinforcement learning from demonstrations, autonomous experience, and expert corrections \citep{physicalintelligence2025pistar06}. SIME selects interaction segments \citep{jin2025sime}; RISE optimizes policies through imagined outcomes \citep{yang2026rise}. ENPIRE supports agent-directed experiments \citep{xiao2026enpire}, while ASPIRE retains validated program repairs \citep{lu2026aspire}. Our agent instead guides world action policy training on existing demonstrations; response-based credit allocates supervision, and evaluation retains complementary task specialists. The reported gains concern this combined procedure.

\section{RAP-WAM: Predictive Rehearsal and Action Perturbations}
\label{sec:method}
Figure~\ref{fig:overview} summarizes RAP-WAM: a future-supervised action interface, predictive rehearsal, and response-weighted supervision on target-task demonstrations. We separately evaluate post-training and task-specific selection, with full protocols in the appendix.

\subsection{A future-supervised action interface}
A shared video--action predictor does not automatically give its action
features an explicit representation of the future. We introduce a
future-supervised interface within that predictor; standard conditioning and
flow matching \citep{lipman2023flow} are described in Appendix~\ref{app:substrate}.
Recent interfaces supervise observed geometry \citep{li2025spatialforcing} or latent foresight \citep{ma2026internvlaa15}; ours aligns queries to recorded future video latents.
We append $K$ learned query tokens $Q$ to video $V$, noisy-action $A$, and observed-state $S$ tokens, forming $[V,A,S,Q]$. Video tokens contain the observed prefix and noisy future latents. Let $z$ be recorded-future latents and $h_k$ the final transformer representation of query $k$. The alignment target is
\begin{equation}
 r_k=\operatorname{sg}\!\left[P_t\operatorname{Pool}_{\mathcal G}(z)_k\right],
 \qquad
 \mathcal L_{\rm align}=\frac1K\sum_{k=1}^{K}
     [1-\cos(P_q h_k,r_k)].
 \label{eq:query-align}
\end{equation}
Here $P_t,P_q$ project targets and queries to a common 128-dimensional space; $\operatorname{sg}$ stops gradients. We pool $z$ onto a time--height--width grid $\mathcal G=(1,4,4)$, yielding $K=16$ targets. Alignment updates $P_q$ and query representations, while $P_t$ remains fixed. Learned query embeddings initialize $Q$ at each forward with action-side noise conditioning; recorded futures are needed only for training supervision. The base objective is
\begin{equation}
 \mathcal L_0=\mathcal L_z+\mathcal L_a+\lambda_q\mathcal L_{\rm align},
 \label{eq:base-objective}
\end{equation}
where $\mathcal L_z,\mathcal L_a$ are mean-squared errors against video and action flow-matching targets;
we set the alignment coefficient $\lambda_q=0.1$.
To encourage actions to use the future-supervised query features, we make attention connectivity directional. Within each block, we remove direct reads from action-token query rows
to video-token key columns. For their index sets $\mathcal A,\mathcal V$,
\begin{equation}
 M_{ij}=\begin{cases}
 -\infty,&i\in\mathcal A,\ j\in\mathcal V,\\
 0,&\text{otherwise}.
 \end{cases}
 \label{eq:directional-attention}
\end{equation}
The mask is added to attention logits. Action rows can read action, state, and predictive-query keys, but not video keys. Video rows can still read action keys, preserving the action-to-video path required by the response probe. State-mediated visual information remains accessible, so queries are not the sole visual pathway. GigaWorld-Policy also restricts future-video-to-action reads \citep{ye2026gigaworldpolicy}.
Coupled Action Projection (CAP) maps final action-token features $h_a$ to the action flow-matching velocity $v_\theta^a$. A compact MLP $H_c$ and a time-modulated head $H_r$ predict the same velocity target through their sum:
\begin{equation}
 \begin{aligned}
 v_\theta^a&=H_c(h_a)+H_r(h_a,e_\sigma),\\
 H_c(h)&=W_2\operatorname{GELU}(W_1\operatorname{LN}(h)+b_1)+b_2.
 \end{aligned}
 \label{eq:additive-readout}
\end{equation}
Here $H_r$ is the sequence head conditioned on the noise-level embedding $e_\sigma$;
$H_c$ uses a hidden width one quarter of the trunk width. Zero-initializing
$W_2,b_2$ adds this path without initially changing the sequence-head output.
Both branches learn through their sum against the same action-flow target,
not separate coarse/fine labels. The MLP has no separate noise-level input, although $h_a$ already depends on that noise level. The added capacity therefore
changes the action projection without introducing a second policy or task router.

\subsection{Rehearsal under predicted inputs}
Training uses recorded images and joint states; deployment instead supplies
observations affected by the robot's preceding actions \citep{ross2011dagger}.
Scheduled sampling \citep{bengio2015scheduled} and Self Forcing \citep{huang2025selfforcing} motivate exposing sequence models to their own predictions.
For two adjacent demonstration chunks, we generate the first chunk's future video and actions without gradients and replace the second chunk's initial observation with the last predicted frame. Scheduled state substitution optionally replaces its initial state with the last predicted absolute-joint action, first unnormalized as an action and then normalized as a state. The predicted image and recorded second-chunk future frames are jointly VAE-encoded; recorded actions remain the targets. Their video/action flow losses form $\mathcal L_1$, without an extra query-alignment loss. This \emph{rehearsal} trains on predicted inputs while retaining recorded supervision; its schedules are in Appendix~\ref{app:substrate}. An initially zero spatial prior $\priorfield$ accumulates the resulting prediction errors across updates.
We average detached second-chunk video errors over channels,
batch and future time, pool them onto a $g_P\times g_P$ grid to obtain
$e^{\rm roll}$, and update
\begin{equation}
 \priorfield\leftarrow\rho\priorfield+(1-\rho)e^{\rm roll}.
 \label{eq:prior}
\end{equation}
We use grid width $g_P=14$ and decay $\rho=0.9$.
The first valid error map initializes $\priorfield$ directly; later updates use the moving average above. Before weighting the video loss, we resize $\priorfield$ to the future-latent spatial grid and repeat it across future latent times. The predicted joint action approximates the next state, without simulator interaction or new corrective labels.
Unlike DAgger, this does not relabel visited states with an expert \citep{ross2011dagger}; it also differs from imagined-return optimization in RISE \citep{yang2026rise}.
The prior records where rehearsal predictions have repeatedly been inaccurate. The next step measures which predictions also depend on the action input.

\begin{table}[!t]
\centering\fontsize{8}{9.3}\selectfont\setlength{\tabcolsep}{1.5pt}
\caption{\textbf{RoboDojo comparison: Score / SR (\%).} Pretr.: external embodied pretraining reported for the method. Video/VLM initialization and target-task demonstrations are excluded. $*$: AHA-WAM uses robot-data pretraining for real-world experiments, but not in its default simulation setup (Appendix~\ref{app:wam}). Score and SR are ranked separately per column; highest and second-highest values are \textbf{bold} and \underline{underlined}, including ties. $\ddagger$: VLA supervised by a frozen video generation model during training. $\S$: matched 42-task coverage for our RAP-WAM base and post-training system; each task is averaged over its valid evaluation episodes.}
\label{tab:primary-leaderboard}\begin{tabular*}{\linewidth}{@{\extracolsep{\fill}}>{\raggedright\arraybackslash}p{35mm}crrrrrrr@{}}
\toprule
Model & Pretr. & Gen-Std & Gen-Rand & Precision & Long-Hor. & Memory & Open & Average \\
\midrule
\multicolumn{9}{@{}l}{\textit{VLA and other action policies}} \\
G0.5 \citep{liu2026g05} & Yes & 27.72/21.44 & \underline{9.20}/\underline{4.22} & \underline{28.25}/\underline{20.42} & 44.12/32.25 & 8.61/7.33 & 1.73/1.58 & \underline{20.23}/\underline{14.88} \\
Xiaomi-Robotics-1 \citep{xiaomi2026robotics1} & Yes & \textbf{35.65}/\textbf{28.00} & \textbf{11.44}/\textbf{6.00} & 26.69/18.83 & 38.39/23.67 & 7.81/6.56 & 3.94/3.58 & 20.07/13.93 \\
Hy-Embodied-0.5-VLA \citep{zhang2026hyvla} & Yes & 21.98/17.00 & 1.57/0.00 & 13.81/8.00 & 25.74/14.92 & \underline{13.37}/\underline{12.11} & 0.65/0.58 & 13.07/8.80 \\
KinRT \citep{yang2026kinrt} & Yes & 20.94/14.78 & 7.09/2.44 & 15.65/9.92 & 26.40/18.08 & 4.82/3.56 & \underline{4.23}/\underline{3.83} & 13.02/8.80 \\
SimpleMemVLA \citep{yin2026simplememvla} & No & 10.38/7.11 & 2.33/0.78 & 7.42/2.92 & 14.58/5.50 & \textbf{33.71}/\textbf{33.22} & 0.85/0.75 & 12.58/9.27 \\
Pi-05 \citep{physicalintelligence2025pi05} & Yes & 20.93/15.00 & 5.82/1.00 & 12.40/5.50 & 23.54/14.67 & 5.78/4.56 & 1.98/1.67 & 11.41/6.91 \\
InternVLA-A1.5$^{\ddagger}$ \citep{ma2026internvlaa15} & Yes & 16.81/12.00 & 3.90/2.00 & 15.23/10.17 & 23.80/13.75 & 4.93/3.56 & 1.43/1.42 & 11.15/7.14 \\
VLAct \citep{yang2026vlact} & Yes & 16.33/12.00 & 2.74/1.00 & 20.62/15.25 & 20.12/13.67 & 0.66/0.56 & 2.37/2.25 & 10.66/7.60 \\
MolmoAct2 \citep{fang2026molmoact2} & Yes & 12.53/7.33 & 4.53/2.00 & 10.92/4.50 & 15.20/7.00 & 5.53/4.56 & \textbf{4.75}/\textbf{4.42} & 8.99/5.03 \\
\midrule
\multicolumn{9}{@{}l}{\textit{World-model-based policies}} \\
OpenWAM-$\alpha$ \citep{openwam2026project} & Yes & \underline{33.16}/\underline{25.56} & 8.26/4.11 & 18.45/9.25 & 34.93/25.33 & 10.41/9.11 & 1.41/1.08 & 17.18/11.92 \\
X-WAM \citep{guo2026xwam} & Yes & 11.24/5.00 & 3.54/1.00 & 6.72/1.83 & 17.47/9.08 & 6.32/4.67 & 0.57/0.25 & 7.69/3.83 \\
GigaWorld-Policy-0 \citep{ye2026gigaworldpolicy} & Yes & 10.28/6.00 & 0.41/0.00 & 6.15/1.83 & 15.51/8.92 & 3.46/2.22 & 0.54/0.50 & 6.20/3.27 \\
AHA-WAM \citep{cai2026ahawam} & Yes$^{*}$ & 10.32/6.00 & 1.26/0.00 & 5.86/2.42 & 8.61/2.67 & 2.97/2.78 & 0.88/0.83 & 4.82/2.39 \\
Fast-WAM \citep{yuan2026fastwam} & No & 4.33/2.00 & 0.34/0.00 & 1.96/0.00 & 9.14/5.17 & 3.55/3.44 & 0.42/0.42 & 3.48/2.03 \\
\midrule
\multicolumn{9}{@{}l}{\textit{World-model-based: ours}} \\
RAP-WAM (base)$^\S$ & No & 21.55/15.00 & 0.82/0.00 & 23.97/16.51 & \underline{45.12}/\underline{32.50} & 7.97/7.00 & 1.06/0.75 & 17.86/12.85 \\
RAP-WAM + refinement$^\S$ (Ours) & No & 27.82/21.00 & 0.93/0.00 & \textbf{28.66}/\textbf{21.03} & \textbf{48.43}/\textbf{36.50} & 10.32/9.33 & 1.16/0.75 & \textbf{20.59}/\textbf{15.62} \\
\bottomrule
\end{tabular*}

\end{table}

\subsection{Response-derived spatiotemporal credit}
\label{sec:complementarity}
Prioritized replay and focal loss allocate learning emphasis using error signals \citep{schaul2016prioritized,lin2017focal}. Our credit mechanism adds action-conditioned response and shares it across visual and action supervision.
\puga{} compares two forwards of the same recorded demonstration. We draw fresh, independent noise for its future latents and actions at a shared noise level, retaining the recorded conditioning inputs, and compute a reference prediction with gradients. A second forward uses the same model, condition, video noise, and noise level but adds $\delta\sim\mathcal N(0,\tau^2I)$ to the normalized noisy actions; this forward has no gradients. At future spatiotemporal position $i$, let $v_i,v_i^+$ denote the reference and perturbed video velocities with $C$ channels.
The target is $v_{z,i}^*=\epsilon_{z,i}-z_i$, where $z_i$ is the clean recorded-future latent and $\epsilon_{z,i}\sim\mathcal N(0,I)$ is its sampled noise, consistent with $z_\sigma=(1-\sigma)z+\sigma\epsilon_z$. Here $i$ indexes a future spatiotemporal location and $*$ denotes the supervision target.
Holding these predictions constant for coefficient construction gives
\begin{equation}
 \responsefield_i=\frac1C\|v_i^+-\operatorname{sg}(v_i)\|^2,\qquad
 \errorfield_i=\frac1C\|\operatorname{sg}(v_i)-v_{z,i}^*\|^2.
 \label{eq:ld}
\end{equation}
We set the perturbation scale $\tau=0.1$. $\errorfield$ measures squared prediction error;
$\responsefield$ measures squared change after the action perturbation.
These are changes in model outputs, not measured physical effects of executed actions
(Appendices~\ref{app:response-proof}--\ref{app:complementarity}).
For each future visual position, we normalize the prior $\priorfield$, response $\responsefield$, and error $\errorfield$ separately, then combine them into a video-loss weight. Here $\mathcal N$ is per-sample min--max normalization over future time and spatial positions, with denominator floor $\epsilon_{\rm n}$:
\begin{equation}
 \weightfield_i
 =1+\alpha_P\mathcal N(\priorfield)_i+\alpha_S\mathcal N(\responsefield)_i
       +\alpha_E\mathcal N(\errorfield)_i.
 \label{eq:spatial-weight}
\end{equation}
The coefficients are nonnegative; we set $\alpha_P=\alpha_S=1$,
$\alpha_E=0.5$, and $\epsilon_{\rm n}=10^{-6}$.
The unit floor gives every location a nonzero loss coefficient, capped at 3.5 with these settings.
To weight action steps, we first average the raw response over spatial positions, then normalize these means across future time:
\begin{equation}
 c_t=1+\alpha_S\mathcal N_t(\operatorname{mean}_{hw}\responsefield)_t,\quad
 w_j=c_{\lfloor j/R\rfloor},\quad R=\lceil H_a/T\rceil.
 \label{eq:temporal-weight}
\end{equation}
Here $H_a$ is the action-chunk length, $T$ the number of future latent steps, and $\mathcal N_t$ min--max normalization over those $T$ steps within each sample. Using zero-based indices, each coefficient is repeated $R$ times and truncated to $H_a$ actions; the reported $T=4,H_a=16$ setting assigns one coefficient to four consecutive actions.
This branch uses the
raw response before spatial min--max normalization. The prior is constant
along time, so its normalized temporal contribution is zero; $\errorfield$
does not enter this branch. Hence $1\leq w_j\leq1+\alpha_S$. Visual and action prediction share one response source, with separate spatial and temporal weighting.
Action coefficients depend on normalized differences between future-step
spatial means, not on the locations of individual response peaks.

\begin{figure}[t]
\centering\includegraphics[width=\linewidth]{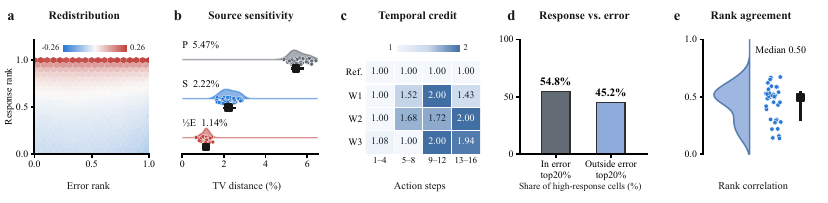}
\caption{\textbf{Complementary response signals and spatiotemporal supervision.}
\textbf{a}, signed spatial credit redistribution. \textbf{b}, sensitivity to removing prior, response, or half-error. \textbf{c}, temporal action credit in three phone-packing windows (W1--W3), with a unit reference. \textbf{d}, response-top20 membership within or outside the error-top20 set. \textbf{e}, rank correlation across 35 condition medians. Panels a, b, d, and e use the same 315 RoboDojo windows. Squares and segments indicate medians and IQRs; these fixed-output diagnostics involve no retraining.}
\label{fig:allocation-fingerprint}
\end{figure}

We apply these weights to video and action prediction errors. Let $i$ enumerate all future video locations across the batch and $j$ all action steps across the batch. At current parameters $\bar\theta$, normalize the detached loss coefficients as
$q_i^v=\weightfield_i/\sum_\ell\weightfield_\ell$ and
$q_j^a=w_j/\sum_\ell w_\ell$. For channel-mean video errors $e_i^z$ and
action-dimension-mean errors $e_j^a$, the auxiliary objective is
\begin{equation}
 \widetilde{\mathcal L}_P(\theta;\bar\theta)
 =r_P(k)\left[
 \lambda_z\sum_i q_i^v(\bar\theta)e_i^z(\theta)
 +\lambda_a\sum_j q_j^a(\bar\theta)e_j^a(\theta)\right],
 \label{eq:puga-objective}
\end{equation}
where $k$ counts training updates, $W_P$ is the credit warmup length, and $r_P(k)=\min(1,k/W_P)$; we set $\lambda_z=0.5$ and $\lambda_a=0.3$.
The full update uses
$\mathcal L=\mathcal L_0+b_c\lambda_c r_c(k)\mathcal L_1+
b_P\widetilde{\mathcal L}_P$, with $b_c\sim\operatorname{Bernoulli}(p_c(k))$
and $b_P\sim\operatorname{Bernoulli}(p_P)$.
Here $r_c(k),p_c(k)$ are the rehearsal warmup and invocation schedules (Appendix~\ref{app:substrate}); $\lambda_c=1$ and $p_P=0.5$.
The video error $e_i^z$ equals $\errorfield_i$ numerically but retains gradients. Both video and action errors reuse the reference forward. Weight construction is detached, so backpropagation updates the predictor through these errors only; no additional forward is needed for this auxiliary loss.
Unlike importance sampling with corrective weights \citep{katharopoulos2018importance}, this deliberately changes the auxiliary loss emphasis.
Positive, bounded coefficients retain every target while regulating relative
emphasis, not gradient norms or capability preservation
(Appendix~\ref{app:surrogate-proof}). No separate region-scoring network or reward head is needed.

The base objective supervises predictions and query features, rehearsal
substitutes predicted images and states, and credit scales individual loss terms.
At deployment, the queries, directional mask, and CAP
remain active during joint video-latent/action denoising. We return
unnormalized actions without decoding future RGB; neither rehearsal nor
credit estimation is invoked. Architecture and sampling details are given in
Appendix~\ref{app:substrate}.

\subsection{Post-training and task-specific policy selection}
\label{sec:agentic}
\begin{figure}[t]
 \centering
 \includegraphics[width=136mm]{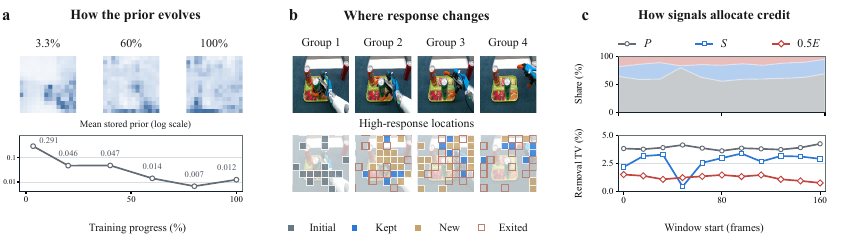}
 \caption{\textbf{Credit across three timescales.}
 \textbf{a}, prior spatial patterns at 3.3\%, 60\%, and 100\% of the same can-placement run (top, independently normalized), and the raw spatial mean at six checkpoints (bottom, logarithmic axis).
 \textbf{b}, response-top40 locations across four future groups for the newly trained three-can tray-placement policy.
 \textbf{c}, stacked component shares (top) and fixed-output removal sensitivity (bottom) for $\priorfield$, $\responsefield$, and $0.5\errorfield$ across all eleven exported windows of the same demonstration. Shares exclude the unit floor of $\weightfield$; sensitivity is total variation (TV) between normalized full and component-removed weights, retaining the unit floor.
 Panel c uses all spatial cells and future groups. All panels use the same training run; b--c use its final checkpoint. Definitions and display protocols appear in Appendix~\ref{app:real-maps}.}
 \label{fig:light-credit-dynamics}
\end{figure}

We evaluate the fixed RAP-WAM base and a refined system separately. The latter extends the learner with an experiment state $X_r=(\mathcal B_r,g_r,\mathcal H_r)$: available policies, task assignments, and intervention history. The initial repertoire contains the base. Following execution-feedback approaches \citep{xiao2026enpire,lu2026aspire}, task evaluation informs subsequent training; accepted specialists remain available rather than replacing the base on every task.

\paragraph{From evidence to a training intervention.}
At round $r$, the agent receives task outcomes, execution evidence, and available diagnostics, denoted $\mathcal F_r$. Together with history $\mathcal H_r$, these specify an intervention $u_r$: the parent checkpoint, target behaviors, demonstration allocation, and trainable parameter set. Here $\mathcal A$ is the agent, $\pi_r$ the selected parent policy, $\mathcal D_r$ the resampled demonstration windows, and $\mathcal U_{\rm RAP}$ the training operator producing candidate $\pi_r^+$:
\begin{equation}
 u_r=\mathcal A(\mathcal F_r,\mathcal H_r),\qquad
 \pi_r^+=\mathcal U_{\rm RAP}(\pi_r,\mathcal D_r;u_r).
 \label{eq:agent-intervention}
\end{equation}
Agent-guided refinement yields multiple task specializations with only a few thousand fine-tuning steps. Parent policies and rejected outcomes remain available.

\paragraph{Replay under recorded targets.}
We implement the replay intervention with 50\% natural replay, 20\% task-balanced replay, 15\% failure-boundary windows, and 15\% late-progress windows. Natural replay preserves coverage; task-balanced replay equalizes task sampling. Boundary replay favors demonstration phases near progress inferred from failed executions' partial-completion scores; late replay favors the final portion of demonstrations. The score-to-progress mapping is a sampling heuristic, not a measured failure time. Training retains recorded images and action targets, uses the RAP-WAM objectives with separate backbone and interface learning rates, and evaluates intermediate checkpoints. Appendix~\ref{app:agentic-protocol} specifies eligibility, fallback rules, optimizer settings, and budgets.

\paragraph{Retention without averaging policies.}
Let $m_t(\pi)=(\mathrm{SR}_t(\pi),\mathrm{Score}_t(\pi))$ denote success rate and mean partial-completion score from the same execution record. A task enters $\mathcal C_r$ if the candidate improves SR over its currently assigned policy, or ties SR and improves Score; a tie in both retains the incumbent. The assignment updates as
\begin{equation}
 g_{r+1}(t)=
 \begin{cases}
 \operatorname{id}(\pi_r^+),& t\in\mathcal C_r,\\
 g_r(t),& t\notin\mathcal C_r.
 \end{cases}
 \label{eq:task-retention}
\end{equation}
Here $\operatorname{id}(\pi)$ is a checkpoint identifier and $g_r(t)$ denotes the selected policy identity; the corresponding checkpoint remains in $\mathcal B_r$. An empty $\mathcal C_r$ leaves the active assignment unchanged. At inference, the task identity selects one whole checkpoint, with its architecture, normalizer, and inference settings. Parameters and within-episode action fragments are not averaged. This assignment operates across policies, whereas credit coefficients weight losses inside a policy's training.

The final assignment uses the base on 14 tasks and 16 other checkpoints on 28 tasks. History retains proposals, outcomes, and acceptance decisions for subsequent interventions. This is the recursive aspect of the experimental procedure; it does not establish autonomous improvement or a causal benefit from history reuse. Selection uses observed evaluation results, so independent policy-selection tests remain necessary. The complete state update and execution rule are retained in Appendix~\ref{app:post-training-details}.

\section{Experiments}
\label{sec:experiments}
\begin{figure}[t]
\centering\includegraphics[width=136mm]{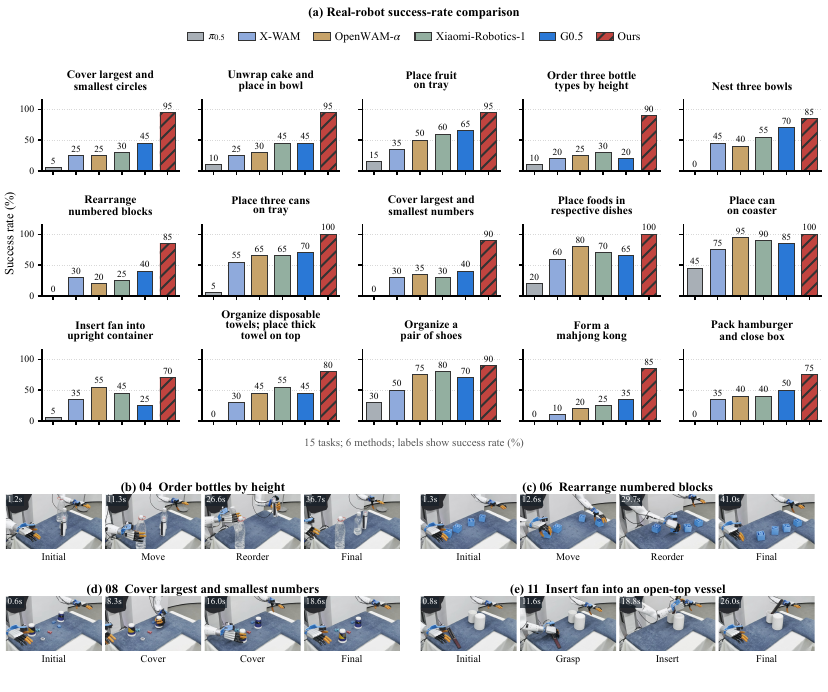}
\caption{\textbf{Real-robot evaluation and four recorded task sequences.} (a) Success rates across fifteen tasks and six methods; bar labels show percentages. (b--e) Tasks 04, 06, 08, and 11 show height-based bottle ordering, numbered-block rearrangement, covering numerical extrema, and fan insertion. Each task occupies four consecutive frames in a two-row, eight-column grid. Timestamps refer to the source video. Frames are taken directly from the original videos with torsos and incidental text masked; task numerals are retained. The phone-box assembly task is provided in Appendix~\ref{app:physical-pending}.}
\label{fig:real-robot-15x6}
\end{figure}

We evaluate the fixed RAP-WAM base and the post-trained task-conditioned system on RoboDojo, analyze their supervision signals, and report physical execution results.

\subsection{Base and post-trained system performance}
We report SR and partial-completion Score, averaged within each capability and then equally across the five capabilities \citep{chen2026robodojo}. Public results are taken from the RoboDojo leaderboard \citep{robodojo2026leaderboard}; coverage and system accounting are specified in Appendices~\ref{app:protocol} and~\ref{app:agentic-protocol}. Across the 42 tasks, the full method achieves \textbf{15.62\% average SR and 20.59 Score}, compared with 12.85\% and 17.86 for the RAP-WAM base. The gains are 2.77 percentage points in SR and 2.73 Score. Precision increases from 16.51\% to 21.03\%, Memory from 7.00\% to 9.33\%, and Generalization from 7.50\% to 10.50\%. These gains arise in the task-conditioned system, where retained specializations contribute across different capability groups.

Long-Horizon success rises from 32.50\% to \textbf{36.50\%}, with Score increasing from 45.12 to \textbf{48.43} (Table~\ref{tab:primary-leaderboard}). At the task level, tower building increases from 60\% to 72\%, pen-holder filling from 6\% to 18\%, and tic-tac-toe from 80\% to 86\%. The task-policy breakdown is given in Appendix~\ref{app:tasks}. Open success remains 0.75\%, and random-Generalization success remains zero, indicating that the observed gains do not resolve broad scene-shift robustness. The candidate comparison in Appendix Figure~\ref{fig:app-candidate-complementarity} helps explain this pattern: the base has the highest aggregate performance among individual checkpoints, yet candidates improve different task subsets. Task-specific retention preserves these local gains despite regressions elsewhere. Appendix Figure~\ref{fig:app-repertoire} shows how the retained specialists divide the task set; their value is not captured by ranking whole-policy averages alone.

\subsection{Action response changes the locations emphasized during training}

Figure~\ref{fig:allocation-fingerprint}a--b uses 315 RoboDojo windows across 35 conditions. Holding model predictions fixed, we separately remove the historical error prior, action response, or current-error contribution and renormalize the remaining weights to compare their effects on supervision allocation (Appendix~\ref{app:mapspace}). Response redistributes
spatial emphasis; the median shares of normalized coefficients redistributed
are 5.47\%, 2.22\%, and 1.14\% for
$\priorfield,\responsefield,0.5\errorfield$, respectively. The rehearsal prior
produces the largest redistribution. These distances combine magnitude and
spatial shape, not additive contributions to task success.

Figure~\ref{fig:allocation-fingerprint}c shows temporal supervision in three phone-packing demonstration windows. Spatially pooled action responses distinguish future action steps without an extra scoring network. Removing the response leaves all coefficients at one: $\priorfield$ is time-constant and this branch excludes $\errorfield$. Appendix~\ref{app:mapspace} details reconstruction and four-action grouping.

Figure~\ref{fig:allocation-fingerprint}d--e shows complementary rankings: on average, \emph{45.21\% of response-top20 positions lie outside the error-top20 set}. The condition-level rank distribution further characterizes this difference over the full field. These measurements explain the local diagnostic signal; the supplementary comparison of normalized auxiliary-credit shares is retained in Appendix~\ref{app:mapspace}.

\paragraph{Local sensitivity and relative credit.}
Under small isotropic action perturbations, $\mathbb E_\delta[\responsefield_i]\approx(\tau^2/C)\|J_i\|_F^2$, where $J_i$ is the video-output Jacobian with respect to action inputs and $C$ is the channel count (Appendix~\ref{app:response-proof}). Response describes coupling at the current condition and diffusion level; a single direction gives a noisy estimate. Per-sample normalization expresses relative priority, not absolute sensitivity across tasks. Current error and rehearsal history complement this local coupling, while the unit floor preserves supervision at low-response locations.

\subsection{Credit evolution within a task}

Figure~\ref{fig:light-credit-dynamics} follows one three-can tray-placement run. Panel a tracks spatial priors across training; b shows response locations across future groups at the final checkpoint; c summarizes component shares and removal sensitivity across eleven demonstration windows. The prior has the largest share and removal sensitivity, while response and error vary across windows. This distinguishes accumulated spatial difficulty from context-dependent reweighting: a final aggregate alone would hide variation over training and predicted time. The analysis characterizes fixed outputs, rather than retraining effects or object localization.

\subsection{Real-robot evaluation and qualitative sequences}
Figure~\ref{fig:real-robot-15x6} compares fifteen tasks, five baselines, and Ours, alongside four recorded task sequences. Event-aligned frames illustrate intermediate object states; twelve further sequences appear in Appendix~\ref{app:physical-pending}. The prediction-horizon analysis (Appendix Figure~\ref{fig:prediction-horizon}) separately evaluates all 50 illustrated training windows against a copy-condition baseline.

Ours achieves 89.00\% task-macro-average SR, compared with 51.33\% for the strongest baseline, G0.5, a margin of 37.67 percentage points. Supplementary phone-box assembly achieves 85\% versus 5\% for $\pi_{0.5}$; it is reported separately and excluded from the fifteen-task average (Appendix~\ref{app:physical-pending}).

\paragraph{Task variation and aggregation.}
Ours has the highest reported SR on each of the 15 physical tasks, but performance is not uniform. SR ranges from 70\% for fan insertion to 100\% for three-can placement, food placement, and coaster placement. Height-based bottle ordering reaches 90\%, compared with 30\% for the best baseline on that task. Each task contributes equally to the reported macro-average; it is not a pooled episode success rate. Trial counts and uncertainty estimates are not supplied with these percentages, so the comparisons describe the recorded outcomes rather than establish statistical significance. The separately reported phone-box assembly result does not change the common 15-task comparison set.

\subsection{Discussion}
Two complementary patterns emerge: response and error prioritize different prediction locations, whereas candidate policies specialize on different tasks. The former concerns within-policy supervision; the latter concerns task-level selection. Separating their contributions requires budget-matched training ablations and policy selection independent of final testing. Weak Open and random-Generalization performance motivates new-layout evaluation. Specialist retention additionally requires checkpoint storage and reliable task identities.

\section{Conclusion}
We presented RAP-WAM, a world action framework coupling future-supervised action representations with predictive rehearsal. Its action-response signal organizes both spatial visual and temporal action supervision, complemented by current and persistent prediction errors. The framework gives future prediction two connected roles: shaping representations for action generation and supplying feedback that guides subsequent learning. Simulation and physical evaluations demonstrate the capabilities of the resulting policies, while task-specific post-training preserves complementary specializations within the complete system. Together, these findings motivate a unified treatment of predictive representation, supervision allocation, and task-level capability retention for world action models.

\label{lastmainpage}
\clearpage
\label{nonmainstart}

\subsection*{AI use statement}
Generative AI assisted with research discussions, post-training intervention design, implementation and experiment scripts, literature retrieval, manuscript drafting and translation, and figure preparation. The improvement agents used in the experimental workflow and their role in proposing training interventions are described in Section~\ref{sec:agentic} and Appendix~\ref{app:agentic-protocol}; their outputs are part of the method under study. During manuscript preparation, assistance included checking citation records against primary sources, comparing reported values with saved experiment records, and compiling and inspecting the manuscript. All results reported in this paper have been manually reviewed by the authors. World-model predictions in Appendix~\ref{app:real-world-predictions} are identified as model outputs and paired with recorded targets. Reported quantitative results come from saved experiments or cited benchmark sources; pending measurements are not replaced with generated values. AI systems are not authors, and responsibility for the scientific claims, source attribution, code, and final submitted content remains with the authors.

\subsection*{Ethics statement}
This research does not involve human-subject experiments or user studies. We evaluate robot manipulation policies using robot demonstrations and simulation and real-robot execution records. Physical execution can cause unintended contact or motion; responsible deployment requires embodiment-specific checks of joint conventions and limits, an emergency stop, and supervised commissioning. Improved benchmark performance does not establish safety outside the evaluated conditions. Recorded demonstration frames and training-sample predictions are presented as qualitative evidence, not as additional autonomous-success trials. Display copies mask incidental identifying text, visible robot torsos in external-view sequences, and background faces in prediction comparisons; task-relevant numerals and mahjong symbols are retained. Original scene content is otherwise preserved, and the processing is disclosed in Appendices~\ref{app:physical-pending} and~\ref{app:real-world-predictions}. Masking alone does not establish consent or authorize redistribution: public release of demonstrations or identifiable recordings remains subject to the applicable dataset terms, participant permissions, and institutional requirements. No approval or consent status is inferred from the availability of these recordings. The limitations of task-conditioned selection, single-seed evaluation, and incomplete physical measurements are stated to avoid overstating generalization or autonomous capability.

\subsection*{Reproducibility statement}
We provide the source code in the supplementary materials.
Section~\ref{sec:method} defines the model interfaces, predictive rehearsal, response-based credit objective, and task-specific policy selection. Appendix~\ref{app:protocol} documents training and evaluation conventions, while Appendix~\ref{app:properties}, including Algorithm~\ref{alg:update}, provides implementation details, assumptions, and derivations. Appendix~\ref{app:agentic-protocol} specifies post-training sampling, optimizer settings, checkpoint retention, inference, and aggregation; Appendix~\ref{app:tasks} reports task-level outcomes. Appendices~\ref{app:mapspace} and~\ref{app:census} describe the diagnostic sampling and reconstruction procedures. The scope and processing of physical recordings and the selection of paired training predictions are described in Appendices~\ref{app:physical-pending} and~\ref{app:real-world-predictions}. The project records preserve source paths, hashes, selection manifests, and figure-generation scripts. These records support auditing the presented artifacts. Remaining reproducibility limits include the absence of independent policy-selection testing, multi-seed estimates, and completed physical-trial statistics.

\bibliography{references_verified_new,references_supplemental}
\bibliographystyle{rapwam}
\clearpage
\appendix
\section*{Appendix: Supplementary Material}
\raggedbottom
\setcounter{topnumber}{3}
\setcounter{bottomnumber}{2}
\setcounter{totalnumber}{5}
\section{Training and Evaluation Protocol}
\label{app:protocol}
Training was performed on NVIDIA A800 GPUs.

\method{} initializes its shared transformer from Wan2.2-TI2V-5B video
weights \citep{wan2025} and jointly learns future video and robot actions from target
RoboDojo demonstrations.
Action/state encoders, queries, and CAP are initialized for
target-task training, without initialization from a separately trained
robot policy. The base architecture contains no additional memory,
mixture-of-experts, or post-training extension.
Thus ``without embodied pretraining'' excludes an intermediate pretraining
stage on external embodied data, whether for action-policy learning or joint
world action modeling. It retains Wan video pretraining, pretrained text
embeddings, and joint world action training on target robot demonstrations.
The collection has
3,500 episodes, with 3,499 used for training and one for validation.
The fixed RAP-WAM base is the reference for local mechanism analyses. Table~\ref{tab:primary-leaderboard} reports the complete post-training system, whose retained task policies build on this substrate. The settings in Table~\ref{tab:contracts} describe the base learner, not the cumulative computation of the entire improvement process.

\begin{table}[ht]
 \centering\small
 \caption{Training settings. Physical training uses the same core with
 task-specific interfaces; physical success rates appear in Figure~\ref{fig:real-robot-15x6}.}
 \label{tab:contracts}
 \begin{tabularx}{\linewidth}{@{}lXX@{}}
 \toprule Setting & RoboDojo & Physical-robot data \\
 \midrule
 Video initialization & Wan2.2-TI2V-5B & Wan2.2-TI2V-5B \\
 Input & Head/two-wrist $384\times320$ mosaic & Head $224\times224$ \\
 Action/state dimension & 14/14 & Phone 28/28; numbers 32/32 \\
 Language & Precomputed instruction embeddings & None \\
 Observation/future/action length & 1/16/16 & 1/16/16 \\
 Queries/alignment width & 16/128 & 16/128 \\
 Per-device/global batch & 1/96 & 1/120 \\
 Optimizer/peak LR & AdamW/$2\times10^{-4}$ & AdamW/$2\times10^{-4}$ \\
 Precision/parameter EMA & BF16/none & BF16/none \\
 Auxiliary warmup (updates) & 18,000 & 6,000 \\
 Rehearsal probability/denoising & 0.25/four iterations & 0.25/four iterations \\
 Conditional state substitution & 0.5 & 0.5 \\
 \puga{} probability & 0.5 & 0.5 \\
 \puga{} video/action coefficients & 0.5/0.3 & 0.5/0.3 \\
 $\priorfield,\responsefield,\errorfield$ coefficients & 1/1/0.5 & 1/1/0.5 \\
 VAE & Cached recorded futures; predicted inputs encoded live & Live \\
 \bottomrule\end{tabularx}
\end{table}

RoboDojo policy evaluation uses 30 denoising steps and
executes eight actions before replanning. Action smoothing
uses a Savitzky--Golay filter with window 9 and polynomial order 2 on all 14 action
coordinates. Four-action boundary blending uses new-chunk coefficients
$(i+1)/5$ for $i=0,1,2,3$. This filter is distinct from classifier-free guidance.
Physical deployment additionally depends on robot-specific joint mappings,
units, and control rates.

Generalization has 12 tasks, Precision 8, Long-Horizon 8, Memory 6, and Open 8.
Binary success rate (SR) and partial-completion Score are averaged within
capabilities and then equally across the five capability means
\citep{chen2026robodojo}:
\begin{equation}
 \mathrm{AvgSR}=\frac15\sum_{d=1}^{5}\frac1{|\mathcal T_d|}
 \sum_{j\in\mathcal T_d}\mathrm{SR}_j.
 \label{eq:metric}
\end{equation}
Here $\mathcal T_d$ contains the evaluated tasks in capability $d$;
Score uses the same aggregation. Row-specific coverage is given in the
table notes and Appendix~\ref{app:tasks}.
Standard and random variants are combined into their parent Generalization task before aggregation. Both the base and agentic system cover all 42 task units: 12 Generalization, eight Precision, eight Long-Horizon, six Memory, and eight Open. For task $t$, SR is the number of successful valid episodes divided by its valid-episode count $n_t$, and Score is the mean partial-completion score over the same episodes. Task means are averaged within capability and then equally across capabilities; tasks are not reweighted by $n_t$. Layout identities and per-task denominators are retained in the result records.

The target alignment projection is stopped along with its target features.
The direct attention mask permits state-mediated information and
shared-parameter updates. At per-device batch one, the optional local min--max
difficulty correction vanishes. Spatial $\priorfield$ broadcasts uniformly in time,
so its normalized temporal contribution also vanishes. The prior is zero until its first valid rehearsal update. Rehearsal and response-based credit are invoked according to their respective sampling probabilities. Only successfully computed auxiliary losses with nonzero weights enter the training objective. For each branch, the effective participation rate is the fraction of training updates satisfying these conditions.

\section{Public References and Pretraining Attribution}
\label{app:wam}
G0.5 pretrains on robot demonstrations and web-scale vision--language data,
with 14 real/simulated embodiments mapped to a shared action space
\citep{liu2026g05}. G0.5 therefore uses external embodied pretraining.
Xiaomi-Robotics-1 and Hy use action-policy pretraining on handheld UMI
demonstrations \citep{xiaomi2026robotics1,zhang2026hyvla}; these are action-labeled
embodied trajectories, not necessarily robot-collected data. OpenWAM-$\alpha$
jointly pretrains future-video and action prediction on egocentric and robot
data \citep{openwam2026project}. Its final mixture contains 6,369 hours of
egocentric human and robot trajectories; human video is not itself labeled
robot-action supervision. Our no-external-embodied-pretraining flag still includes
Wan video initialization, pretrained text embeddings, and joint world action
training on target robot demonstrations. Table~\ref{tab:primary-leaderboard} therefore
defines ``Pretrained'' as an intermediate stage on external embodied data,
including robot/UMI action-policy and joint world action pretraining, rather
than asserting that every pretraining trajectory was robot-collected.

\paragraph{External embodied pretraining.}
The pretraining column summarizes the method's reported use of external embodied data. InternVLA-A1.5 uses robot-episode pretraining \citep{ma2026internvlaa15}; VLAct and X-VLA use cross-embodiment robot pretraining, and MolmoAct2 uses action-data pretraining \citep{yang2026vlact,zheng2025xvla,fang2026molmoact2}. Spatial Forcing inherits the Pi-05 base, so both carry a positive embodied-pretraining flag \citep{physicalintelligence2025pi05}.

X-WAM reports more than 5,800 hours of robot-data pretraining \citep{guo2026xwam}. GigaWorld-Policy introduces embodied-video pretraining using robot and egocentric human data before target-task action training \citep{ye2026gigaworldpolicy}. AHA-WAM reports no robot-data pretraining in its default simulation setting, but its real-world comparison pretrains both AHA-WAM and Fast-WAM on 24,600 RoboCOIN trajectories (approximately 165 hours) \citep{cai2026ahawam}. The AHA-WAM Yes flag refers to this reported real-world variant; it does not establish the initialization of the RoboDojo checkpoint. X-WAM and GigaWorld-Policy-0 are marked Yes based on their reported pretraining. Fast-WAM's reported simulation configuration uses Wan initialization without external embodied pretraining \citep{yuan2026fastwam}.

\paragraph{SimpleMemVLA training and comparison source.}
SimpleMemVLA uses timestamped visual history with a Qwen3.5-4B backbone and a flow-matching action head \citep{yin2026simplememvla}. Its released training recipe initializes the backbone from Qwen3.5-4B, constructs a new action head, and trains on each benchmark's demonstrations without a separate embodied-pretraining stage. We therefore mark No under the definition above. The RoboDojo values use the official 23 September 2026 leaderboard snapshot, including separate standard and randomized Generalization results.

\paragraph{World-model classification.}
Table~\ref{tab:primary-leaderboard} separates VLA and other action policies from world-model-based policies. The latter group includes OpenWAM-$\alpha$, X-WAM, GigaWorld-Policy-0, AHA-WAM, Fast-WAM and our policy, which learn an explicit predictive world-model component alongside action generation. InternVLA-A1.5 is grouped with VLAs: its foresight tokens receive latent supervision from a frozen pretrained video model, but the policy does not learn pixel-level generation and discards the video branch at inference \citep{ma2026internvlaa15}. Its $\ddagger$ marker records this supervision source without classifying the policy as a world action model.

Xiaomi, Hy, Pi-05, VLAct, X-VLA and MolmoAct2 document action-generation
objectives without an explicit predictive world-model component
\citep{xiaomi2026robotics1,zhang2026hyvla,physicalintelligence2025pi05,
yang2026vlact,zheng2025xvla,fang2026molmoact2}. Spatial Forcing aligns observed
geometry, not future scene evolution \citep{li2025spatialforcing}; its
benchmark identity is the Pi-05-based variant. Action flow matching, current
depth prediction, and textual subtask planning alone do not qualify as world
modeling here. KinRT uses kinematics-supervised expert routing for action generation and is grouped with non-world-model policies \citep{yang2026kinrt}. SimpleMemVLA is grouped with VLAs: its video context represents observed history, rather than an explicitly predicted future world state.
These method-level labels
neither make all architectures identical nor explain performance causally.

\section{Mathematical Properties and Objective Details}
\label{app:properties}
\begin{table}[!t]
\centering\small
\caption{Post-training system results across 42 tasks. SR is in percent; Score measures partial completion. Each task is averaged over its valid evaluation episodes.}\label{tab:primary-all-tasks}
\begin{tabular}{@{}p{5.0cm}lrr@{}}
\toprule Task & Capability & SR & Score \\
\midrule

Align blocks & Open & 0.00 & 0.00 \\
Arrange largest number & Generalization & 4.00 & 5.80 \\
Build tower & Precision & 72.00 & 79.60 \\
Classify objects & Long-Horizon & 2.00 & 6.50 \\
Classify objects by language & Open & 0.00 & 1.60 \\
Cover blocks & Memory & 22.00 & 26.10 \\
Deposit coin & Precision & 10.20 & 14.69 \\
Fasten screws & Precision & 8.00 & 27.60 \\
Fill egg holder & Long-Horizon & 0.00 & 12.95 \\
Fill pen holder & Long-Horizon & 18.00 & 43.40 \\
Fold clothes & Generalization & 12.00 & 16.00 \\
General pickup & Open & 6.00 & 6.00 \\
Hang mugs & Generalization & 2.00 & 7.30 \\
Imitate sorting sequence & Memory & 2.00 & 3.80 \\
Insert key & Precision & 0.00 & 15.00 \\
Insert tubes & Precision & 58.00 & 72.40 \\
Make kong & Long-Horizon & 82.00 & 82.00 \\
Make toast & Generalization & 10.00 & 18.00 \\
Match and pick from conveyor & Memory & 32.00 & 32.00 \\
Organize table & Long-Horizon & 10.00 & 48.50 \\
Pack objects into box & Generalization & 6.00 & 17.20 \\
Pick from conveyor by image & Open & 0.00 & 0.00 \\
Play xylophone & Precision & 2.00 & 2.00 \\
Play stacking toy & Long-Horizon & 0.00 & 2.00 \\
Play tic tac toe & Long-Horizon & 86.00 & 96.00 \\
Plug in charger & Precision & 4.00 & 4.00 \\
Pour balls into vase & Precision & 14.00 & 14.00 \\
Pour by language & Open & 0.00 & 0.00 \\
Pour liquid into cup & Generalization & 4.00 & 4.00 \\
Press by number & Memory & 0.00 & 0.00 \\
Push t & Generalization & 0.00 & 0.00 \\
Put bottles into dustbin & Long-Horizon & 94.00 & 96.10 \\
Solve equation & Open & 0.00 & 0.00 \\
Sort nesting dolls by size & Generalization & 8.00 & 8.00 \\
Stack blocks & Generalization & 24.00 & 28.20 \\
Stack blocks by language & Open & 0.00 & 1.20 \\
Stack bowls & Generalization & 42.00 & 43.20 \\
Store laptop and headphones & Generalization & 14.00 & 24.80 \\
Store tools in toolbox & Open & 0.00 & 0.50 \\
Swap t & Memory & 0.00 & 0.00 \\
Swap blocks & Memory & 0.00 & 0.00 \\
Sweep blocks & Generalization & 0.00 & 0.00 %
\\
\bottomrule
\end{tabular}
\end{table}
\begin{table}[!t]
\centering\small
\caption{RAP-WAM base results across 42 tasks, using the single step-742500 checkpoint before post-training. Task order and metric definitions match Table~\ref{tab:primary-all-tasks}. SR (\%) and Score are computed over valid evaluation episodes.}\label{tab:base-all-tasks}
\begin{tabular}{@{}p{5.0cm}lrr@{}}
\toprule Task & Capability & SR & Score \\
\midrule
Align blocks & Open & 0.00 & 0.00 \\
Arrange largest number & Generalization & 2.00 & 3.80 \\
Build tower & Precision & 60.00 & 69.80 \\
Classify objects & Long-Horizon & 0.00 & 4.40 \\
Classify objects by language & Open & 0.00 & 1.60 \\
Cover blocks & Memory & 16.00 & 20.60 \\
Deposit coin & Precision & 4.08 & 7.76 \\
Fasten screws & Precision & 2.00 & 18.60 \\
Fill egg holder & Long-Horizon & 0.00 & 12.95 \\
Fill pen holder & Long-Horizon & 6.00 & 32.00 \\
Fold clothes & Generalization & 6.00 & 9.20 \\
General pickup & Open & 6.00 & 6.00 \\
Hang mugs & Generalization & 0.00 & 3.80 \\
Imitate sorting sequence & Memory & 0.00 & 1.20 \\
Insert key & Precision & 0.00 & 14.40 \\
Insert tubes & Precision & 54.00 & 69.20 \\
Make kong & Long-Horizon & 74.00 & 74.00 \\
Make toast & Generalization & 4.00 & 14.00 \\
Match and pick from conveyor & Memory & 26.00 & 26.00 \\
Organize table & Long-Horizon & 10.00 & 48.50 \\
Pack objects into box & Generalization & 2.00 & 14.70 \\
Pick from conveyor by image & Open & 0.00 & 0.00 \\
Play xylophone & Precision & 0.00 & 0.00 \\
Play stacking toy & Long-Horizon & 0.00 & 0.60 \\
Play tic tac toe & Long-Horizon & 80.00 & 94.50 \\
Plug in charger & Precision & 0.00 & 0.00 \\
Pour balls into vase & Precision & 12.00 & 12.00 \\
Pour by language & Open & 0.00 & 0.00 \\
Pour liquid into cup & Generalization & 0.00 & 0.00 \\
Press by number & Memory & 0.00 & 0.00 \\
Push t & Generalization & 0.00 & 0.00 \\
Put bottles into dustbin & Long-Horizon & 90.00 & 94.00 \\
Solve equation & Open & 0.00 & 0.00 \\
Sort nesting dolls by size & Generalization & 2.00 & 2.00 \\
Stack blocks & Generalization & 24.00 & 27.00 \\
Stack blocks by language & Open & 0.00 & 0.40 \\
Stack bowls & Generalization & 40.00 & 42.10 \\
Store laptop and headphones & Generalization & 10.00 & 17.60 \\
Store tools in toolbox & Open & 0.00 & 0.50 \\
Swap t & Memory & 0.00 & 0.00 \\
Swap blocks & Memory & 0.00 & 0.00 \\
Sweep blocks & Generalization & 0.00 & 0.00 %
\\
\bottomrule
\end{tabular}
\end{table}

\subsection{Predictive representation and training update}
\label{app:substrate}
\paragraph{Joint flow-matching substrate.}
We use the linear interpolation and velocity-regression formulation of flow matching \citep{lipman2023flow}.

A frozen VAE encodes observed and subsequent demonstration video, with fixed camera
mosaics for multi-view observations. Actions and states use separate
training-set normalizers. Video and actions share a noise level $\sigma$
but receive independent noises:
\begin{equation}
 z_\sigma=(1-\sigma)z+\sigma\epsilon_z,\quad
 a_\sigma=(1-\sigma)a+\sigma\epsilon_a,\quad
 (v_z^*,v_a^*)=(\epsilon_z-z,\epsilon_a-a).
 \label{eq:joint-noise}
\end{equation}
Observed visual latents are clamped and excluded from the future-only video
loss. The video-initialized trunk is trained together with newly initialized
action/state encoders, queries, and CAP on target demonstrations.

\paragraph{Tokenization, conditioning, and action projection.}
Action and state encoders use Linear--SiLU--Linear projections, learned
position embeddings, and LayerNorm. All four token groups share each block's
self-attention projections, text cross-attention, and feed-forward layers;
there are no modality-specific transformer experts. Rotary position encoding
applies only to video-token queries and keys. Predictive queries use action-side
diffusion-time modulation. State tokens receive no additional diffusion-time
modulation, but still undergo the block's learned modulation and all updates.
The mask in Equation~\ref{eq:directional-attention} is applied identically
at training and sampling: for $i\in\mathcal A,j\in\mathcal V$, the logit is
suppressed; state-mediated paths remain.

The two projections in Equation~\ref{eq:query-align} have output width 128.
The entire target branch, including $P_t$, is evaluated without gradients.
Future latents are pooled over time and onto a spatial $4\times4$ grid;
the resulting targets supervise 16 distinct query slots. They are neither
an online memory bank nor an independently trained teacher.
In Equation~\ref{eq:additive-readout}, the MLP output projection and bias are
zero-initialized; the time-modulated branch starts from the newly initialized
sequence head. Its exact form is
\begin{equation}
 H_r(h,e)=W_r[\operatorname{LN}(h)\odot(1+e+m_s)+(e+m_b)]+b_r,
 \label{eq:modulated-readout}
\end{equation}
with learned offsets $m_s,m_b$. The two branches have separate normalization
operators. The construction introduces no low-pass filter, task routing, or
independently supervised coarse/fine decomposition.

\paragraph{Generated-image and state conditions.}
Rehearsal is invoked with $p_c(k)=\bar p_c r_c(k)$, where
$r_c(k)=\min(1,k/W_c)$. Conditional state substitution has probability
$\bar p_s\min(1,k/W_s)$. In the reported configuration, $\bar p_c=0.25$
and $\bar p_s=0.5$. The first chunk uses $S_c$ gradient-free denoising steps,
with $S_c=4$, before RGB decoding. Its last
image replaces the next observed image; this prefix and the recorded next
future frames are jointly re-encoded by the frozen VAE. Thus the recorded
RGB targets are retained, but their temporal VAE latents need not equal those
encoded with the recorded initial image. For action and state normalizers $N_a,N_s$,
the substituted normalized state is
\begin{equation}
 \widetilde s_1=N_s\!\left((1-b_s)s_1^{\rm raw,*}
 +b_s\operatorname{sg}\!\left[N_a^{-1}(\widehat a_{0,H_a-1})\right]\right).
 \label{eq:state-rehearsal}
\end{equation}
Here $\widehat a_0$ is the sampled normalized action chunk and $b_s$ selects
state substitution. This mapping assumes compatible absolute-joint action
and state schemas. It does not model actuator dynamics or contact feedback.
The base loss includes query alignment; rehearsal and credit auxiliaries
supervise video/action flow outputs without additional alignment losses.

The persistent prior is initially zero, so credit remains defined before a
rehearsal update. The first valid rehearsal-error map initializes the prior
directly; subsequent maps use the moving average. The reference predictions use the
original demonstration batch with fresh noising, not the rehearsed input.
Each credit invocation uses one gradient-carrying reference prediction and one gradient-free
perturbed forward, in addition to the base and any rehearsal computations.

\begin{algorithm}[ht]
\caption{One training update of \method}
\label{alg:update}
\begin{algorithmic}[1]
\Require Adjacent demonstration chunks, model $\theta$, spatial prior $\priorfield$ (initially zero), step $k$
\State Sample base noising; compute joint loss $\mathcal L\gets\mathcal L_0$
\If{$b_c\sim\mathrm{Bernoulli}(p_c(k))$ is one}
  \State Generate the first chunk without gradients using $S_c$ denoising steps
  \State Form the second-chunk image from the last predicted frame
  \State Optionally substitute the last raw predicted joint action; normalize as state
  \State Compute video/action flow loss $\mathcal L_1$ against the recorded future/actions
  \State Update $\priorfield$ from detached second-chunk spatial errors
  \State $\mathcal L\gets\mathcal L+\lambda_c r_c(k)\mathcal L_1$
\EndIf
\If{$b_P\sim\mathrm{Bernoulli}(p_P)$ is one}
  \State Fresh auxiliary noising; compute gradient-carrying $(v^z,v^a)$
  \State Evaluate the action perturbation without gradients; detach $\priorfield,\responsefield,\errorfield$
  \State Compute $\weightfield,c,w$ and normalize their masses to $q^v,q^a$
  \State $\mathcal L\gets\mathcal L+\widetilde{\mathcal L}_P(\theta;\bar\theta)$
\EndIf
\State Backpropagate $\mathcal L$ and update the trainable parameters
\end{algorithmic}
\end{algorithm}

The algorithm assumes valid training inputs, with a zero prior before its
first valid rehearsal update. It specifies the update rule, not the empirical
frequency with which each auxiliary branch contributes during training
(Appendix~\ref{app:protocol}). It neither sends actions to an environment nor
requests new corrective demonstration labels at predicted states.

\paragraph{Joint sampling and deployment.}
Inference begins with independent future-video and action noise. At adjacent
noise levels, Euler integration updates both modalities:
\begin{equation}
 (z_{n+1},a_{n+1})=(z_n,a_n)
 +(\sigma_{n+1}-\sigma_n)(v_\theta^z,v_\theta^a).
 \label{eq:joint-sampling}
\end{equation}
The observed visual prefix is clamped at every iteration. Queries, directional
attention, and CAP remain in each forward; the action output
is $N_a^{-1}(a_N)$. Returning actions does not decode future RGB, but it does
retain future-latent computation. Alignment targets, rehearsal, prior updates,
and perturbation-based credit construction are absent at deployment.

\subsection{Local interpretation of the perturbation response}
\label{app:response-proof}
Fix $\theta$, observation, state, language, video noise, and diffusion level.
Let $f_i:\mathbb R^d\to\mathbb R^C$ map the flattened normalized noisy
action chunk to the predicted video velocity at location $i$.
For $g\sim\mathcal N(0,I_d)$ and $\delta=\tau g$,
$\responsefield_i=C^{-1}\|f_i(a+\tau g)-f_i(a)\|^2$.
At the noisy action input $a_\sigma$, the local interpretation is
\begin{equation}
 \mathbb E_\delta[\responsefield_i]
 =\frac{\tau^2}{C}\|J_a f_i(a_\sigma)\|_F^2+O(\tau^4).
 \label{eq:response-interpretation}
\end{equation}
This perturbation argument \citep{bishop1995noise} motivates a forward-only sensitivity probe without explicit Jacobians or second-order derivatives, under the regularity conditions stated below.

\paragraph{Proposition 1 (response interpretation).}
Suppose $f_i$ is three times continuously differentiable on $\mathbb R^d$
and its third derivative has globally bounded operator norm. At fixed $a$,
\begin{equation}
 \mathbb E\responsefield_i=\frac{\tau^2}{C}\|J_a f_i(a)\|_F^2+O(\tau^4)
 \quad\text{as }\tau\to0.
\end{equation}
For an affine $f_i$, the expression without the remainder is exact for every
$\tau$. This is a standard local sensitivity calculation, not a new control
guarantee or an assertion that a finite-precision model satisfies the assumptions.

\paragraph{Proof.}
Write the Taylor expansion as
\begin{equation}
 f_i(a+\tau g)-f_i(a)=A+B+R,\quad
 A=\tau Jg,\quad B=\tfrac12\tau^2 H[g,g],\quad
 \|R\|\leq M_3\tau^3\|g\|^3/6.
\end{equation}
Here $M_3$ bounds the third derivative, $M_1=\|J\|_{\rm op}$, and
$M_2=\|H\|_{\rm op}$ at $a$. Gaussian second moments give
$\mathbb E\|A\|^2=\tau^2\|J\|_F^2$, and symmetry makes the cubic
linear--quadratic cross term vanish. With $m_p=\mathbb E\|g\|^p$, the remaining
terms satisfy
\begin{align}
 \left|\mathbb E\responsefield_i-\frac{\tau^2}{C}\|J\|_F^2\right|
 &\leq\frac1C\left[
  \tau^4\left(\frac{M_2^2}{4}+\frac{M_1M_3}{3}\right)m_4\right.\nonumber\\
 &\hspace{2.5em}\left.+\tau^5\frac{M_2M_3}{6}m_5
        +\tau^6\frac{M_3^2}{36}m_6\right].
\end{align}
All Gaussian moments are finite, proving the asymptotic bound. Affine maps
have $H=R=0$. The constants can depend on dimension and the fixed predictor.
\hfill$\square$

\paragraph{Single-direction and scale limitations.}
For an affine map,
\begin{equation}
 \operatorname{Var}(\responsefield_i)=\frac{2\tau^4}{C^2}
       \operatorname{tr}[(J^\top J)^2].
\end{equation}
Thus shrinking $\tau$ does not eliminate the relative uncertainty of a single
direction. The implementation uses one perturbation, and generally
$\mathbb E\mathcal N(\responsefield)\ne\mathcal N(\mathbb E\responsefield)$; its normalized map is
not an exact Jacobian-norm ranking. Nor does per-coordinate $\tau=0.1$ imply
a small whole-chunk perturbation: the RMS norm is $\tau\sqrt d$, approximately
1.50 for $d=16\times14$ and 2.12 for $d=16\times28$ in normalized coordinates.
The analysis motivates perturbation-scale and repeatability checks, not an
environment-causality interpretation or a prediction of task success.

\subsection{Response and prediction error need not agree}
\label{app:complementarity}
\begin{figure}[!htb]
\centering\includegraphics[width=\linewidth]{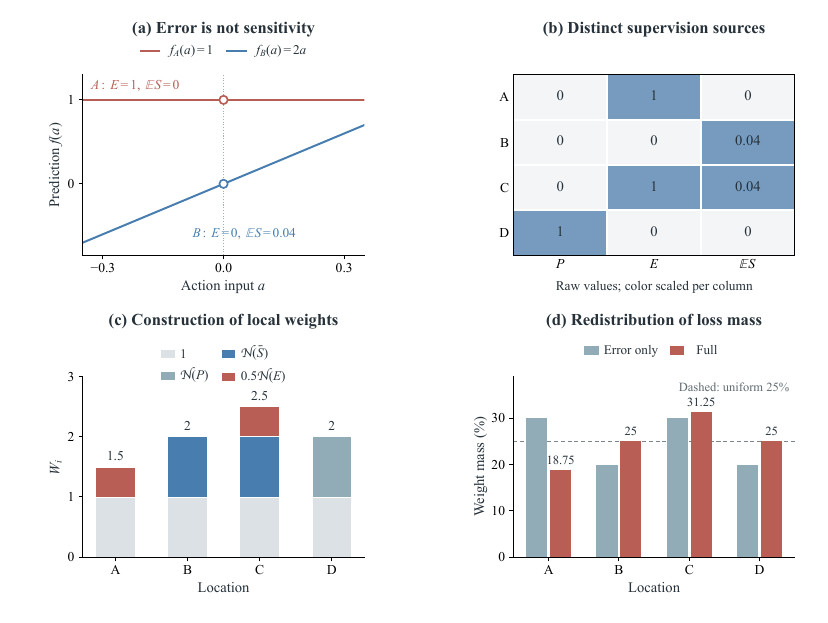}
\caption{\textbf{Why error, response, and rehearsal prior provide different supervision.} An exact constructed example, not measured robot results. At four locations, let $f_A(a)=1$, $f_B(a)=2a$, $f_C(a)=1+2a$, and $f_D(a)=0$, with $a=0$, target zero, and prior $P=(0,0,0,1)$. For $\delta\sim\mathcal N(0,0.1^2)$, use the expected response $\bar S=\mathbb E_\delta S$. (a) A has error without action sensitivity; B has sensitivity without error. (b) The four locations separate residual, response, combined, and historical signals. (c) Applying the spatial rule to $P,\bar S,E$ gives positive weights $(1.5,2,2.5,2)$. (d) After normalization, error-only allocation gives $(30,20,30,20)\%$, whereas the full rule (Full) gives $(18.75,25,31.25,25)\%$. Thus current error alone cannot reproduce the allocation. This illustration uses expected response; training uses a single sampled perturbation, as discussed in Appendix~\ref{app:response-proof}.}
\label{fig:app-matched-probe}
\end{figure}

Figure~\ref{fig:app-matched-probe} illustrates how prediction error, action response, and historical error yield different supervision weights.

For $f_1(a)=b$ and $f_2(a)=\kappa a$, suitable targets give
\begin{equation}
 \errorfield_1=(b-y_1)^2>0,\quad \errorfield_2=0,\qquad
 \mathbb E\responsefield_1=0,\quad
 \mathbb E\responsefield_2=\tau^2\kappa^2.
 \label{eq:response-residual-example}
\end{equation}
A constant action-independent output can have nonzero error
but zero response; an accurately predicted linear output can have zero error
but nonzero expected response. This demonstrates non-equivalence of signals,
not an empirical improvement in robot success. The finite-perturbation and
normalization qualifications above still apply.

\subsection{Frozen-weight surrogate and redistribution identity}
\label{app:surrogate-proof}
For the main visual objective, bars denote uniform empirical averages and
$\operatorname{Cov}_u$ the corresponding covariance over active locations:
\begin{equation}
 \sum_i q_i^v e_i^z=\bar e^z+
 \frac{\operatorname{Cov}_u(\weightfield,e^z)}{\bar{\weightfield}},
 \label{eq:weight-covariance}
\end{equation}
This separates prediction error from emphasis on particular locations. The
general form and its gradient analogue follow below.
Let $i$ enumerate all active batch/spatiotemporal locations, with channel-mean
errors $e_i(\theta)$ and positive weights $\weightfield_i$. At current $\bar\theta$,
freeze $q_i=\weightfield_i/\sum_j\weightfield_j$. The local surrogate and its gradient are
\begin{equation}
 \widetilde F(\theta;\bar\theta)=\sum_iq_i(\bar\theta)e_i(\theta),\qquad
 g_q=\sum_iq_i(\bar\theta)\nabla e_i(\bar\theta).
\end{equation}
This is the precise backward-pass meaning of stop-gradient. At differentiable
points, the fully adaptive value $F(\theta)=\sum_iq_i(\theta)e_i(\theta)$ has
the additional derivative $\sum_i e_i\nabla q_i$, which the implementation
intentionally omits. Ties in min--max normalization do not prevent defining the
frozen-weight surrogate.

\paragraph{Proposition 2 (redistribution identities).}
Let bars and $\operatorname{Cov}_u$ use uniform empirical averaging over $n$
locations, with covariance denominator $n$, and let
$g_i=\nabla e_i(\bar\theta)$. Then
\begin{equation}
 \sum_iq_i e_i=\bar e+\frac{\operatorname{Cov}_u(\weightfield,e)}{\bar{\weightfield}},
 \qquad
 g_q-\bar g=\frac{\operatorname{Cov}_u(\weightfield,g)}{\bar{\weightfield}}.
\end{equation}
The second covariance is defined componentwise. These are explanatory
identities, not original theorems of optimization or closed-loop improvement.

\paragraph{Proof.}
For either scalar $x_i=e_i$ or vector $x_i=g_i$, expanding the empirical
covariance gives $n^{-1}\sum_i\weightfield_i x_i=\bar{\weightfield}\bar x+
\operatorname{Cov}_u(\weightfield,x)$. Divide by $\bar{\weightfield}>0$ and use the definition of $q$.
\hfill$\square$

Consequently, normalized weighting can change gradient direction and/or magnitude
when priorities correlate with location-wise gradients. It need not decrease
the fully adaptive value $F$, and it is not a corrected importance sampler for
the uniform risk. Within a single sample, $\errorfield_i=e_i$, so its spatial/temporal
empirical covariance satisfies
$\operatorname{Cov}_{u,s}(\mathcal N(\errorfield),e)=
\operatorname{Var}_{u,s}(e)/\max(\max_i e_i-\min_i e_i,\epsilon_{\rm n})\geq0$:
part of \puga{} explicitly increases coefficients at high-error locations. Whether adding response $\responsefield$
or persistent prior $\priorfield$ improves control is an empirical question.

\paragraph{Bounded emphasis without exclusion.}
Let $\kappa_v=1+\alpha_P+\alpha_S+\alpha_E$ and $\kappa_a=1+\alpha_S$.
For $N_v$ video positions and $N_a$ action positions in their respective
normalizations, nonnegative source coefficients imply
\begin{equation}
 \frac1{\kappa_v N_v}\le q_i^v\le\frac{\kappa_v}{N_v},\qquad
 \frac1{\kappa_a N_a}\le q_j^a\le\frac{\kappa_a}{N_a}.
 \label{eq:credit-support}
\end{equation}
Our settings give $\kappa_v=3.5$ and $\kappa_a=2$.
Thus selectivity retains every location. Detachment removes direct gradients
through credit while allowing it to adapt across updates. These bounds
regulate emphasis, not gradient norms, unbiased estimation, or preservation
of previously acquired capabilities.

\subsection{Limits of the interpretation}
\label{app:decisive}
The identities characterize the weighting rule, not its causal effect on
success. No matched-budget, independently evaluated component ablation is
reported here. A claim of signal-specific control benefit would require
matching auxiliary computation, initialization, data, and checkpoint selection,
and separating weighting information from gradient-scale changes. Normalizing
weight sums does not match the full AdamW update. Numerical checks on
affine and smooth toy maps verify the identities, not robot performance.

\section{Task-level Base and Post-training Results}
\label{app:tasks}
Tables~\ref{tab:primary-all-tasks} and~\ref{tab:base-all-tasks} report the post-training system and the fixed base on the same 42 tasks. The base achieves 12.85\% SR and 17.86 Score under the capability-balanced aggregation. Base and system use identical task coverage and aggregation rules; per-task SR and Score come from one policy's valid evaluation records. Generalization variants are combined within their parent task.

\begin{figure}[!htb]
\centering\includegraphics[width=\linewidth]{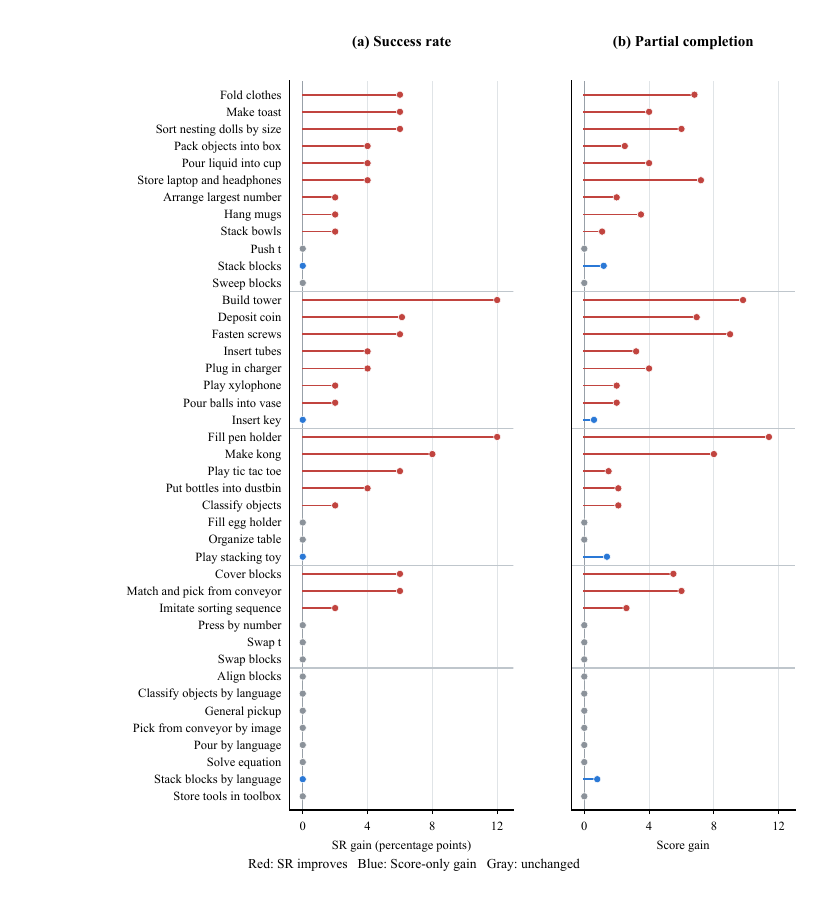}
\caption{\textbf{Task-level changes under the selected assignment.} Aligned rows show all 42 tasks, grouped by capability: (a) SR changes and (b) partial-completion Score changes relative to the base. Red indicates 24 tasks with higher SR, blue four tasks with higher Score only, and gray 14 unchanged tasks. These are observed results under the selected task-policy assignment, not an independent test of regression-free improvement.}
\label{fig:app-task-gains}
\end{figure}

Figure~\ref{fig:app-task-gains} compares the per-task changes in SR and Score relative to the base.

\section{Supervision Allocation and Response Analysis}
\label{app:mapspace}
The analyses use fixed-policy predictions to quantify supervision allocation. Simulation statistics use the RAP-WAM base and 35 conditions of nine demonstration windows each, separate from benchmark evaluation episodes. Physical-demonstration sources are specified in Appendix~\ref{app:real-maps}.

\paragraph{Field notation.}
$\priorfield,\responsefield,\errorfield,\weightfield$ denote the cross-chunk prior,
action response, prediction error, and composite visual-loss coefficient, respectively.
The three simulation component fields used in the analyses below are normalized.

\paragraph{Arithmetic source-removal analysis.}
Figure~\ref{fig:allocation-fingerprint}a--b uses the same simulation
component maps, with model outputs held fixed.
Each of 315 windows has $N=4\times24\times20=1{,}920$ native positions,
giving 604,800 positions in total, not that many independent observations.
Cast the normalized $\priorfield,\responsefield,\errorfield$ arrays to float64 and reconstruct
$\weightfield=1+\priorfield+\responsefield+0.5\errorfield$ instead of using the rounded stored composite.
For each $X\in\{\priorfield,\responsefield,0.5\errorfield\}$, remove that source
within the original window and recompute the sum-normalization:
\begin{equation}
 q_i^{-X}=\frac{\weightfield_i-X_i}{\sum_j(\weightfield_j-X_j)},\qquad
 \mathrm{TV}_X=\frac12\sum_i\left|q_i-q_i^{-X}\right|.
 \label{eq:allocation-removal}
\end{equation}
The unit floor and all other normalized components remain fixed.
All 315 windows share the same saved $\priorfield$ map; its removal is not removal of
a separate learned task-specific prior.

For panel a, average tied ranks separately within each window and divide each
rank by 1,920. The horizontal coordinate is the $\errorfield$ rank and the vertical
coordinate is the $\responsefield$ rank. Hexagonal binning with grid size 18 reports the
arithmetic mean of $\Delta_i=1{,}920(q_i-q_i^{-\responsefield})$ over positions in each
displayed hexagon. Color measures signed change relative to the uniform
per-position coefficient $1/N$, not a ratio to $q_i^{-\responsefield}$ or a density count.
The diverging scale is symmetric about zero and covers all displayed means;
no smoothing is applied. The native changes sum to zero in every window;
recovering this property from bin means requires weighting by bin counts,
not an unweighted mean of the displayed hexagons. Pooling preserves equal
window representation overall, but the conditional mean in a hexagon weights
conditions by their number of positions falling there.

For panel b, $\mathrm{TV}_X$ equals the sum of positive coefficient changes, the sum of negative
changes in magnitude, and the largest absolute change in total coefficients
assigned to any subset of native positions. Writing
$\alpha_X=\sum_iX_i/\sum_i\weightfield_i$ and $p_i^X=X_i/\sum_jX_j$
for $\sum_iX_i>0$ gives
\begin{equation}
 q-q^{-X}=\alpha_X(p^X-q^{-X}),\qquad
 \mathrm{TV}_X=\alpha_X\,\mathrm{TV}(p^X,q^{-X}).
 \label{eq:removal-interpretation}
\end{equation}
Thus removal sensitivity combines the source's summed coefficients with its
normalized spatial shape. It is neither a causal importance estimate nor an
additive decomposition across sources. If $\sum_iX_i=0$, removal instead
leaves the normalized coefficients unchanged and $\mathrm{TV}_X=0$.
Compute TV on each window, take its
median across nine windows per condition, and show all 35 condition medians.
The cross-condition median [Q1,Q3] in percent is 5.4707 [5.2891,5.7790] for
$\priorfield$, 2.2159 [1.9935,2.5100] for $\responsefield$, and 1.1405 [1.0422,1.2434] for $0.5\errorfield$.
Each density is normalized to its own peak. Density smoothing is display-only;
the marks are descriptive IQRs, not
confidence intervals. These arithmetic removals hold model outputs fixed,
do not retrain a policy, and measure neither gradient magnitude nor success rate.

Panel c instead uses raw float32 $\responsefield$ from one phone-packing
validation demonstration, at window starts 63, 126, and 188. These windows
use the physical-image diagnostic model and data described in
Appendix~\ref{app:census}. Spatial means of raw $\responsefield$ are normalized over the four native
future bins using the implemented $10^{-6}$ denominator floor. Their temporal
ranges are respectively $3.6025\times10^{-6}$, $5.6887\times10^{-6}$, and
$1.8406\times10^{-6}$, so the floor is inactive in all three retained windows.
The display shows four bin weights, each repeated over four consecutive action
steps, alongside the exact unit reference when $\responsefield$ is removed. It does not
use a normalized-map temporal proxy or 16 independently probed actions.
Because $\priorfield$ is time-constant and this branch excludes $\errorfield$,
removing $\responsefield$ leaves unit temporal coefficients. The same action
perturbation thus provides spatial and temporal selectivity without a second
coefficient predictor. The displayed $w$ precedes the objective's normalization
$q_j^a=w_j/\sum_k w_k$; even its unit reference is not a retrained ablation.
The companion reproducibility materials retain the source correspondence
and display specification.

\paragraph{Diagnostic conditions.}
Table~\ref{tab:mapspace-condition-ids} identifies the 35 conditions used for the response and error analyses in Figure~\ref{fig:allocation-fingerprint}. Each condition contains nine demonstration windows.

\begin{table}[!t]
\centering\fontsize{7.5}{9}\selectfont
\caption{Task identities for the 35 diagnostic conditions.}
\label{tab:mapspace-condition-ids}
\begin{tabular}{@{}rl@{}}
\toprule Index & Task \\
\midrule

00 & Arrange largest number \\
01 & Build tower \\
02 & Classify objects \\
03 & Cover blocks \\
04 & Deposit coin \\
05 & Spell RoboDojo \\
06 & Fasten screws \\
07 & Fill egg holder \\
08 & Fill pen holder \\
09 & Fold clothes \\
10 & Hang mugs \\
11 & Imitate sorting sequence \\
12 & Insert key \\
13 & Insert tubes \\
14 & Make kong \\
15 & Make toast \\
16 & Match + pick from conveyor \\
17 & Organize table \\
18 & Pack objects into box \\
19 & Play xylophone \\
20 & Play stacking toy \\
21 & Play tic-tac-toe \\
22 & Plug in charger \\
23 & Pour balls into vase \\
24 & Pour liquid into cup \\
25 & Press by number \\
26 & Push T \\
27 & Put bottles in dustbin \\
28 & Sort nesting dolls by size \\
29 & Stack blocks \\
30 & Stack bowls \\
31 & Store laptop + headphones \\
32 & Swap T pieces \\
33 & Swap blocks \\
34 & Sweep blocks \\
\bottomrule
\end{tabular}
\end{table}

\paragraph{Priority-set flow and condition rainclouds.}
In each simulation window, top-20\% membership is defined on all 1,920 native
positions, giving 384 effective members in either map. Boundary ties receive
fractional membership. For memberships $a_i,b_i\in[0,1]$ in $\errorfield,\responsefield$, the four
joint masses are the position means of $(1-a_i)(1-b_i)$, $(1-a_i)b_i$,
$a_i(1-b_i)$ and $a_ib_i$, followed by an equal mean over all 315 windows.
These fractions are 0.709583, 0.090417, 0.090417 and 0.109583, respectively.
Figure~\ref{fig:allocation-fingerprint}d conditions on response-top20 membership:
dividing the shared and response-only masses by 0.2 gives the displayed
54.7916\% and 45.2084\% bars. Each percentage uses the response top set as its denominator. Shared mass remains 10.9583\% of all positions.
The overlap compares signals within the same window.

Raincloud points use the within-condition median of each window's
Spearman rank correlation. The cloud density is normalized to its own peak.
The cloud and median/IQR marks
describe 35 curated condition medians, not 315 independent evaluations.
The separately summarized overlap median 0.559896 differs from the alluvial mean 0.547916.
Distribution smoothing in the cloud is a display device, not an uncertainty
model; no confidence interval, significance test, or success gain is inferred.

Figure~\ref{fig:allocation-fingerprint}c labels the three original phone-packing windows (starts 63, 126, 188) as W1, W2, W3. The five-panel arrangement retains the original underlying statistics.

\paragraph{Normalized auxiliary weight share.}
Appendix~\ref{app:mapspace} evaluates $\sum_i b_i\weightfield_i/\sum_i\weightfield_i$
on the same windows under uniform $\weightfield_i=1$, post-hoc error-only
$\weightfield_i=1+0.5\errorfield_i$, and full \puga{} $\weightfield_i=1+\priorfield_i+\responsefield_i+0.5\errorfield_i$ rules.
The saved component maps are normalized; full weights are reconstructed from
them rather than taken from float16-rounded stored $\weightfield$. Each point is a median
over nine windows in one condition, with median/IQR marks across 35 conditions.
The respective medians are 20.00\%, 20.9448\% and 24.4003\%; the error-only
and \puga{} IQRs are 20.7050--21.2332\% and 23.0325--25.3654\%.
Every condition-median share increases; the median paired difference is 3.7816
percentage points (range 0.8839--5.7004), not the difference of group medians.
Adding both $\priorfield$ and
$\responsefield$ prevents attribution to either signal alone; enrichment partly follows
from including $\responsefield$ in both the weights and the selected set. This measures
normalized auxiliary coefficients, not gradient magnitude, task gain, or a retrained
error-only control.

\begin{figure}[!htb]
\centering\includegraphics[width=\linewidth]{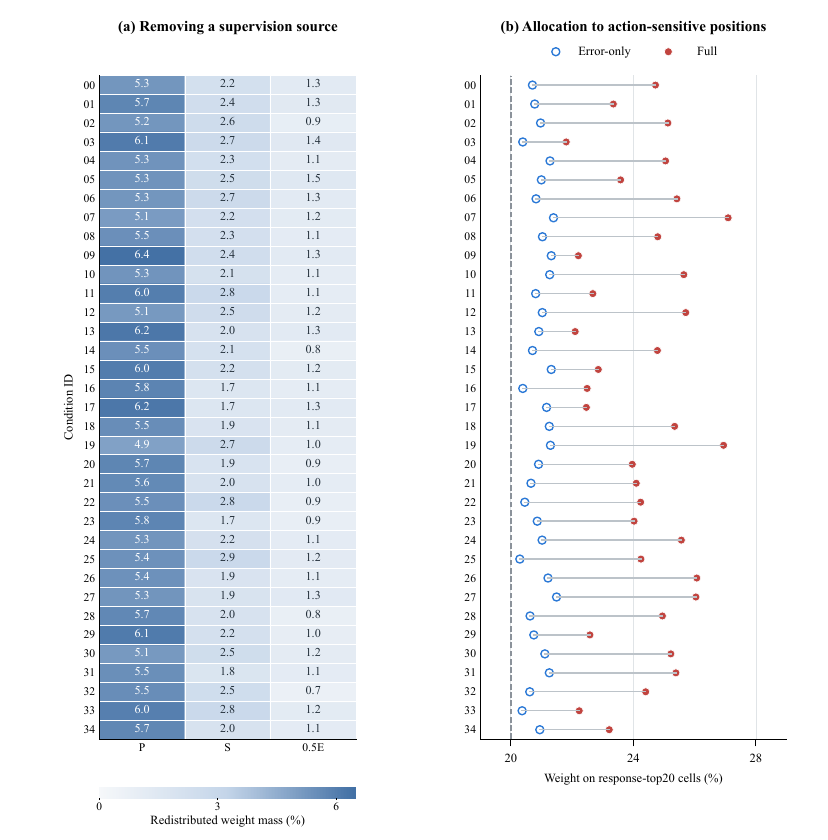}
\caption{\textbf{Condition-resolved supervision allocation.} Both panels retain all 35 conditions, each with nine windows; IDs follow Table~\ref{tab:mapspace-condition-ids}. (a) Median total-variation distance after removing each source from the normalized spatial weights, in percent. (b) Median normalized weight mass assigned to response-top20 positions under error-only reweighting (blue) and the full credit rule (red, labeled Full); each connector pairs the same condition. The dashed line is uniform allocation, 20\%. The fields and predictions are fixed throughout.}
\label{fig:app-condition-credit}
\end{figure}
Figure~\ref{fig:app-condition-credit} shows that the full credit rule assigns a larger weight share to response-top20 positions than error-only reweighting in every displayed condition. The medians across conditions are 24.40\% and 20.94\%, respectively. This supports the mechanism's ability to alter supervision beyond error-only weighting; it does not measure a retrained success-rate gain. The source-removal distances also vary across conditions, showing that the same coefficient rule induces different allocations for different inputs. The 315 windows are descriptive samples, not independent policy evaluations; no confidence intervals or significance claims are inferred from these plots.

\section{Spatiotemporal Diagnostic Protocol}
\label{app:census}\label{app:real-maps}
The simulation analysis uses 35 conditions with nine distinct demonstrations each. Conditions are defined using initial image, state, and logarithmic trajectory length; each demonstration contributes one middle valid window and one seeded draw of diffusion noise and action perturbation. Each map contains four future bins on a $24\times20$ grid. Within-window ranks use all 1,920 positions; condition summaries aggregate nine windows. Pixels and windows from the same condition are not independent policy evaluations.

The temporal-credit example in Figure~\ref{fig:allocation-fingerprint}c uses a phone-packing demonstration with window starts 63, 126, and 188. Its raw response maps have four $14\times14$ bins; spatial averaging and the normalization in Equation~\ref{eq:temporal-weight} produce the displayed action weights. Figure~\ref{fig:light-credit-dynamics} instead uses the three-can tray-placement run, with checkpoint and window details below. Both analyses use recorded demonstrations, not autonomous physical success trials.

\paragraph{Three-timescale display and component accounting.}
Panel a of Figure~\ref{fig:light-credit-dynamics} reads the native $14\times14$ EMA prior directly from the same can-placement run at steps 690, 4140, 8280, 12420, 16560, and 20700. Percentages are relative to the final step. The upper row selects steps 690, 12420 and 20700, independently normalized to show spatial shape. The lower row reports the arithmetic mean of the raw stored EMA at all six checkpoints on a logarithmic axis. The curve measures the training prior rather than held-out prediction error. The final prior is the one used in panels b--c. Panels b--c use the verified step-20,700 can-placement checkpoint and recorded training demonstration 8574282; these are not physical executions. Panel b retains start 112 (RGB offsets 114, 118, 122, 126), selected after screening 32 archived windows. Its square crop is $x\in[48,176)$, $y\in[96,224)$; top40 membership and transitions are computed over the full $14\times14$ grid before cropping. Thus the crop can show fewer than 40 selected cells. The archived full-field tensors retain background responses.

Panel c instead includes all eleven exported windows, starts 0,16,$\ldots$,160, without localization-based selection. For each window let $C_j\in\{\priorfield,\responsefield,0.5\errorfield\}$ and $\weightfield=1+\sum_j C_j$. The upper row reports $100\sum C_j/\sum(\weightfield-1)$, with sums over all four future groups and all 196 spatial positions. The unit floor is excluded only from this composition ratio. The lower row normalizes $\weightfield$ and $\weightfield-C_j$ separately over the same positions and reports $100\,\mathrm{TV}$, where $\mathrm{TV}(p,q)=\tfrac12\sum|p-q|$. The unit floor is retained in both distributions; no remaining component is recomputed or renormalized before this final distribution normalization. This measures sensitivity of the allocation rule at fixed outputs, not the effect of training without a component. Shared row scales are 0--100\% and 0--5\%; lower-row points denote available windows and lines connect adjacent observations. Upper-row stacked shares are linearly connected between adjacent windows.

Component share measures weight magnitude, whereas removal TV measures its effect on the normalized allocation. These complementary statistics describe one demonstration at fixed model parameters.

\section{Scope of Physical-Robot Evidence}
\label{app:physical-pending}
\begin{figure}[!htb]
\centering\includegraphics[width=\linewidth]{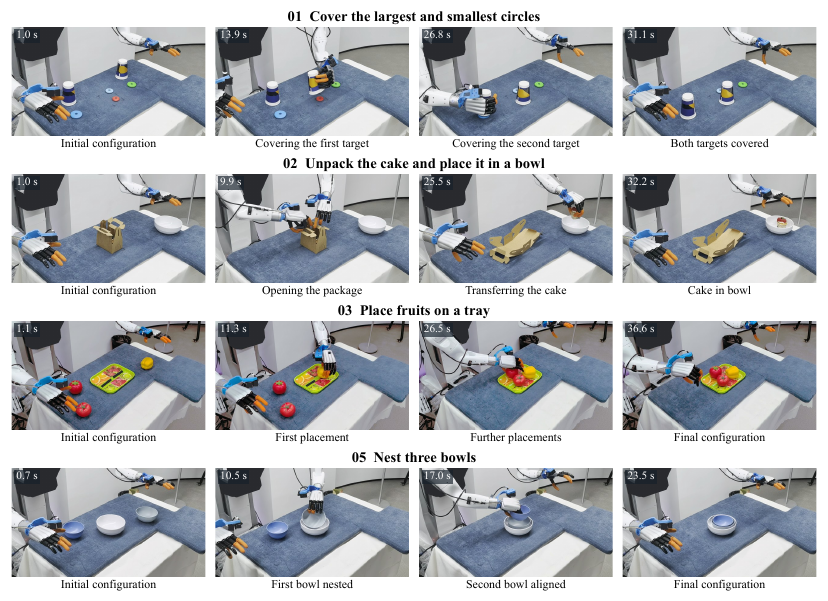}
\caption{\textbf{Additional recorded task sequences (I).} Tasks 01, 02, 03, and 05 show circle-size selection, cake unpacking and placement, fruit placement, and bowl nesting. Frames are event-aligned within each source video; torsos and incidental text are masked before display downsampling and compression.}
\label{fig:physical-catalog-a}
\end{figure}

\begin{figure}[!htb]
\centering\includegraphics[width=\linewidth]{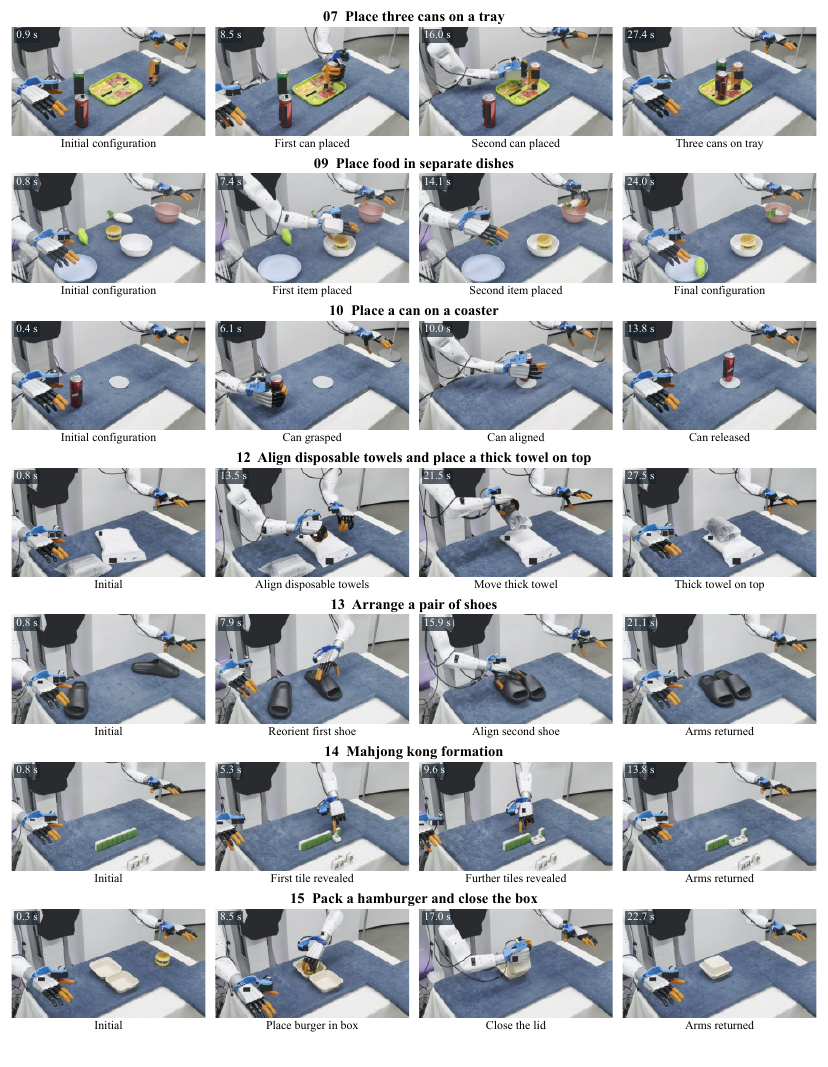}
\caption{\textbf{Additional recorded task sequences (II).} Tasks 07, 09, and 10 show can placement, food placement, and coaster placement. Task 12 aligns disposable towels and places a thick towel on top; the torso and packaging text are masked; the recorded scene is otherwise retained. Tasks 13 and 14 show shoe organization and mahjong kong formation. Task 15 places a hamburger in a box and closes its lid, ending with both arms returned. Timestamps and frame-selection conventions follow Figure~\ref{fig:physical-catalog-a}.}
\label{fig:physical-catalog-b}
\end{figure}

\begin{figure}[!htb]
\centering\includegraphics[width=\linewidth]{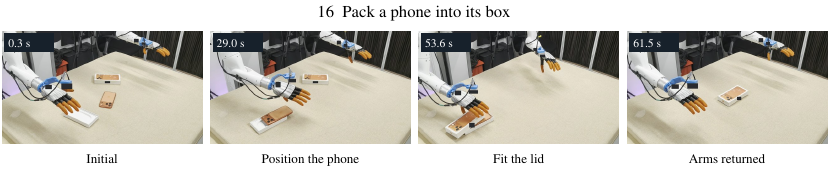}
\caption{\textbf{Additional recorded task sequence (III).} Task 16 packs a phone into its box and fits the lid. The four original-video frames show the initial configuration, phone positioning, lid placement, and the final configuration with both arms returned. Torso and incidental packaging and equipment text are masked; timestamps refer to the source video.}
\label{fig:physical-catalog-c}
\end{figure}

Sixteen task videos provide recorded manipulation sequences. Tasks 04, 06, 08, and 11 appear in Figure~\ref{fig:real-robot-15x6}; Figures~\ref{fig:physical-catalog-a}--\ref{fig:physical-catalog-c} cover the twelve other recorded tasks. Each row contains four chronologically ordered frames from one video: initial configuration, task-specific intermediate events, and a visible terminal configuration. Selection follows actions rather than equal time intervals; timestamps are source-video times. Task 15 places a hamburger in a takeaway box and closes the lid; its final frame shows both arms returned. Task 16 packs a phone into its box and fits the lid. This supplementary task is excluded from the fifteen-task main comparison.

All 64 frames are decoded directly from the sixteen original videos at the reported timestamps. Tight opaque masks cover incidental packaging and equipment text, including chest branding; additional masks cover the visible torso while preserving foreground arms and hands. Task-defining numerals and mahjong symbols, timestamps, and stage labels are retained. Original cloth marks, object details, and robot poses are retained. Display frames are downsampled and JPEG-compressed; no generative editing or inpainting is applied. Original videos and unmasked extracted frames are retained.

\paragraph{Supplementary phone packing.} For packing a phone into its box and closing the lid, the supplied results report SRs of 5\% for $\pi_{0.5}$ and 85\% for Ours.

\section{Post-training Implementation and Evaluation}
\label{app:agentic-protocol}
\paragraph{Agent implementation and experimental workflow.}
Completed task results and failure evidence are supplied to an improvement agent based on Claude Opus 5 or GPT 5.6 Sol. The agent returns a post-training proposal specifying the parent policy, target behaviors, training-data allocation, and parameter-update scope. Training produces a candidate for task-level evaluation. Improved task capabilities enter the repertoire; unsuccessful candidates are excluded from active execution. Evaluation outcomes feed the next proposal, linking policy learning to repeated experimental decisions.

\paragraph{Evaluation object.}
The full system combines task-specialized policies developed on the RAP-WAM substrate. Its task-to-policy assignment is part of the evaluated system specification. SR and Score for a task refer to the same policy execution record; parameters and within-episode action fragments are not averaged across policies. The numerical source manifest preserves task identities, policy identities, and original summary paths.

\paragraph{Aggregation.}
The complete system has Generalization SR/Score of 10.50/14.375, Precision 21.0255/28.6617, Long-Horizon 36.5000/48.4318, Memory 9.3333/10.3167, and Open 0.75/1.1625. Applying Equation~\ref{eq:metric} gives 15.6218\% SR and 20.5895 Score, rounded to 15.62\% and 20.59 in the main text. Table~\ref{tab:primary-leaderboard} shows Generalization's standard and random columns separately, reconstructed from the corresponding task-policy records. The local base comparison uses identical capability denominators.

\paragraph{Inference configuration.}
Evaluation uses 30 denoising steps, replanning every eight actions, nine-step Savitzky--Golay smoothing, and four-action boundary blending. Each task assignment identifies the policy architecture, checkpoint, normalizer, and inference path. 

\paragraph{Component comparisons.}
Separating policy learning from agentic decision-making requires matched data, training compute, interaction counts, and retry budgets. The relevant comparisons are continued training without agent-guided intervention, refinement with history reset, and the complete feedback loop. Removing response-based credit tests the local learning rule; resetting experiment history tests the role of cross-round evidence. The reported system comparison evaluates the combined procedure.

\paragraph{Post-training implementation.}
In the continuation experiment with demonstration replay, we start from base step 742500, updating all action-generating parameters with AdamW and cosine decay. Learning-rate warmup lasts 2,000 updates; rehearsal, state-substitution, and credit schedules restart over 5,000 updates. Its immutable demonstration sampler uses 50\% natural replay, 20\% task-balanced replay, 15\% reliable failure-boundary windows, and 15\% late-progress windows. The source covers 35 training tasks; failure-boundary evidence is available for 21. Five virtual epochs correspond to ten source-dataset equivalents, with a final training step of 145,585. The final repertoire uses intermediate checkpoints at 12,500, 25,000, and 50,000 steps; the last training checkpoint is not automatically the retained policy for every task.

\paragraph{Task-level improvement.}
The final assignment retains the base for 14 of 42 tasks and uses 16 specialist checkpoints for the other 28, drawing on 30 recorded candidate identities. Nine specialists are obtained within 5,000 additional optimization steps, indicating that useful task-specific refinements can emerge from short training runs. Success improves on 24 tasks; four other substitutions improve Score at unchanged SR. The capability-balanced SR gain is 2.7697 percentage points, from 12.8520\% to 15.6218\%. These checkpoint budgets describe the retained solutions; total development cost also includes longer training branches and candidate evaluation.

\begin{figure}[!htb]
\centering\includegraphics[width=\linewidth]{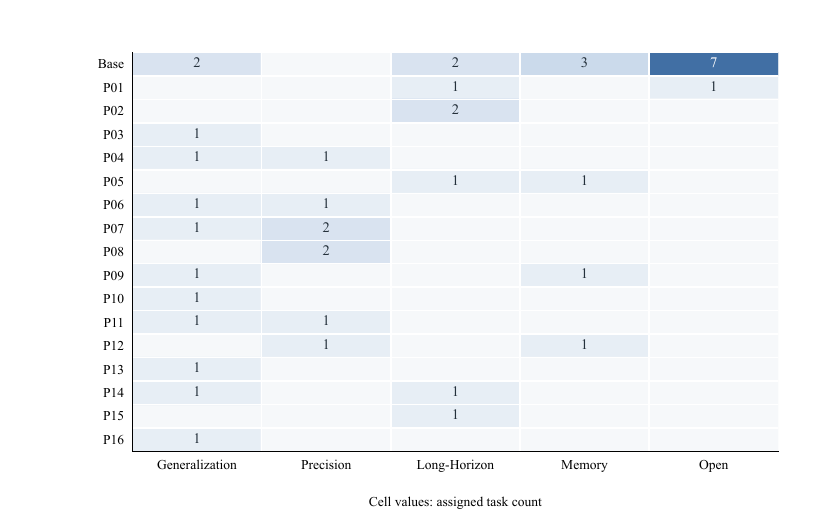}
\caption{\textbf{Task assignments in the retained repertoire.} Each cell counts tasks assigned to the base or a retained specialist within one capability group; blank cells denote zero assignments. P01--P16 identify existing checkpoints, with identities recorded in the accompanying manifest. This is the final repertoire snapshot, not an iteration history or a policy-removal experiment.}
\label{fig:app-repertoire}
\end{figure}
The base serves 14 tasks, while 16 other checkpoint identities serve 28. Different specialists serve different capability groups. Assignments use the observed evaluation results, as specified in Appendix~\ref{app:agentic-protocol}.

\paragraph{Evaluation scope.}
The RAP-WAM base and post-trained repertoire use no external embodied pretraining. Task-policy selection uses observed evaluation results; the reported evidence does not include an independent held-out policy-selection test. Runs also differ in RNG and live-VAE metadata, so they are not paired stochastic replicates. This RoboDojo comparison evaluates integrated simulation performance. Real-robot results are reported separately in Figure~\ref{fig:real-robot-15x6}.

\paragraph{Replay-intervention parameter values.}
The sampler uses $\tau_{\rm rel}=0.15$, $b_{\min}=0.08$, $b_{\max}=0.22$, $\kappa=0.5$, $\tau_{\rm boundary}=0.30$, $K_{\rm fallback}=128$, and $\tau_{\rm late}=0.65$. The replay proportions are $(0.50,0.20,0.15,0.15)$ in the order used in Section~\ref{sec:agentic}, giving $2N$ virtual windows. Learning rates are $\eta_{\rm backbone}=10^{-5}$ and $\eta_{\rm interface}=3\times10^{-5}$. AdamW uses betas $(0.9,0.95)$ and weight decay $10^{-4}$ for decayed parameter groups, with zero decay for the excluded groups; warmup lasts 2,000 optimizer steps, followed by cosine decay. Candidate checkpoints are saved every 2,500 steps. These values instantiate the replay intervention used in the experiments.

\paragraph{Execution evidence to a training distribution.}
For each task, the execution profile records the fraction of failed episodes with positive partial-completion scores, together with a center $c_t$ and quartiles $q_{25,t},q_{75,t}$. Boundary replay is enabled for eligible tasks whose positive-failure fraction reaches a reliability threshold $\tau_{\rm rel}$. A demonstration window at offset $k$ has normalized progress $p=k/\max(1,L-H)$, with $p=1$ if no offset range is available; $L$ is episode length and $H$ the required sequence span. Set $b_t=\operatorname{clip}(b_{\min}+\kappa(q_{75,t}-q_{25,t}),b_{\min},b_{\max})$ and proximity $w=\exp[-(p-c_t)^2/(2b_t^2)]$. The boundary pool contains windows with $w\geq\tau_{\rm boundary}$; an empty pool falls back to the $K_{\rm fallback}$ highest-proximity windows, capped by the available count. The late-progress pool uses $p\geq\tau_{\rm late}$. This transfers a task-level score profile to demonstration progress as a sampling heuristic, rather than estimating a physical failure time.

\paragraph{Continued training and candidate generation.}
The training distribution combines natural replay, task-balanced replay, boundary replay, and late-progress replay with nonnegative proportions $\alpha_{\rm natural},\alpha_{\rm balanced},\alpha_{\rm boundary},\alpha_{\rm late}$ summing to one. A virtual epoch preserves one complete pass over the source windows and adds the other pools in these proportions. Additional samples are balanced across eligible tasks and drawn uniformly within each task pool, with replacement only when the requested count exceeds pool size. Images and action labels remain unchanged. The selected parent is then optimized using the existing RAP-WAM objectives. The backbone and action interface use separate learning rates $\eta_{\rm backbone}$ and $\eta_{\rm interface}$; the latter covers the action head, action/state encoders, query tokens, and projections. The optimizer runs within a specified update budget, saving candidates at a prescribed interval for evaluation. The replay-intervention parameter values above specify the thresholds, mixture proportions, optimizer configuration, and checkpoint schedule.

\paragraph{Per-task episode accounting.}
All reported aggregates use the same 42-task definition. Each task uses its valid evaluated layouts; the evaluation records contain 49 for Deposit coin and 44 for Fill egg holder, shared by the base and candidate policies. The source summaries and per-layout records were cross-checked; duplicate records were counted once. Candidate selection uses observed SR first and Score to break ties, with both metrics taken from one policy. Deposit coin uses P07 at 12,500 post-training steps (5/49 successes, Score 14.6939); Fill egg holder retains the base (0/44, Score 12.9545). Metrics are computed directly over these valid episodes. The unified base aggregate is 12.8520\% SR / 17.8602 Score, and the system aggregate is 15.6218\% / 20.5895, a gain of 2.7697 percentage points and 2.7293 Score.

\begin{figure}[!htb]
\centering\includegraphics[width=\linewidth]{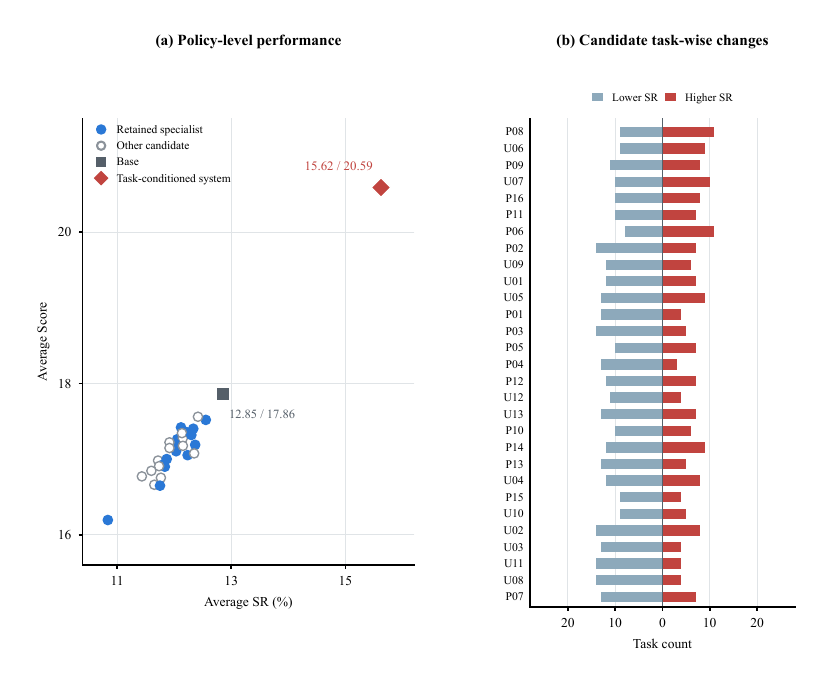}
\caption{\textbf{Why retain task specialists?} (a) All 30 candidate checkpoint identities, evaluated using the same 42-task aggregation. Filled blue circles are retained specialists, open circles other candidates, the square is the base, and the diamond is the task-conditioned system. Axis origins are truncated to display the observed range. (b) For each of the 29 non-base candidates, tasks with lower SR (left) or higher SR (right) relative to the base; unchanged tasks are omitted from the bars. Candidates are ordered by average SR, with Score breaking ties. P identifiers match Figure~\ref{fig:app-repertoire}; U identifiers denote unused candidates.}
\label{fig:app-candidate-complementarity}
\end{figure}
The base has the highest aggregate SR and Score among the individual checkpoints in this candidate pool. Nevertheless, refinements improve subsets of tasks while regressing on others. The retained assignment combines these local gains and reaches 15.62\% SR and 20.59 Score. This distinguishes a task-specialized improvement from a universal improvement of one checkpoint. All candidate results are shown, including candidates excluded from the repertoire. The comparison uses the observed results involved in policy selection; the system point is reconstructed from its task assignments, rather than an independent validation or evidence that the same gains persist under new seeds. Inference metadata differences are described in Appendix~\ref{app:agentic-protocol}.

\subsection{Post-training and policy-retention details}
\label{app:post-training-details}
Building on execution-feedback approaches \citep{xiao2026enpire,lu2026aspire}, the outer improvement loop uses experiment results to guide continued training, retain useful task specializations, and combine them through task-conditioned execution. Its state $X_r=(\mathcal B_r,g_r,\mathcal H_r)$ contains the policy repertoire, task assignment, and experiment history; the repertoire initially contains RAP-WAM. Training creates candidates, while evaluation and retention determine which capabilities remain available.

\paragraph{From execution evidence to continued training.}
At round $r$, completed experimental results are supplied to a language-model-based improvement agent. Its input $\mathcal F_r$ contains task outcomes, execution trajectories, and available model diagnostics; the agent uses these results to design the next post-training procedure. The agent compares weak task behavior with previous interventions in $\mathcal H_r$ to formulate a hypothesis about a learning bottleneck. It then specifies $u_r$: the parent policy, target tasks or execution phases, the data or training-code change, and the trainable parameter set. This is a decision about the learning procedure; the policy optimizer carries out the resulting update:
The intervention and optimization are defined by Equation~\ref{eq:agent-intervention}.
Agent-guided refinement yields multiple task specializations with only a few thousand fine-tuning steps, reusing existing policies and demonstrations. The operator $\mathcal U_{\rm RAP}$ updates designated parameters from a parent checkpoint while preserving that policy. The replay intervention below instantiates this process; implementation details and workflow provenance appear in Appendix~\ref{app:agentic-protocol}.

\paragraph{Evidence-guided continued training.}
The intervention mixes ordinary, task-balanced, failure-boundary, and late-progress demonstrations: $q_r=\sum_{j\in\{\mathrm{natural,balanced,boundary,late}\}}\alpha_jq_j$, with $\alpha_j\geq0$ and $\sum_j\alpha_j=1$. Boundary windows are selected by proximity to a score-derived task profile; reliability thresholds and fallback rules handle weak evidence or empty pools. Ordinary replay maintains demonstration coverage. All pools retain recorded observations and action targets. Training optimizes the RAP-WAM objectives from the parent checkpoint, with intermediate checkpoints evaluated as candidates. Appendix~\ref{app:agentic-protocol} specifies window selection, optimizer settings, and budgets. The score-to-progress mapping is a sampling heuristic, not a measurement of physical failure time.

\paragraph{Retention and task-conditioned composition.}
Task-wise comparison with the current assignment $g_r$ produces a result vector $\mathcal E_r$ of success, partial completion, and observed failures. Let $\mathcal C_r$ contain the tasks on which the candidate is accepted. The comparison retains a task-level change vector $\Delta_t=(\Delta\mathrm{SR}_t,\Delta\mathrm{Score}_t)$ relative to the incumbent on each task, so improvements and regressions are not hidden by an aggregate mean. Both metrics must come from the same candidate execution record. Their assignments switch to the candidate; other tasks retain their policies. A candidate with an accepted improvement enters the repertoire; if $\mathcal C_r=\varnothing$, it is discarded from the active policy set and $g_r$ is unchanged:
\begingroup
\small
\begin{equation}
\mathcal B_{r+1}=\operatorname{Retain}(\mathcal B_r,\pi_r^+,\mathcal C_r),\quad g_{r+1}=\operatorname{Assign}(\mathcal B_{r+1},\mathcal C_r,g_r),\quad \mathcal H_{r+1}=\mathcal H_r\cup\{(\mathcal F_r,u_r,\mathcal E_r,\mathcal C_r)\}.
\label{eq:recursive-state}
\end{equation}
\endgroup
The base and previously useful policies remain available. For task $t$ and observation history $o_{\leq k}$, the system executes
\begin{equation}
 a_k\sim\pi_{g_r(t)}(\,\cdot\mid o_{\leq k},t).
\label{eq:system-policy}
\end{equation}

The outer assignment selects a complete learned policy; credit coefficients govern supervision within policy training. Task-level comparison determines retention, with evaluation accounting specified in Appendix~\ref{app:agentic-protocol}.

\paragraph{Closing the improvement loop.}
History retains interventions, task-wise outcomes, and acceptance decisions, including rejected candidates. Later rounds can revise unsuccessful hypotheses and continue from useful specialists. Each evaluation is supplied back to the agent to design the next post-training round. The resulting loop is evaluate, propose, train, re-evaluate, and retain or discard. Discarding a policy does not imply discarding the experimental evidence used for the next decision.

\section{Additional Related Work}
\label{app:extended-related}
\paragraph{Generalist representations and action interfaces.}
OpenVLA adapts vision--language representations to robot actions
\citep{kim2024openvla}. G0.5 places reasoning and action in one autoregressive
stream \citep{liu2026g05}, while Xiaomi-Robotics-1 scales action-labeled
pretraining \citep{xiaomi2026robotics1} and Hy-Embodied uses modality-specific
computation with a flow-based action expert \citep{zhang2026hyvla}.
Spatial Forcing supplies geometric alignment targets \citep{li2025spatialforcing}.
Our model learns its action interface on the target data. Its loss-coefficient rule
uses the joint model's own outputs and existing future targets, rather than
requiring an additional pretrained robot expert or geometric-label pipeline.

\paragraph{Training with predicted inputs and adaptive supervision.}
Self Forcing trains video diffusion on self-generated context with sequence-level supervision \citep{huang2025selfforcing}. Our rehearsal instead supervises an adjacent recorded segment and retains its demonstration targets.

Scheduled sampling and interactive imitation learning address discrepancies
between recorded demonstration inputs and histories produced by policy execution
\citep{bengio2015scheduled,ross2011dagger}. Our adjacent-segment training replaces
observed inputs with predictions while retaining recorded demonstration targets; it neither collects
new interactions nor inherits interactive data-aggregation guarantees.
Prioritized replay and focal loss emphasize selected errors
\citep{schaul2016prioritized,lin2017focal}; importance sampling targets estimator
variance \citep{katharopoulos2018importance}, and value-aware model learning
uses decision structure \citep{farahmand2017valueaware}.
\puga\ combines persistent error with action response, and projects the same
response into spatial and temporal supervision. RoboDojo and RoboTwin offer
complementary manipulation benchmarks \citep{chen2026robodojo,chen2025robotwin};
we report RoboDojo and real-robot task outcomes, alongside prediction diagnostics.

\paragraph{Synthetic trajectories and joint world action pretraining.}
UWM couples video and action diffusion with independent modality timesteps \citep{zhu2025uwm}; DreamZero jointly models video and actions to transfer across tasks and embodiments \citep{ye2026dreamzero}. Motus learns latent actions from optical flow \citep{bi2025motus}, and MotuBrain extends joint modeling to multiview and cross-embodiment settings \citep{motubrain2026}. DreamGen generates synthetic task videos and recovers pseudo-actions using latent-action or inverse-dynamics models \citep{jang2025dreamgen}. Our rehearsal instead changes the conditioning context while retaining recorded demonstration targets; response-based credit then redistributes supervision within that training. These approaches provide architectural and data-generation context, not controlled evidence for the effect of our credit rule.

\section{Recorded and Predicted Futures in Real-Robot Training}
\label{app:real-world-predictions}
Figures~\ref{fig:wm-circle}--\ref{fig:wm-phone} show one reproducible training example for 15 of the 16 real-robot tasks with external-view video sequences in Appendix~\ref{app:physical-pending}. Task 16 uses the completed Tencent Cloud phone-training run; its configuration, checkpoint, and sample identity were independently verified. Each example uses the final checkpoint of its audited task-training run. We select the median-length eligible continuous training segment and its middle legal two-chunk window, before inspecting predictions or attention. The archived exports used previously are replaced by these identified samples so that predictions and attention come from the same forward trajectory. Recorded targets belong to the same training sample as the prediction; the separately filmed external-view execution need not be the same episode. Each model receives only the conditioning observation and recorded current state at inference. There are no weight updates. The first column repeats the condition; offsets $t+1,t+4,t+8,t+12,t+16$ index the sampled future sequence, not seconds.

\paragraph{Prediction metrics.}
Every predicted frame is labeled with RGB SSIM \citep{wang2004ssim}; figure-level SSIM and LPIPS \citep{zhang2018perceptual} are arithmetic means over the five displayed future frames, excluding the condition. SSIM uses an $11\times11$ Gaussian window with $\sigma=1.5$, population covariance, and channel averaging, excluding a five-pixel image border. LPIPS uses the official AlexNet network, version 0.1, at native resolution with inputs mapped to $[-1,1]$. Both metrics use decoded float images clipped to $[0,1]$, before privacy masks, quantization, or JPEG display compression. The orange dashed baseline (``Repeat current frame'') uses the current conditioning image itself as the prediction at every future position, without running the world model. For current image $I_t$ and recorded future $I_{t+h}$, its plotted error is $\operatorname{LPIPS}(I_t,I_{t+h})$. Thus it measures how much the recorded scene departs from remaining visually unchanged. Higher SSIM and lower LPIPS indicate closer images. We also retain metrics for all 16 predicted frames in the accompanying data.

\paragraph{Attention and local error.}
The bottom strip shows the condition with a fixed task-region rectangle, predictive-query attention, mean absolute RGB prediction error over the five displayed future frames, and the horizon-wise LPIPS curve. Rectangles were marked from condition images and checked for visible task coverage, independently of attention magnitude; they are coarse task regions, not object segmentations, and do not track moving objects. Region SSIM and MAE use the same rectangle in each future frame. A read-only probe records the actual self-attention query/key tensors after normalization and rotary encoding, retaining the model's additive mask and full softmax denominator. We average all 24 heads and 16 predictive-query rows at layers 10, 20, and 30, and denoising iterations 4, 8, and 16. We then display attention to the conditioning-time visual tokens. The spatial grid is $7\times7$, displayed without smoothing. Attention color is density relative to a uniform spatial distribution after conditioning on the displayed view, using a common $0$--$4\times$ scale; error uses a common $[0,0.15]$ scale. Values beyond either scale are clipped for display only. Region enrichment is its fraction of attention divided by its fraction of image area; 1 denotes uniform allocation. Raw attention masses and layer/iteration maps are retained.

\begin{figure}[H]
\centering\includegraphics[width=\linewidth]{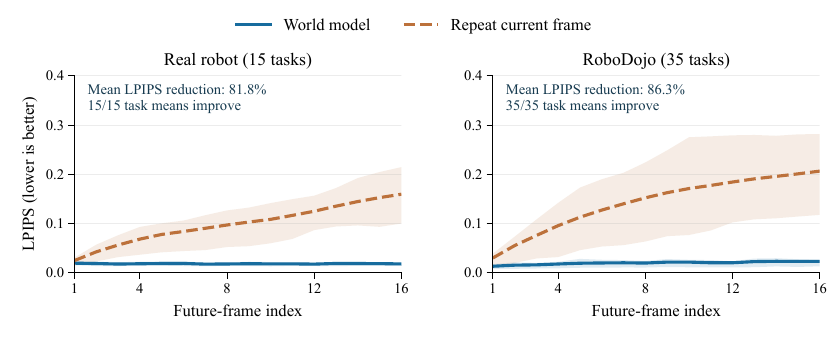}
\caption{\textbf{Prediction error over the full future horizon.} Solid blue curves show the world model. Dashed orange curves show the LPIPS between the unchanged current image and each recorded future frame: the current image is reused as the future prediction, without model inference. Lines are task means, and shaded bands show the 25th--75th percentiles across the selected task windows, not confidence intervals. Both panels use the same vertical scale. Relative reductions use all 16 frames; the condition is excluded.}
\label{fig:prediction-horizon}
\end{figure}

\paragraph{Scope of the evidence.}
Across these 15 selected examples, mean SSIM is 0.879 versus 0.814 for persistence, and mean LPIPS is 0.018 versus 0.095. Under the fixed aggregation, 1 of 15 task regions have attention enrichment above 1. Thus these maps do not establish preferential attention to task objects or accurate object recognition. They are descriptive internal diagnostics, not causal explanations. The examples come from training data and different task checkpoints; they do not establish held-out prediction quality, task success, or the separate contribution of response-based credit or agentic refinement. Incidental text and visible background faces are masked at paired locations; task numerals and mahjong symbols remain. Display editing is limited to deterministic masking, display downsampling, and compression.
\paragraph{Horizon-wise comparison.}
The fourth diagnostic panel in each task figure plots LPIPS over all 16 predicted frames, against the copy-condition baseline, using a common vertical scale within each domain. Figure~\ref{fig:prediction-horizon} aggregates all 15 real-robot and 35 simulation windows with equal task and frame weights. Full-horizon mean LPIPS decreases by 81.8\% for real robots and 86.3\% for simulation, computed as one minus the ratio of the model and baseline grand means. Every task has a lower full-horizon mean, although individual frames can favor persistence; all frames and tasks are retained. These comparisons demonstrate improved future-image fidelity over repeating the current observation on the selected training windows. They do not isolate the contribution of response-based credit or compare against an independently trained base model.

\begingroup
\setlength{\intextsep}{6pt}

\begin{figure}[H]
\centering\includegraphics[width=\linewidth]{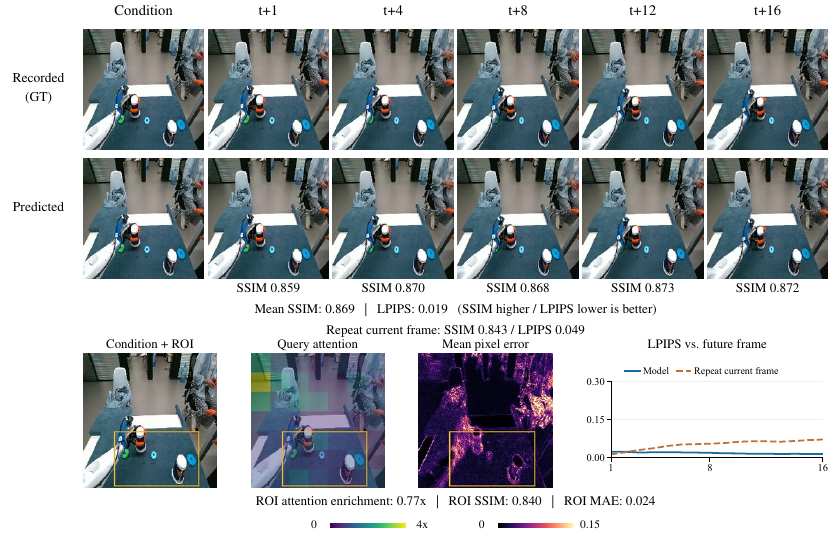}
\caption{\textbf{Task 01: Cover the largest and smallest circles.}}
\label{fig:wm-circle}
\end{figure}

\begin{figure}[H]
\centering\includegraphics[width=\linewidth]{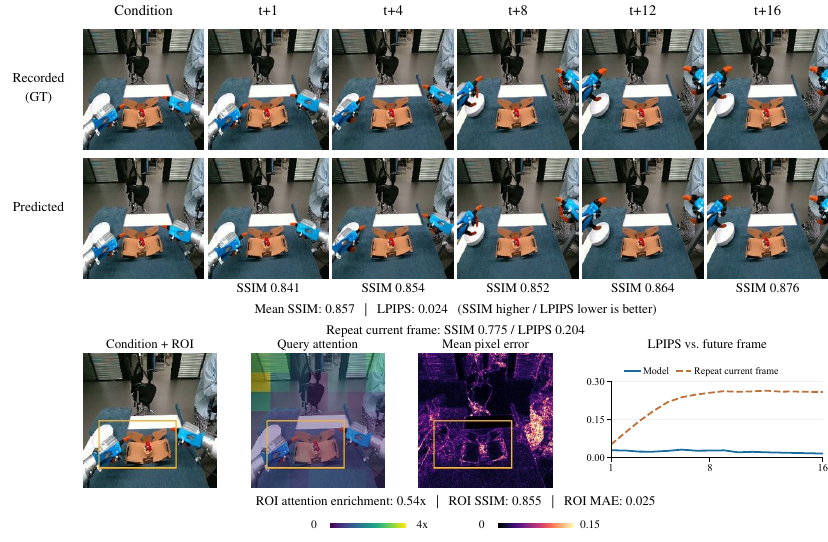}
\caption{\textbf{Task 02: Unpack the cake and place it in a bowl.}}
\label{fig:wm-cake}
\end{figure}

\begin{figure}[H]
\centering\includegraphics[width=\linewidth]{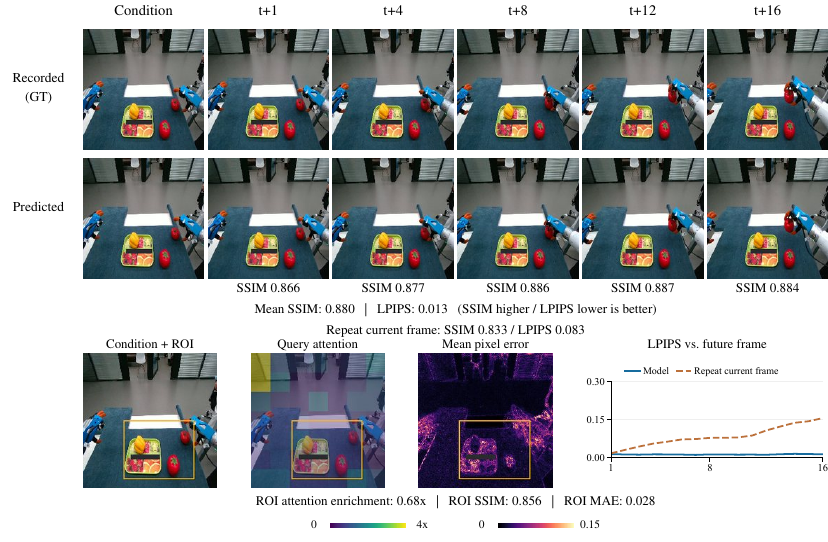}
\caption{\textbf{Task 03: Place fruits on a tray.}}
\label{fig:wm-apple}
\end{figure}

\begin{figure}[H]
\centering\includegraphics[width=\linewidth]{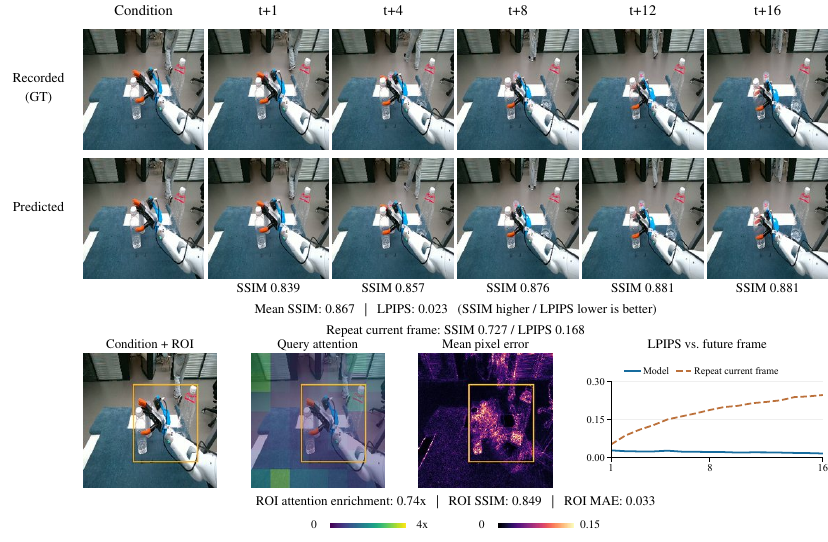}
\caption{\textbf{Task 04: Reorder three different types of bottles by height.}}
\label{fig:wm-highlow}
\end{figure}

\begin{figure}[H]
\centering\includegraphics[width=\linewidth]{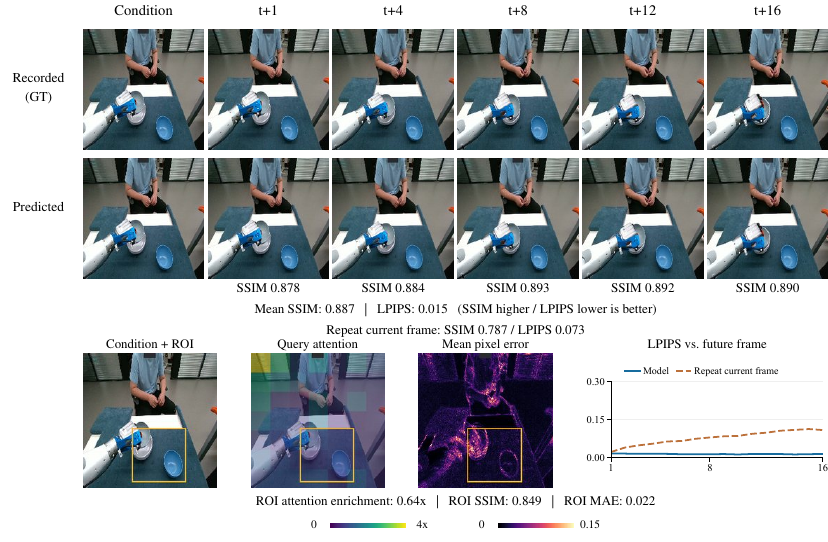}
\caption{\textbf{Task 05: Nest three bowls.}}
\label{fig:wm-bowl}
\end{figure}

\begin{figure}[H]
\centering\includegraphics[width=\linewidth]{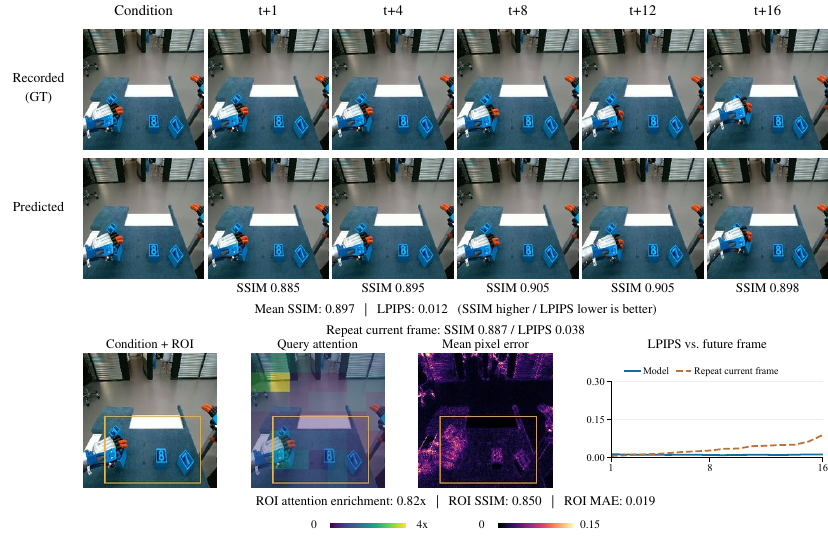}
\caption{\textbf{Task 06: Rearrange numbered blocks.}}
\label{fig:wm-number}
\end{figure}

\begin{figure}[H]
\centering\includegraphics[width=\linewidth]{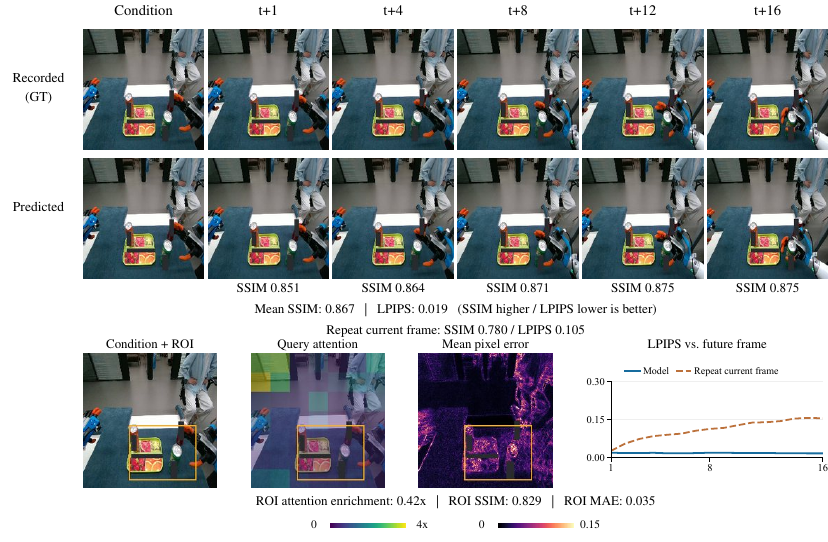}
\caption{\textbf{Task 07: Place three cans on a tray.}}
\label{fig:wm-can3}
\end{figure}

\begin{figure}[H]
\centering\includegraphics[width=\linewidth]{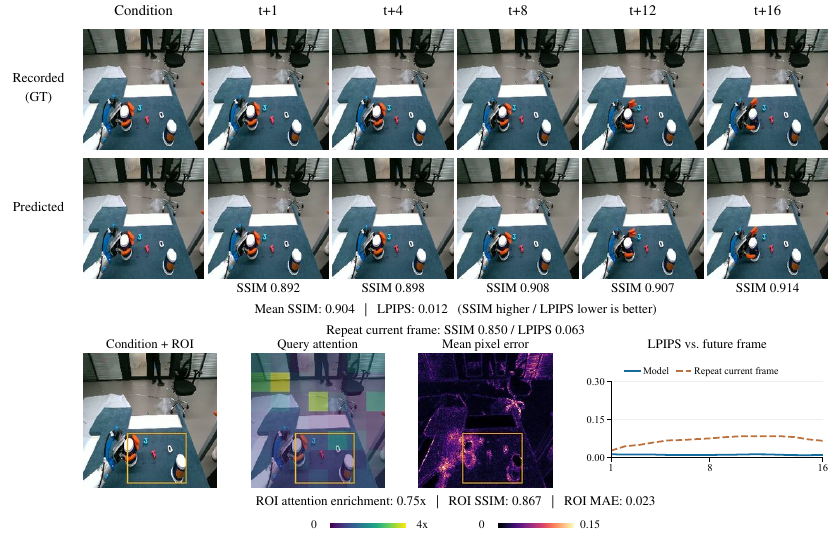}
\caption{\textbf{Task 08: Cover the largest and smallest numbers.}}
\label{fig:wm-bigsmall}
\end{figure}

\begin{figure}[H]
\centering\includegraphics[width=\linewidth]{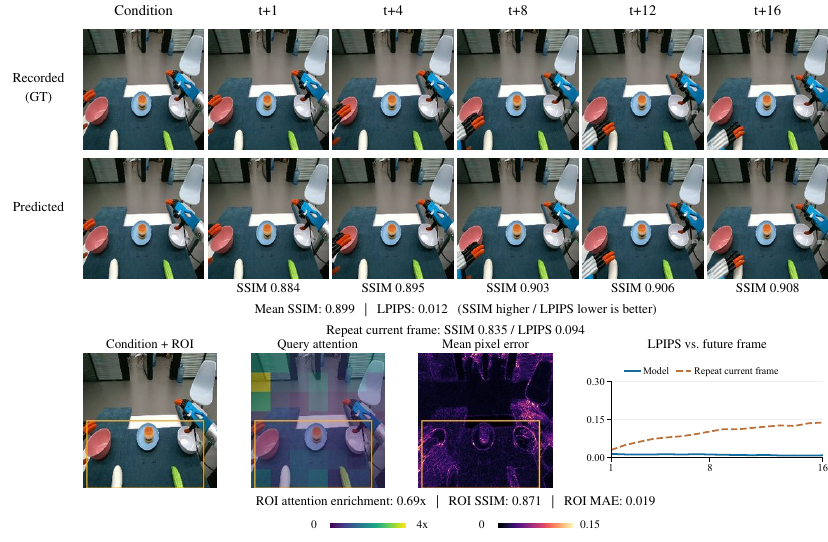}
\caption{\textbf{Task 09: Place food in separate dishes.}}
\label{fig:wm-threefood}
\end{figure}

\begin{figure}[H]
\centering\includegraphics[width=\linewidth]{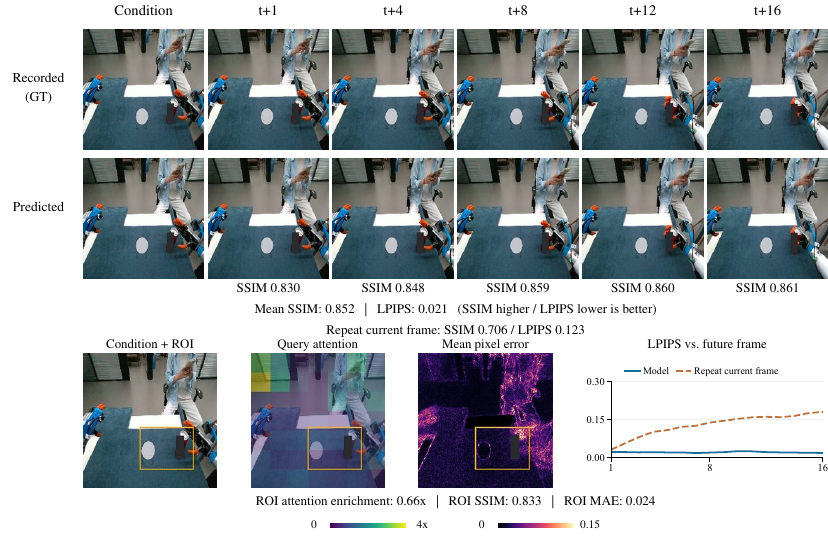}
\caption{\textbf{Task 10: Place a can on a coaster.}}
\label{fig:wm-can}
\end{figure}

\begin{figure}[H]
\centering\includegraphics[width=\linewidth]{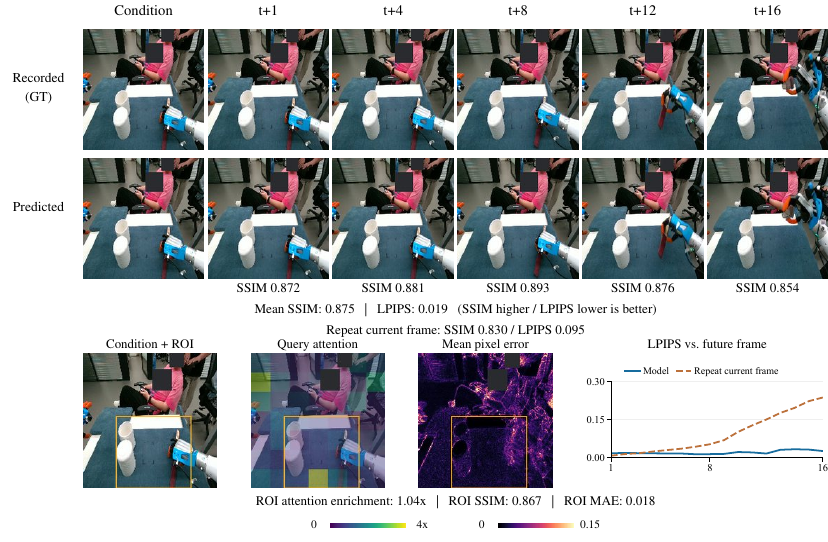}
\caption{\textbf{Task 11: Insert a folding fan into an open-top vessel.}}
\label{fig:wm-fan}
\end{figure}

\begin{figure}[H]
\centering\includegraphics[width=\linewidth]{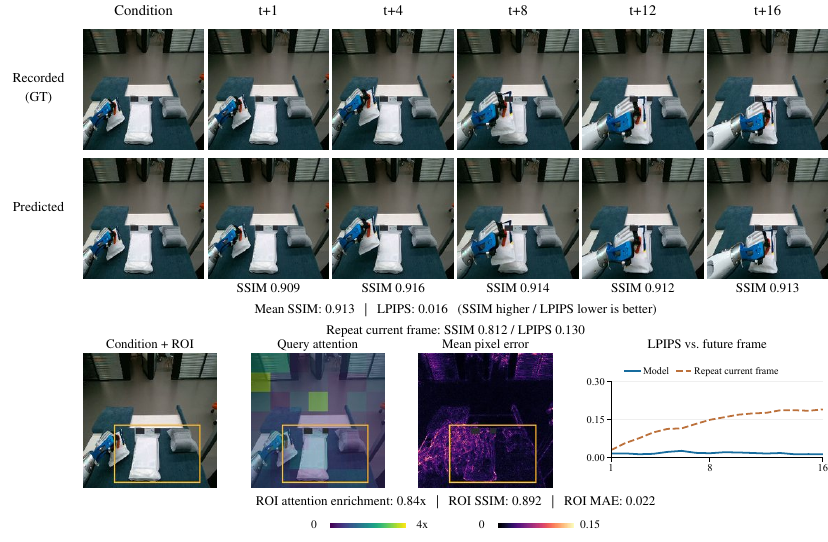}
\caption{\textbf{Task 12: Align disposable towels and place a thick towel on top.}}
\label{fig:wm-towel}
\end{figure}

\begin{figure}[H]
\centering\includegraphics[width=\linewidth]{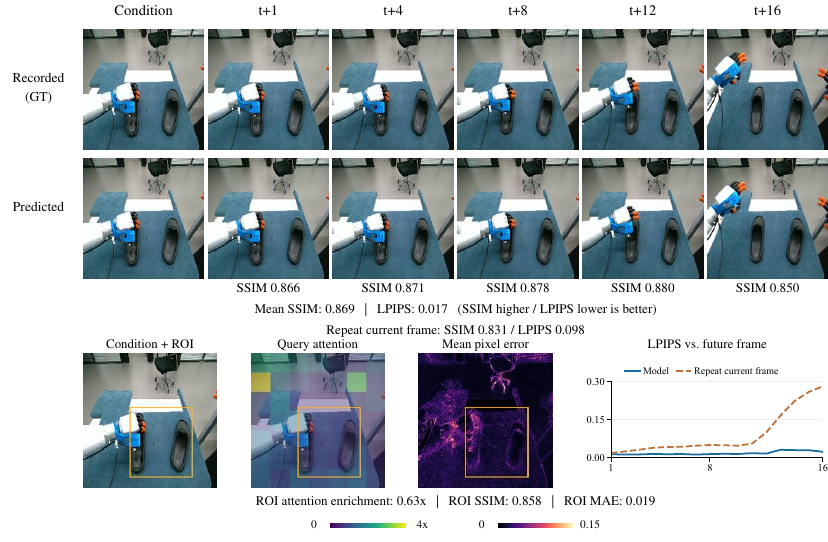}
\caption{\textbf{Task 13: Arrange a pair of shoes.}}
\label{fig:wm-shoes}
\end{figure}

\begin{figure}[H]
\centering\includegraphics[width=\linewidth]{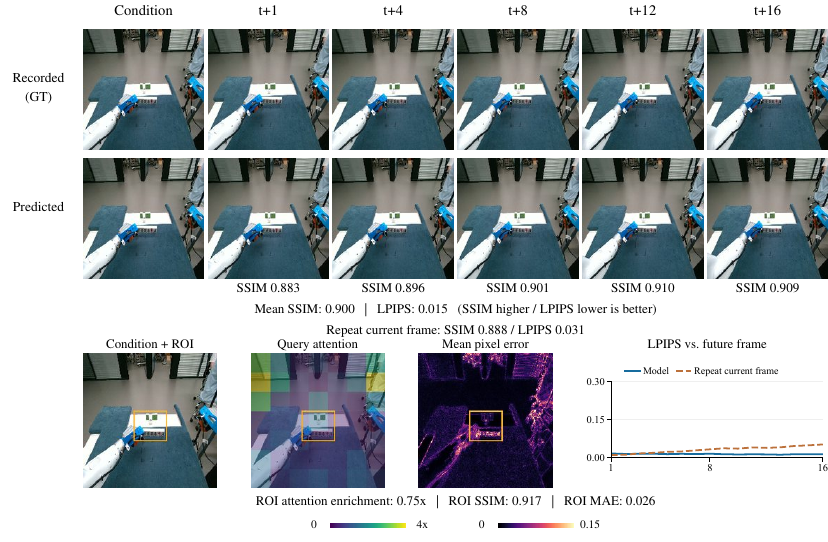}
\caption{\textbf{Task 14: Mahjong kong formation.}}
\label{fig:wm-gangpaifinal}
\end{figure}

\begin{figure}[H]
\centering\includegraphics[width=\linewidth]{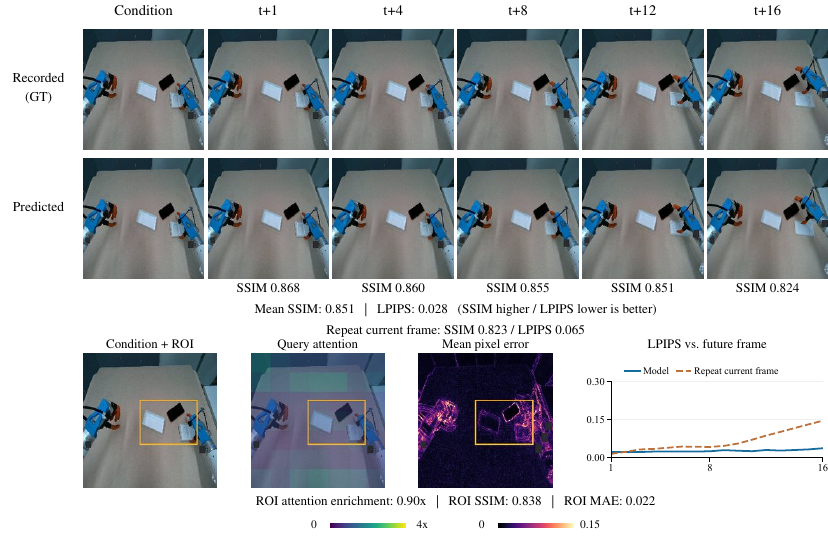}
\caption{\textbf{Task 16: Pack a phone into its box.}}
\label{fig:wm-phone}
\end{figure}
\endgroup

\section{Recorded and Predicted Futures in RoboDojo Simulation}
\label{app:robodojo-predictions}
Figures~\ref{fig:robodojo-task_00}--\ref{fig:robodojo-task_34} show one identified training example for each of the 35 RoboDojo training tasks (task IDs 00--34). All examples use the retained post-training checkpoint P09 (step 50,000) described in Appendix~\ref{app:agentic-protocol}. We preserve the training loader's seed-42 validation exclusion, select the median-length eligible training episode per task, and use its middle legal two-chunk window. Task IDs are verified against dataset metadata, and the corresponding task-language embedding is explicitly supplied during inference. Each figure pairs the recorded future with a 16-frame prediction sampled using 16 denoising steps. Inputs are the conditioning image, current state, and task-language embedding. The 30-step setting in Appendix~\ref{app:protocol} applies to policy evaluation. The original head-and-two-wrist camera mosaic is retained; task numbers denote dataset IDs, not difficulty ranks.

SSIM, LPIPS, the persistence baseline, and the fixed attention aggregation follow Appendix~\ref{app:real-world-predictions}. Global similarity metrics use the full original $384\times320$ camera mosaic. The diagnostic bottom strip shows only the head view, the upper $256\times320$ region; region metrics and attention enrichment are restricted to that view. The original attention grid is $12\times10$ over the mosaic, with an $8\times10$ head-view subgrid. It is displayed without smoothing across camera boundaries, with the same attention and error color scales as the real-robot figures. Region rectangles were marked from conditions and checked for visible task coverage, independently of attention magnitude.

Across the 35 selected training examples, mean SSIM is 0.942 versus 0.828 for the copy-condition baseline, and mean LPIPS is 0.018 versus 0.134. SSIM improves on persistence in 34 of 35 examples and LPIPS in all 35. None of the 35 marked regions has attention enrichment above 1 under this fixed aggregation, so these maps do not support a claim of object-localized attention or object-recognition accuracy. Coverage of all training task types is not a held-out evaluation: one window per task cannot establish average trajectory fidelity, causal component gains, or rollout success. Incidental text is masked for display while task-defining numbers, letters, and mahjong symbols are retained; metrics are computed before masking and compression.
\begingroup
\setlength{\intextsep}{6pt}

\begin{figure}[H]
\centering\includegraphics[width=\linewidth]{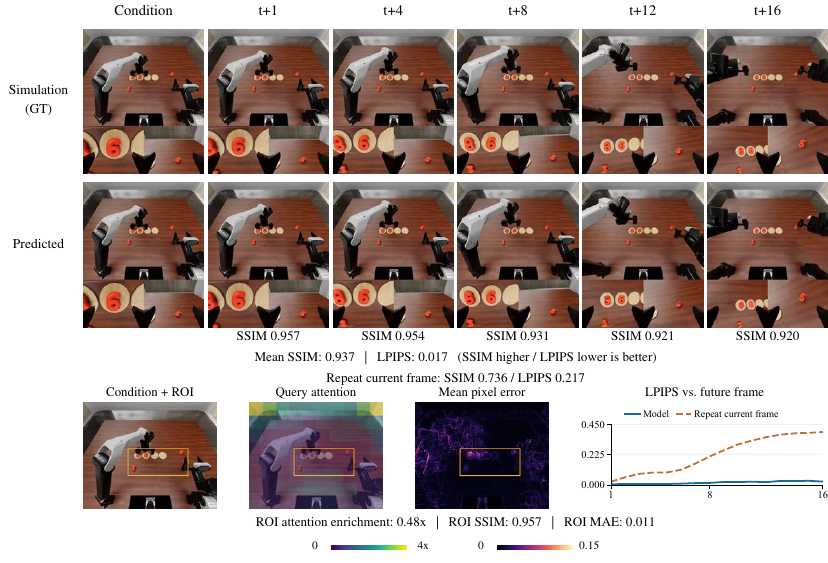}
\caption{\textbf{RoboDojo task 00: Arrange numbers.}}
\label{fig:robodojo-task_00}
\end{figure}

\begin{figure}[H]
\centering\includegraphics[width=\linewidth]{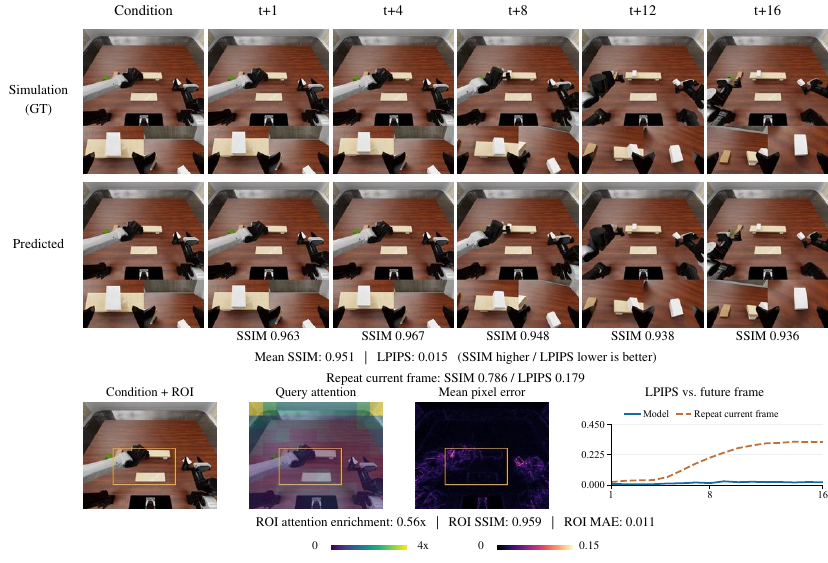}
\caption{\textbf{RoboDojo task 01: Build a tower.}}
\label{fig:robodojo-task_01}
\end{figure}

\begin{figure}[H]
\centering\includegraphics[width=\linewidth]{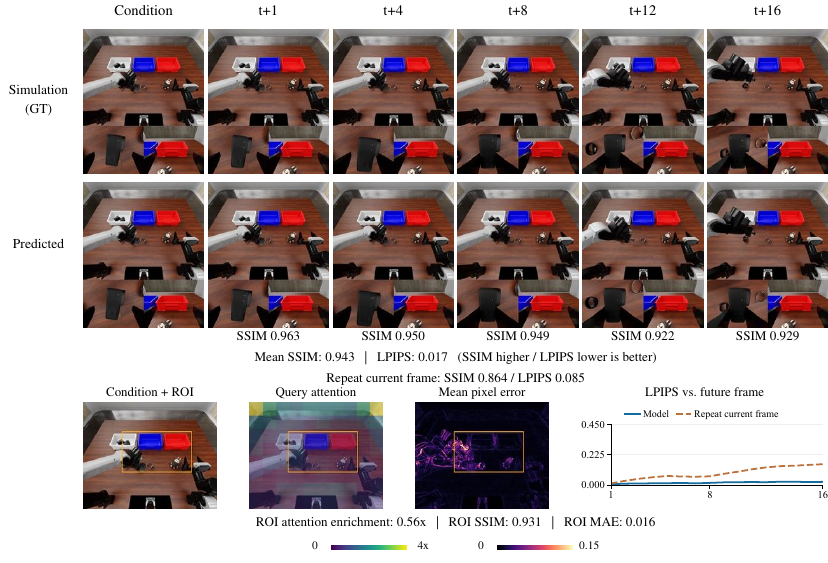}
\caption{\textbf{RoboDojo task 02: Sort objects by category.}}
\label{fig:robodojo-task_02}
\end{figure}

\begin{figure}[H]
\centering\includegraphics[width=\linewidth]{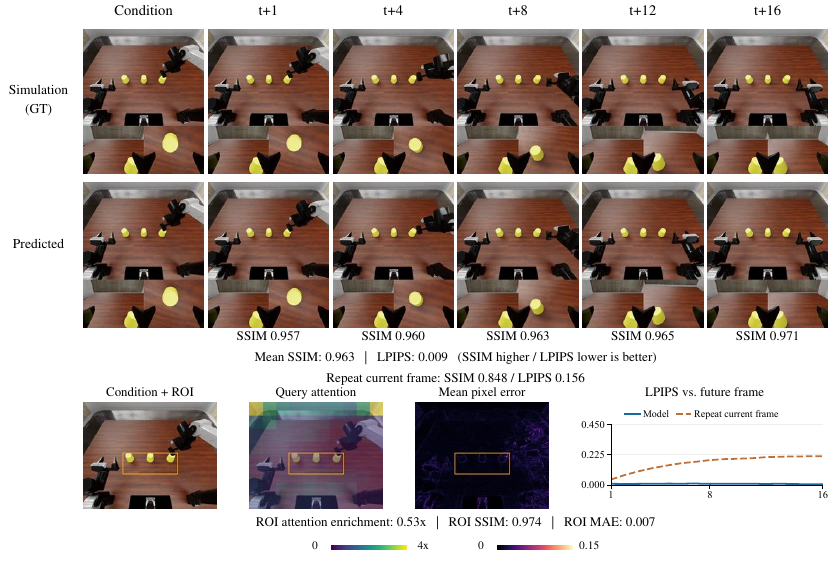}
\caption{\textbf{RoboDojo task 03: Remember covered block colors.}}
\label{fig:robodojo-task_03}
\end{figure}

\begin{figure}[H]
\centering\includegraphics[width=\linewidth]{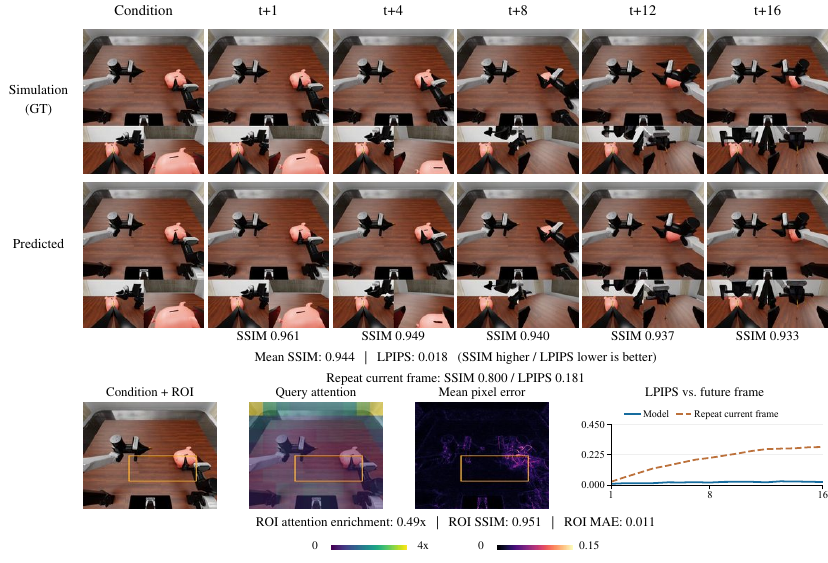}
\caption{\textbf{RoboDojo task 04: Insert a coin.}}
\label{fig:robodojo-task_04}
\end{figure}

\begin{figure}[H]
\centering\includegraphics[width=\linewidth]{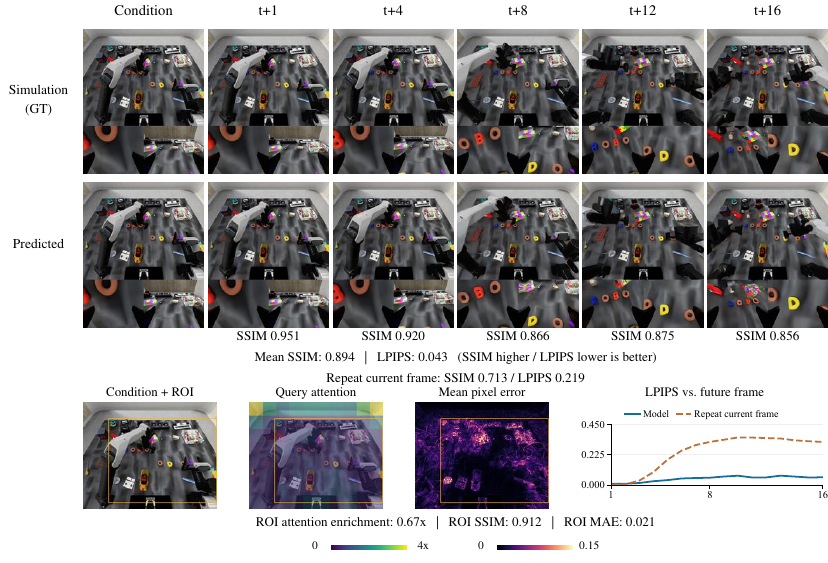}
\caption{\textbf{RoboDojo task 05: Spell RoboDojo.}}
\label{fig:robodojo-task_05}
\end{figure}

\begin{figure}[H]
\centering\includegraphics[width=\linewidth]{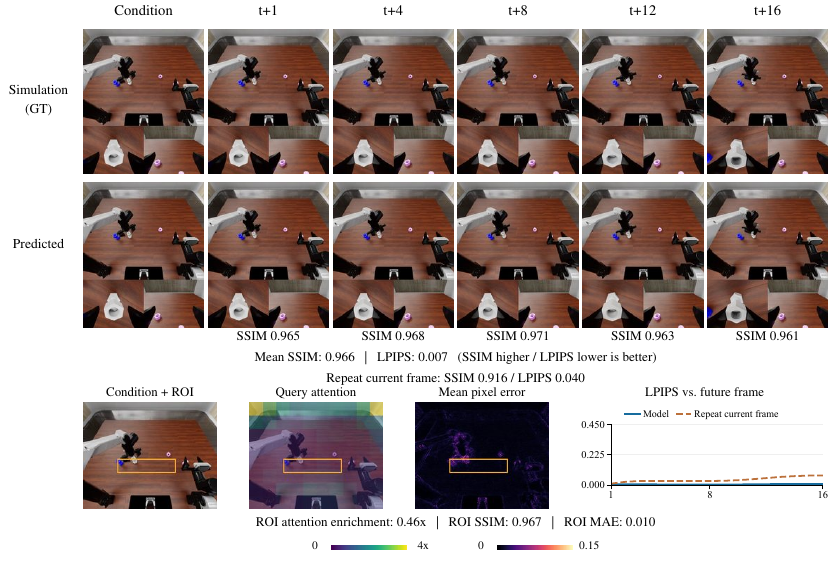}
\caption{\textbf{RoboDojo task 06: Tighten color-matched screws.}}
\label{fig:robodojo-task_06}
\end{figure}

\begin{figure}[H]
\centering\includegraphics[width=\linewidth]{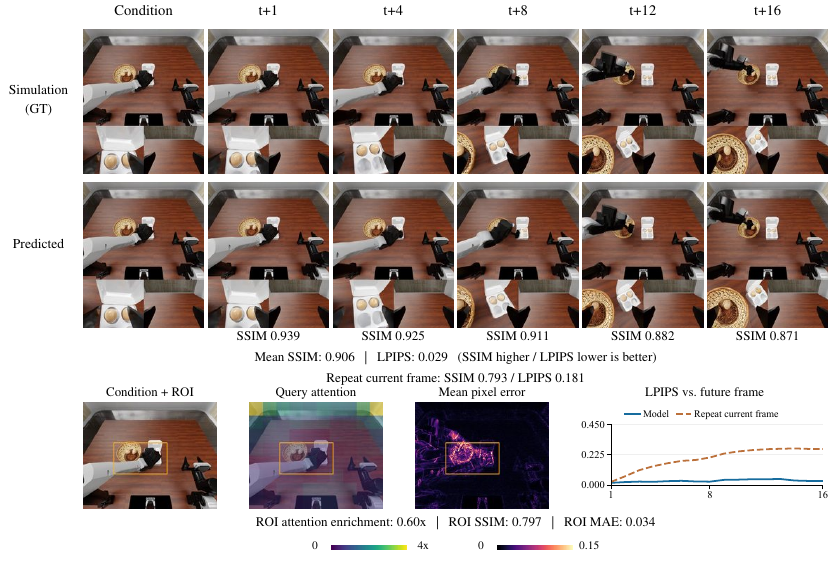}
\caption{\textbf{RoboDojo task 07: Store eggs and close the holder.}}
\label{fig:robodojo-task_07}
\end{figure}

\begin{figure}[H]
\centering\includegraphics[width=\linewidth]{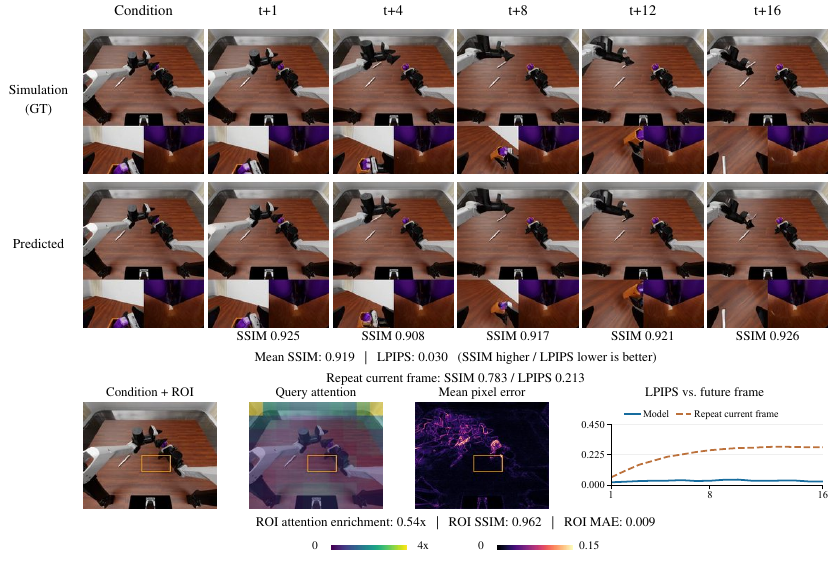}
\caption{\textbf{RoboDojo task 08: Put pens into a holder.}}
\label{fig:robodojo-task_08}
\end{figure}

\begin{figure}[H]
\centering\includegraphics[width=\linewidth]{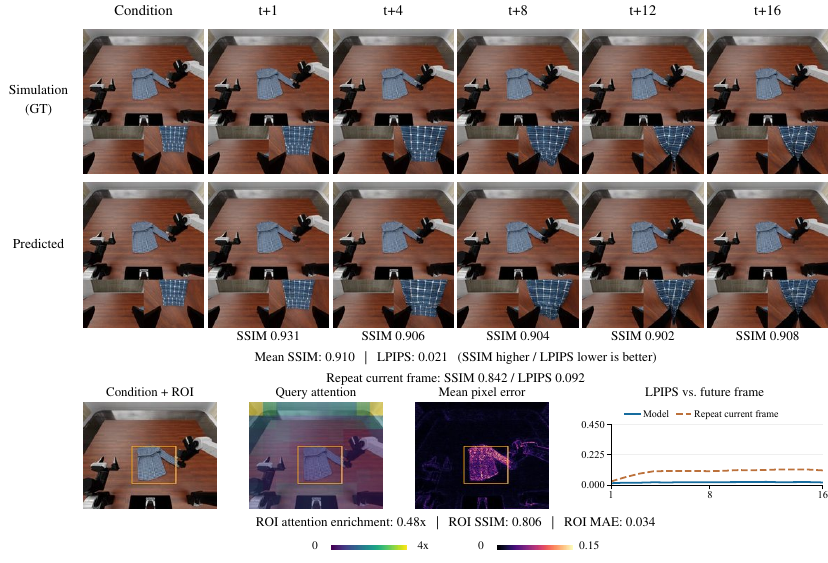}
\caption{\textbf{RoboDojo task 09: Fold clothes.}}
\label{fig:robodojo-task_09}
\end{figure}

\begin{figure}[H]
\centering\includegraphics[width=\linewidth]{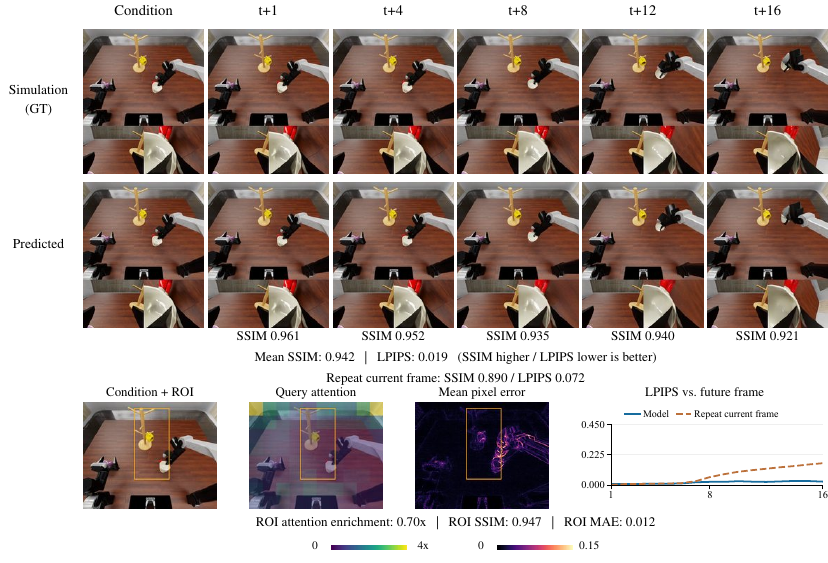}
\caption{\textbf{RoboDojo task 10: Hang mugs.}}
\label{fig:robodojo-task_10}
\end{figure}

\begin{figure}[H]
\centering\includegraphics[width=\linewidth]{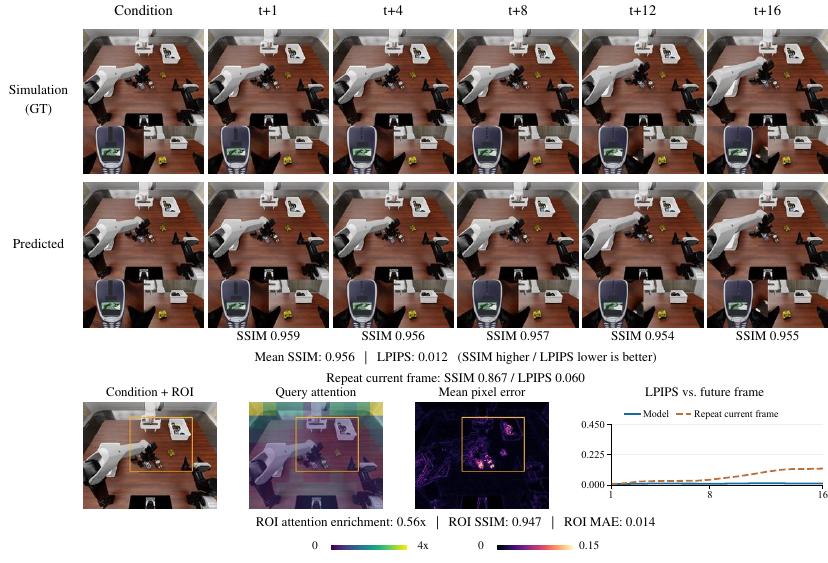}
\caption{\textbf{RoboDojo task 11: Recall object placement order.}}
\label{fig:robodojo-task_11}
\end{figure}

\begin{figure}[H]
\centering\includegraphics[width=\linewidth]{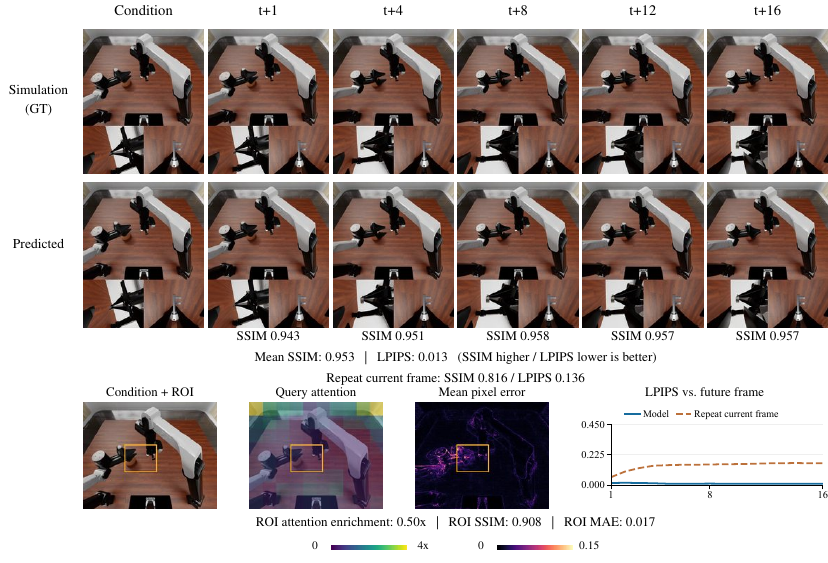}
\caption{\textbf{RoboDojo task 12: Insert a thick card.}}
\label{fig:robodojo-task_12}
\end{figure}

\begin{figure}[H]
\centering\includegraphics[width=\linewidth]{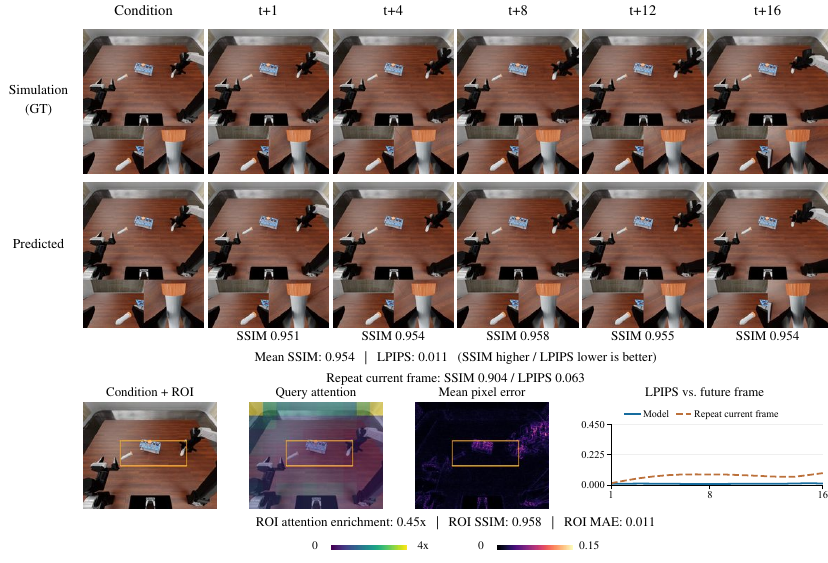}
\caption{\textbf{RoboDojo task 13: Place tubes in a rack.}}
\label{fig:robodojo-task_13}
\end{figure}

\begin{figure}[H]
\centering\includegraphics[width=\linewidth]{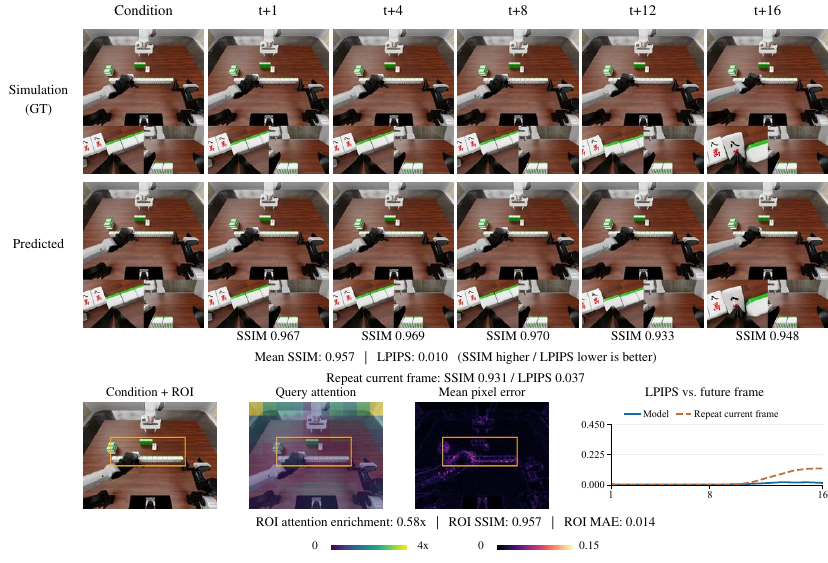}
\caption{\textbf{RoboDojo task 14: Form a mahjong kong.}}
\label{fig:robodojo-task_14}
\end{figure}

\begin{figure}[H]
\centering\includegraphics[width=\linewidth]{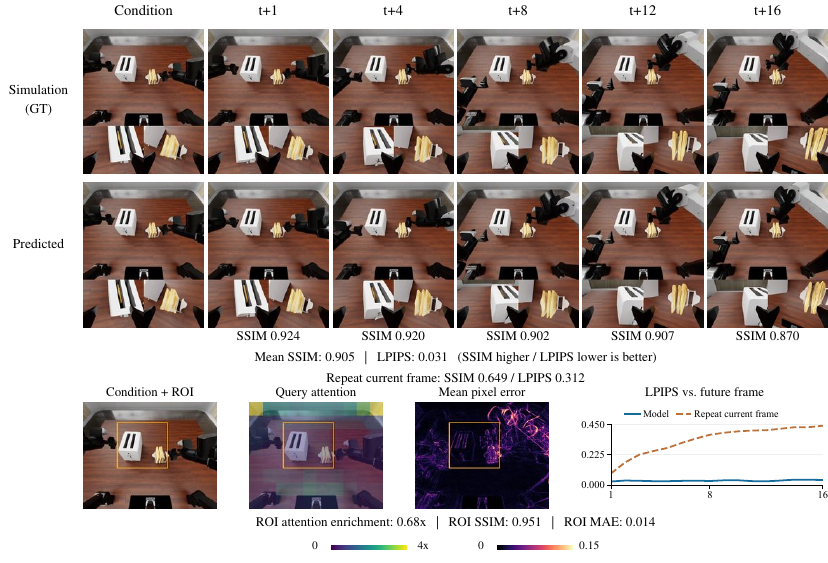}
\caption{\textbf{RoboDojo task 15: Toast bread.}}
\label{fig:robodojo-task_15}
\end{figure}

\begin{figure}[H]
\centering\includegraphics[width=\linewidth]{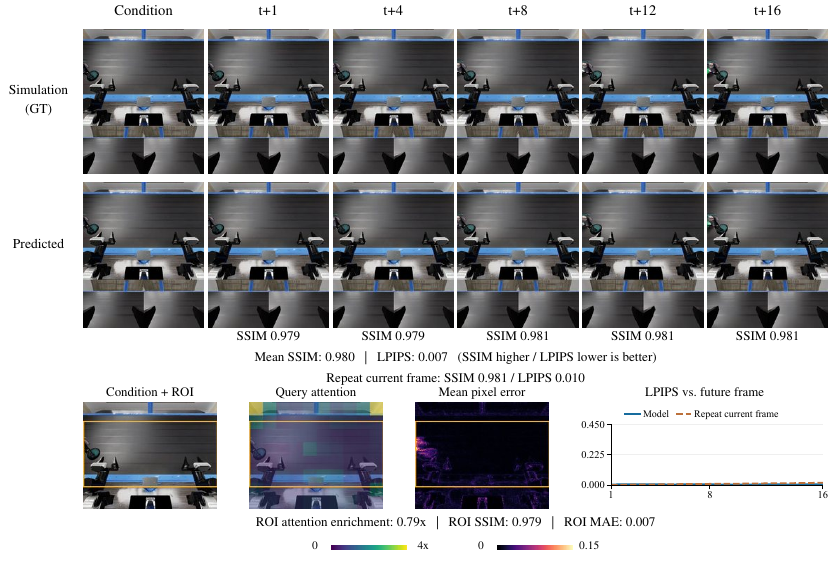}
\caption{\textbf{RoboDojo task 16: Match a remembered conveyor object.}}
\label{fig:robodojo-task_16}
\end{figure}

\begin{figure}[H]
\centering\includegraphics[width=\linewidth]{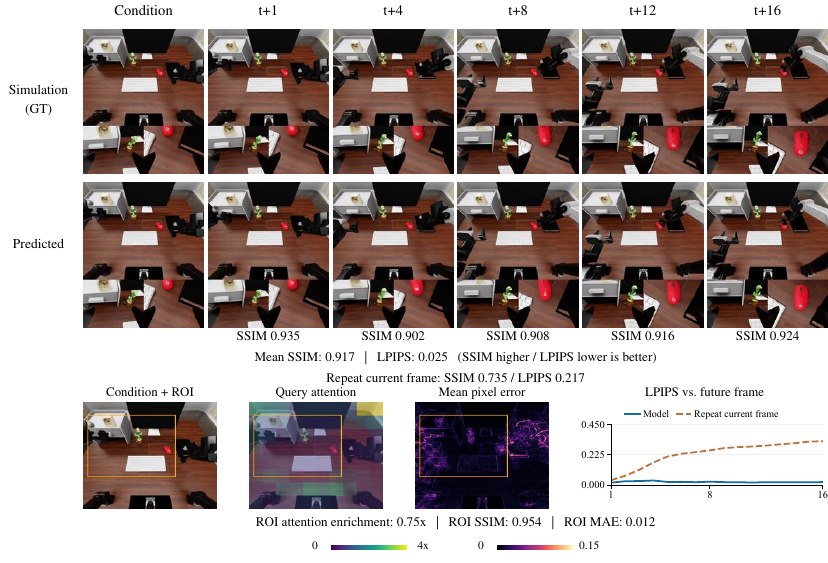}
\caption{\textbf{RoboDojo task 17: Organize a desk.}}
\label{fig:robodojo-task_17}
\end{figure}

\begin{figure}[H]
\centering\includegraphics[width=\linewidth]{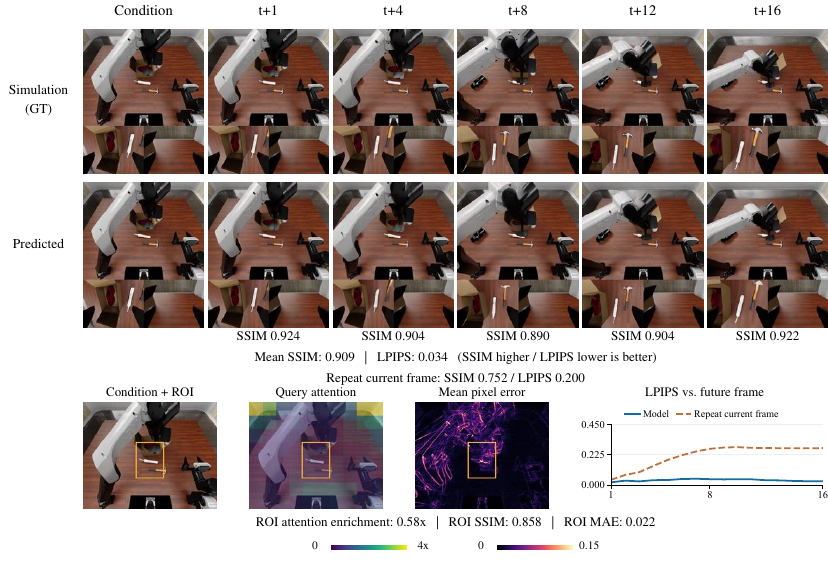}
\caption{\textbf{RoboDojo task 18: Pack objects facing left.}}
\label{fig:robodojo-task_18}
\end{figure}

\begin{figure}[H]
\centering\includegraphics[width=\linewidth]{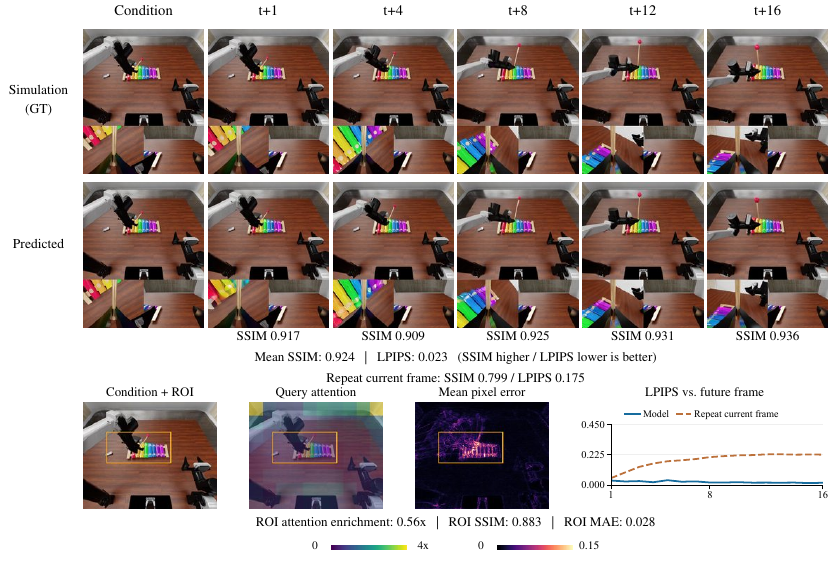}
\caption{\textbf{RoboDojo task 19: Play the xylophone.}}
\label{fig:robodojo-task_19}
\end{figure}

\begin{figure}[H]
\centering\includegraphics[width=\linewidth]{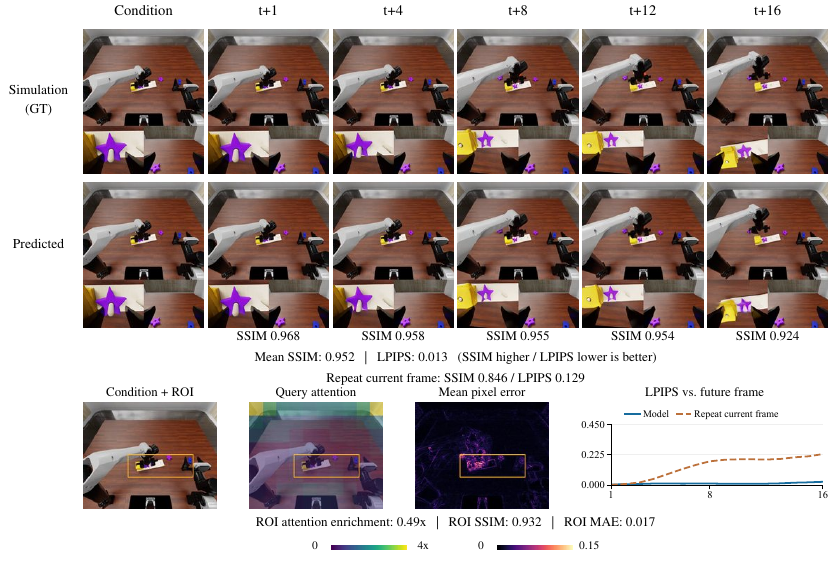}
\caption{\textbf{RoboDojo task 20: Assemble a stacking toy.}}
\label{fig:robodojo-task_20}
\end{figure}

\begin{figure}[H]
\centering\includegraphics[width=\linewidth]{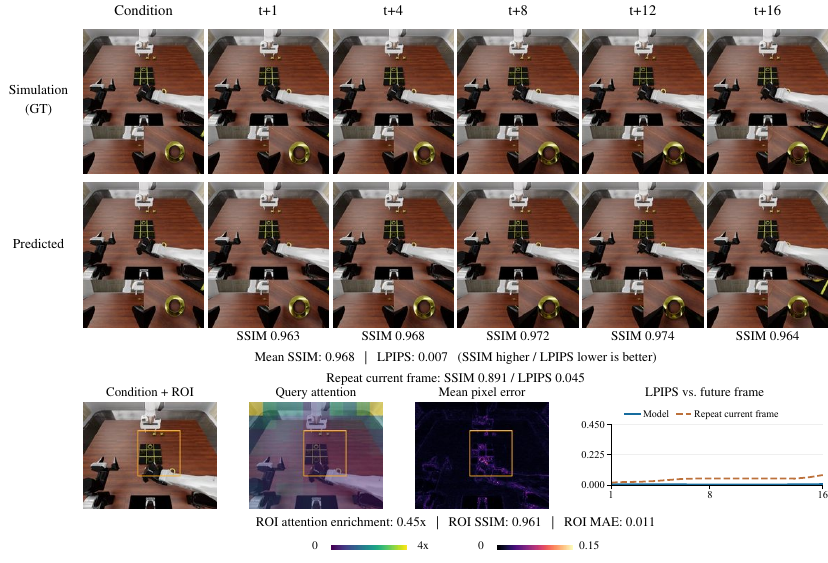}
\caption{\textbf{RoboDojo task 21: Play tic-tac-toe.}}
\label{fig:robodojo-task_21}
\end{figure}

\begin{figure}[H]
\centering\includegraphics[width=\linewidth]{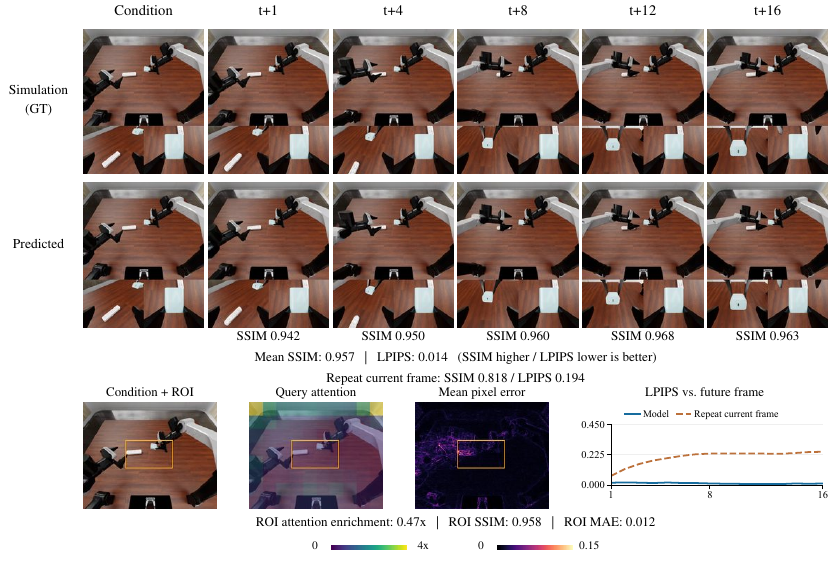}
\caption{\textbf{RoboDojo task 22: Plug in a charger.}}
\label{fig:robodojo-task_22}
\end{figure}

\begin{figure}[H]
\centering\includegraphics[width=\linewidth]{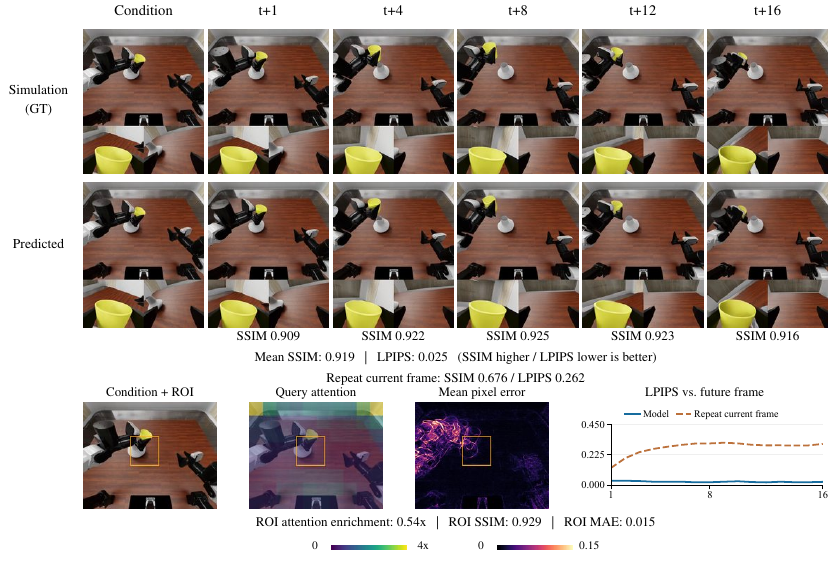}
\caption{\textbf{RoboDojo task 23: Pour balls into a vase.}}
\label{fig:robodojo-task_23}
\end{figure}

\begin{figure}[H]
\centering\includegraphics[width=\linewidth]{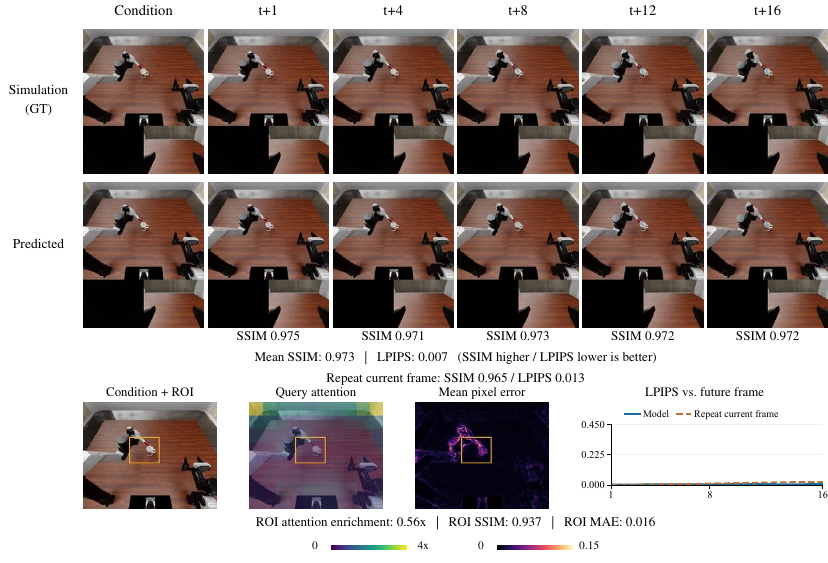}
\caption{\textbf{RoboDojo task 24: Pour liquid into a cup.}}
\label{fig:robodojo-task_24}
\end{figure}

\begin{figure}[H]
\centering\includegraphics[width=\linewidth]{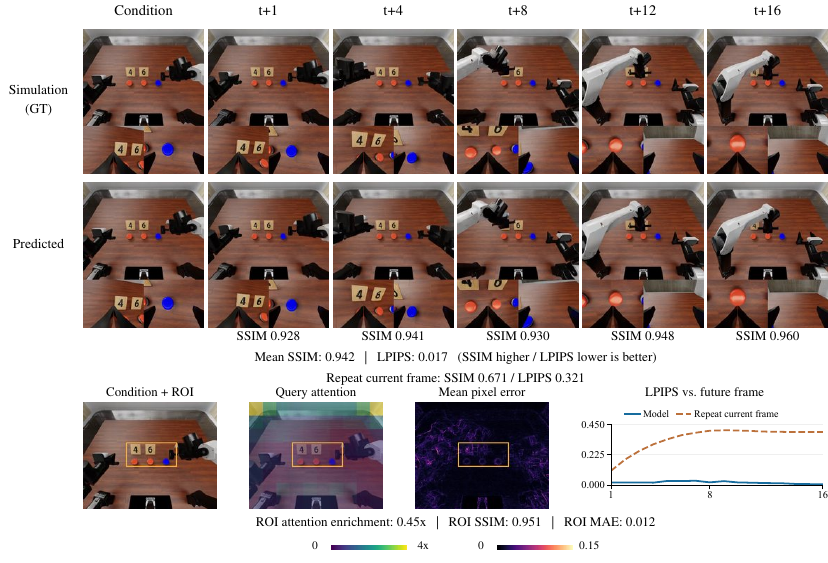}
\caption{\textbf{RoboDojo task 25: Count button presses.}}
\label{fig:robodojo-task_25}
\end{figure}

\begin{figure}[H]
\centering\includegraphics[width=\linewidth]{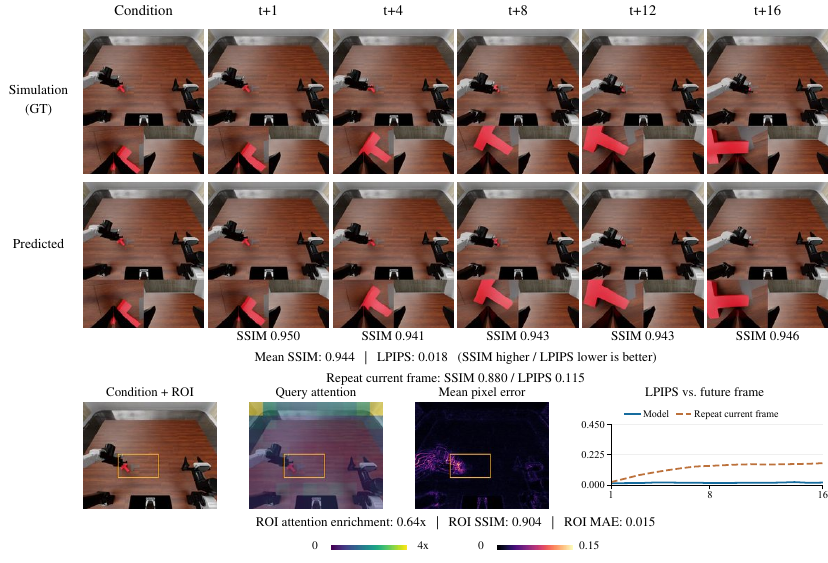}
\caption{\textbf{RoboDojo task 26: Align a T-shaped block.}}
\label{fig:robodojo-task_26}
\end{figure}

\begin{figure}[H]
\centering\includegraphics[width=\linewidth]{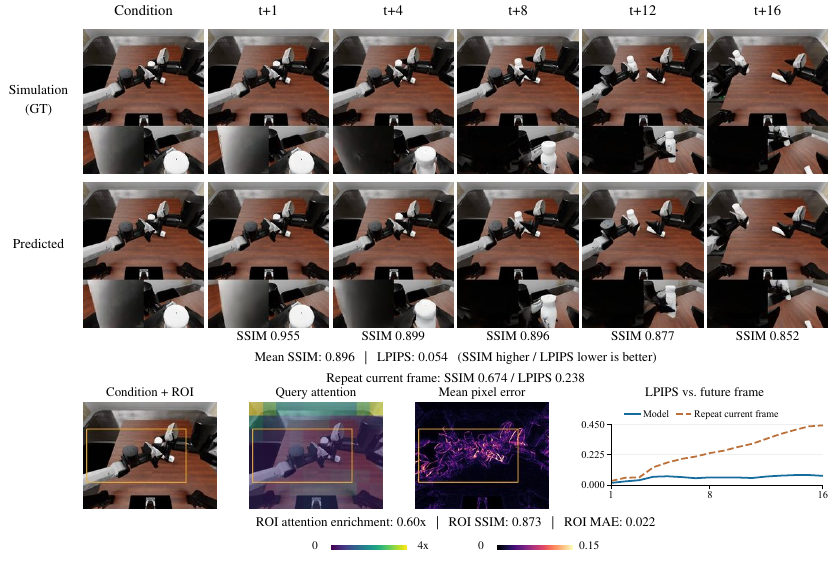}
\caption{\textbf{RoboDojo task 27: Discard bottles.}}
\label{fig:robodojo-task_27}
\end{figure}

\begin{figure}[H]
\centering\includegraphics[width=\linewidth]{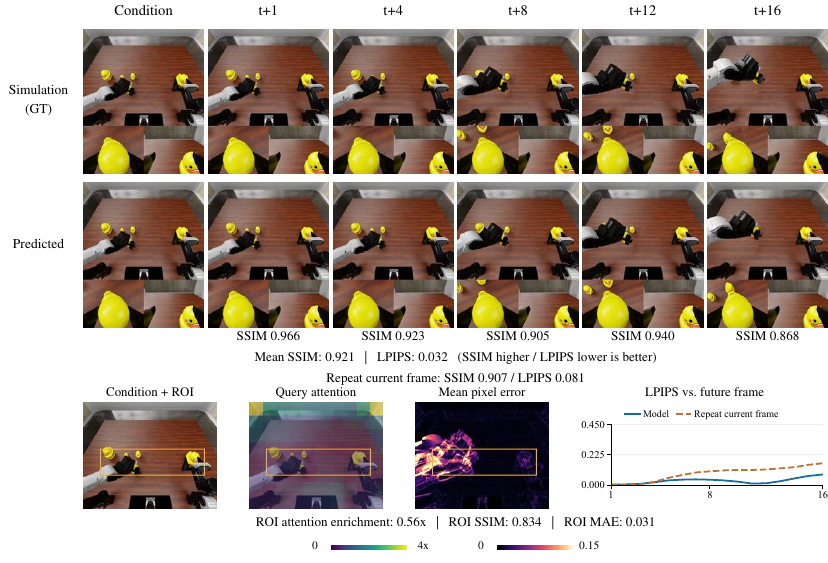}
\caption{\textbf{RoboDojo task 28: Order nesting dolls.}}
\label{fig:robodojo-task_28}
\end{figure}

\begin{figure}[H]
\centering\includegraphics[width=\linewidth]{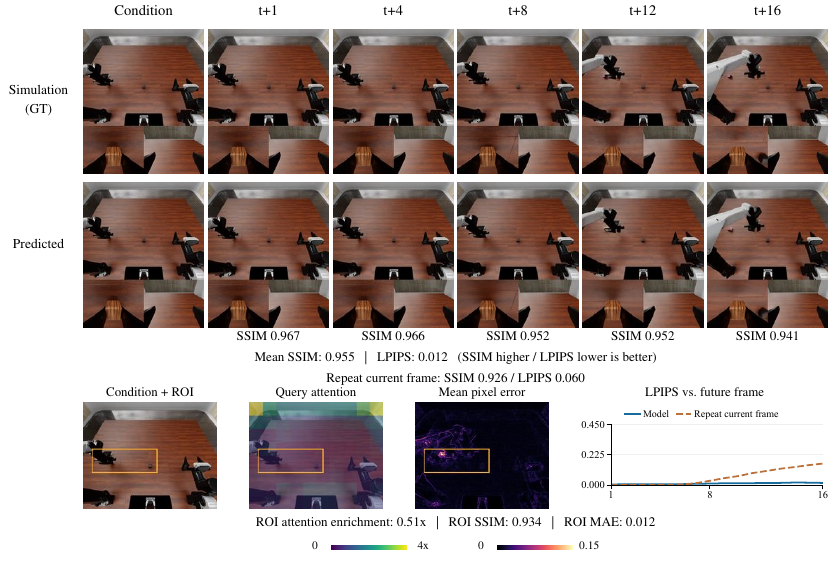}
\caption{\textbf{RoboDojo task 29: Stack textured blocks.}}
\label{fig:robodojo-task_29}
\end{figure}

\begin{figure}[H]
\centering\includegraphics[width=\linewidth]{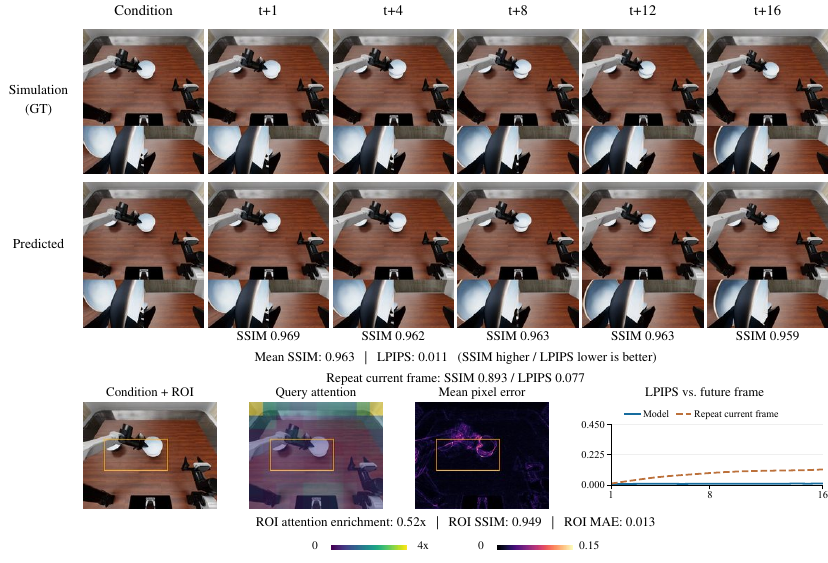}
\caption{\textbf{RoboDojo task 30: Stack bowls.}}
\label{fig:robodojo-task_30}
\end{figure}

\begin{figure}[H]
\centering\includegraphics[width=\linewidth]{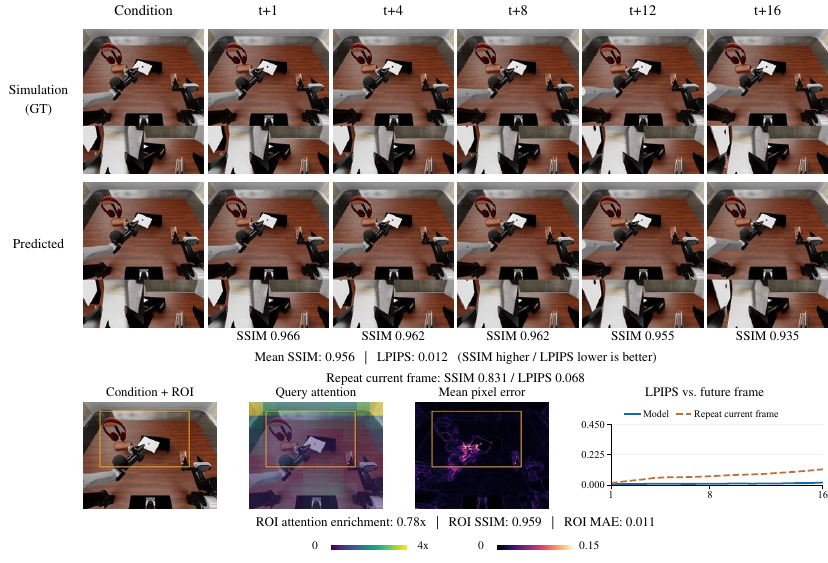}
\caption{\textbf{RoboDojo task 31: Store headphones and a laptop.}}
\label{fig:robodojo-task_31}
\end{figure}

\begin{figure}[H]
\centering\includegraphics[width=\linewidth]{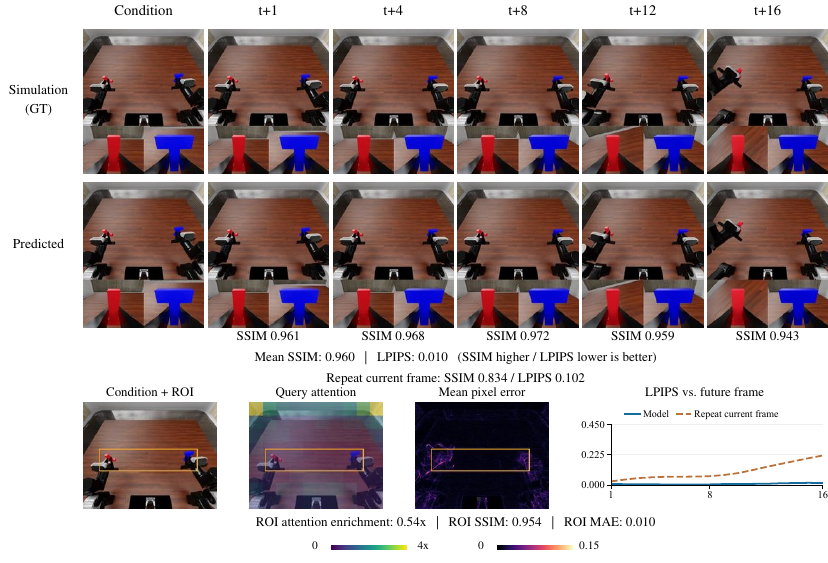}
\caption{\textbf{RoboDojo task 32: Swap T-shaped blocks.}}
\label{fig:robodojo-task_32}
\end{figure}

\begin{figure}[H]
\centering\includegraphics[width=\linewidth]{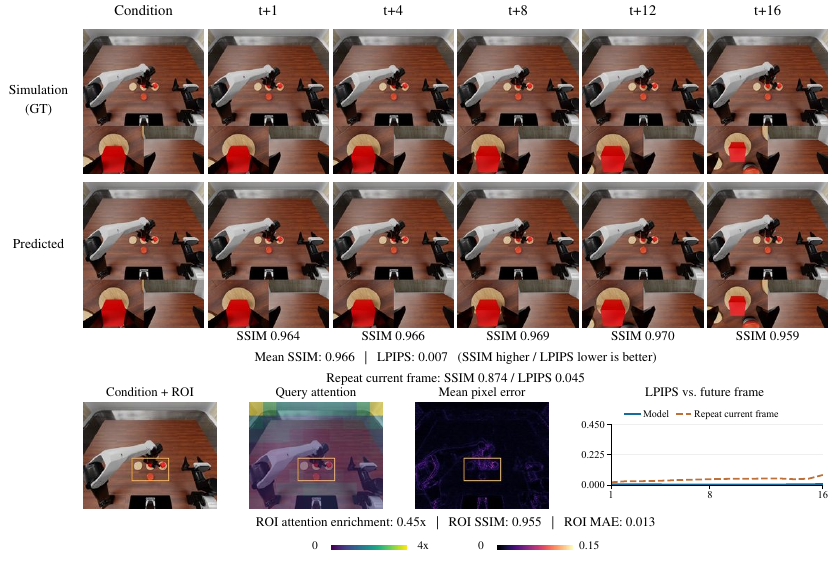}
\caption{\textbf{RoboDojo task 33: Swap blocks via an empty mat.}}
\label{fig:robodojo-task_33}
\end{figure}

\begin{figure}[H]
\centering\includegraphics[width=\linewidth]{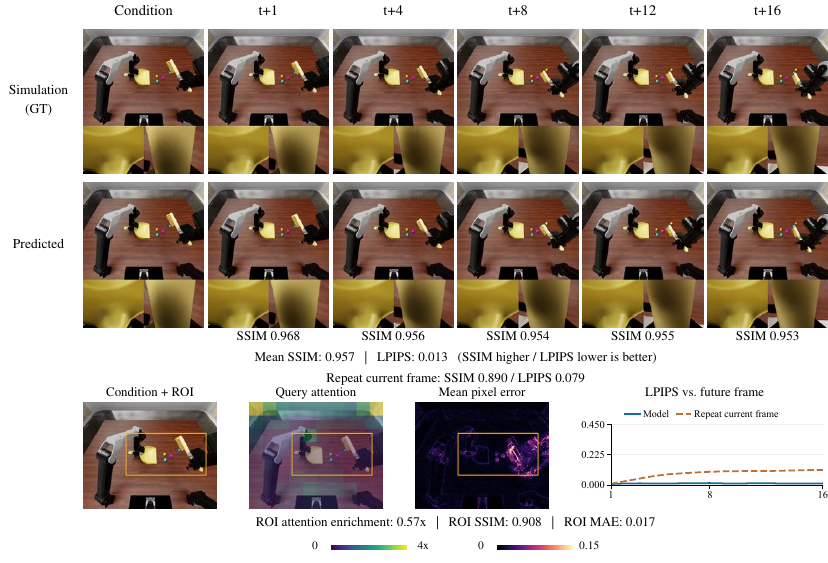}
\caption{\textbf{RoboDojo task 34: Sweep blocks.}}
\label{fig:robodojo-task_34}
\end{figure}
\endgroup

\end{document}